\documentclass{article}

\usepackage{PRIMEarxiv}

\usepackage[utf8]{inputenc} 
\usepackage[T1]{fontenc}    
\usepackage{hyperref}       
\usepackage{url}            
\usepackage{booktabs}       
\usepackage{amsfonts}       
\usepackage{nicefrac}       
\usepackage{microtype}      
\usepackage{lipsum}
\usepackage{fancyhdr}       
\usepackage{graphicx}       

\usepackage{balance} 
\usepackage{algorithm}
\usepackage{algorithmic}
\usepackage{multirow}
\usepackage{graphicx}

\usepackage{subcaption}
\usepackage[table,dvipsnames]{xcolor}
\usepackage{placeins}
\usepackage{hyperref}
\usepackage{xr-hyper}
\usepackage{natbib}
\usepackage{amsmath}
\usepackage{amsthm}
\usepackage{tabularx}

\newcolumntype{Y}{>{\raggedright\arraybackslash}X}

\newcommand{\E}{\mathbb{E}}
\newcommand{\I}{\mathbb{I}}

\newtheorem{lemma}{Lemma}
\newtheorem{theorem}{Theorem}
\newtheorem{proposition}{Proposition}
\newtheorem{corollary}{Corollary}

\graphicspath{{media/}}     

\title{Reward-Weighted Preference Optimization for Selective Combinatorial Optimization Problems
}

\author{
  Haohui Zhang \\
  University of Twente \\
  Enschede \\
  Netherlands\\
  \texttt{h.zhang-5@utwente.nl} \\
   \And
  Xinyu Hu \\
  University of Bath \\
  Bath \\
  United Kingdom\\
  \texttt{xh922@bath.ac.uk} \\
  \AND
  Wouter van Heeswijk \\
  University of Twente \\
  Enschede \\
  Netherlands \\
  \texttt{w.j.a.vanheeswijk@utwente.nl} \\
  \And
  Neil Yorke-Smith \\
  Delft University of Technology \\
  Delft \\
  Netherlands \\
  \texttt{n.yorke-smith@tudelft.nl} \\
  \And
  Martijn Mes \\
  University of Twente \\
  Enschede \\
  Netherlands \\
  \texttt{m.r.k.mes@utwente.nl} \\
}

\begin{document}
\maketitle

\begin{abstract}
Reinforcement learning has emerged as an effective approach for combinatorial optimization, but most existing methods focus on problems such as the Travelling Salesperson Problem (TSP) in which all elements must be served.  Selective combinatorial optimization problems (SCOPs), including the Orienteering Problem (OP), Prize-Collecting TSP (PCTSP), Stochastic PCTSP and Knapsack Problem (KP), additionally require selecting which elements to include.  This selection creates sampled solutions with substantially different objective values, yet standard preference optimization reduces such comparisons to binary rankings and discards the magnitude of these differences.

We introduce Reward-Weighted Preference Optimization (RWPO), a pairwise training objective that incorporates reward-gap information into preference learning.  For candidate trajectories sampled from the same instance, RWPO weights each preference loss according to the normalized difference in their rewards, assigning greater training emphasis to comparisons that represent larger improvements in solution quality.  We further combine RWPO with scale-specific fine-tuning and reward- and likelihood-weighted tilted sampling to adapt trained policies to larger and shifted instance distributions without relying on local-search post-processing.

Experiments on the OP, PCTSP, Stochastic PCTSP, and KP show that RWPO consistently improves over unweighted preference optimization and policy-gradient baselines across a wide range of instance sizes. On the OP, the resulting solver exceeds the mean reward of the Compass metaheuristic at scales from 100 to 2,000 nodes.
The results demonstrate that preserving reward-gap information can substantially improve preference-based learning for selective combinatorial optimization and support generalization to large-scale instances.
\end{abstract}

\keywords{Combinatorial Optimization\and Reinforcement Learning \and Preference Optimization \and Large-Scale Combinatorial Optimization \and 
Vehicle Routing Problems \and Machine Learning}

\section{Introduction}
\label{sec:intro}

Many routing and resource allocation problems involve deciding which customers to serve or which items to select under resource constraints. These decisions are naturally formulated as \emph{selective combinatorial optimization problems (SCOPs)}, including the Orienteering Problem (OP), Prize-Collecting Travelling Salesperson Problem (PCTSP), Stochastic PCTSP (SPCTSP), and the Knapsack Problem (KP) \citep{gunawan2016orienteering,balas1989prize,martello1990knapsack}. Unlike the canonical Travelling Salesperson Problem (TSP) and Capacitated Vehicle Routing Problem (CVRP), SCOPs require choosing which elements to include in a feasible solution. In selective routing, this choice also affects route construction: visiting a customer changes both the objective value and the resources required to maintain feasibility. 
Sequential construction therefore produces candidate solutions with potentially large differences in objective value. Preference-based training can exploit these comparisons, but binary preference labels discard the magnitude of these quality differences.

Neural constructive solvers have emerged as effective approaches to combinatorial optimization, learning policies that construct solutions sequentially and enabling fast inference on new instances \cite{bengio2021machine,kool2019attention}. To tackle large-scale combinatorial optimization problems (COPs), recent efforts have introduced sophisticated scaling techniques such as divide-and-conquer strategies and advanced search guidance \cite{cheng2023select,hou2023generalize,zheng2024udc,ye2024glop,pan2025hierarchical,chen2026dragon}. Although neural-based methods have been applied to selective problems such as OP, PCTSP, and KP \cite{kool2019attention,xin2021multi,kim2022sym,drakulic2023bq,zheng2024udc,ye2024glop}, their performance at large and shifted instance scales remains challenging.

\begin{figure*}[t]
\includegraphics[width=\linewidth]{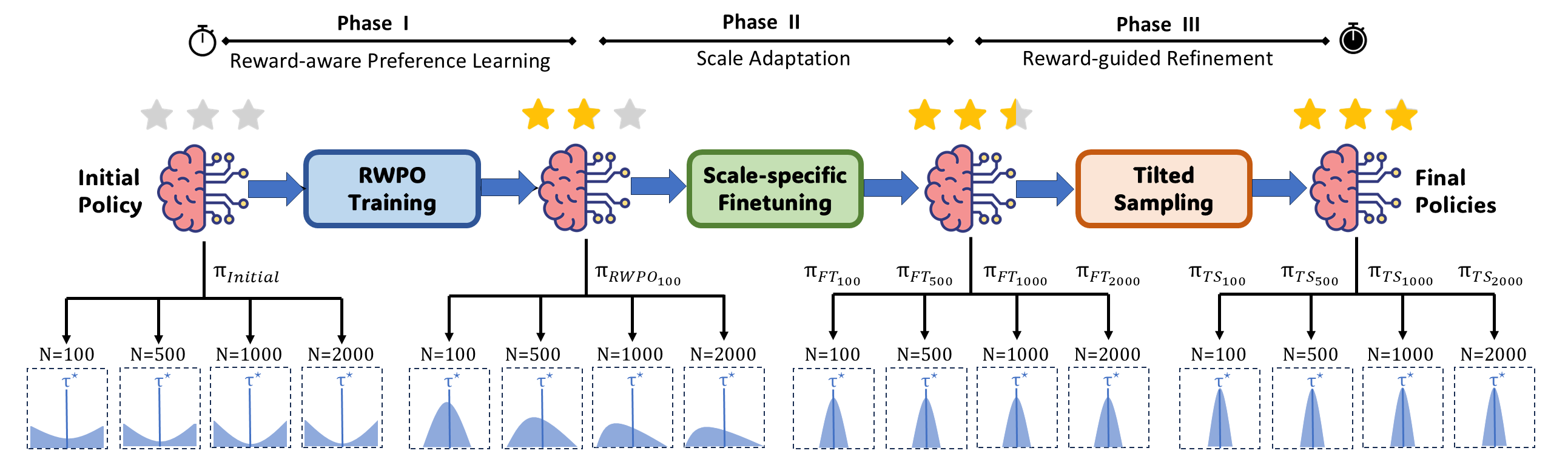}
\caption{Overview of the RWPO training and adaptation pipeline. The distributions below are conceptual visualizations of probability mass over solutions (with empirical validation provided in Appendix~\ref{app:comparison_ls_scale_ft_ts}). Starting from a random initial policy, RWPO learns reward-aware preferences, scale-specific fine-tuning adapts the policy to different scales ($N$), and tilted sampling further refines the policy toward high-quality solutions.
 Yellow stars denote solution quality, and $\tau^\star$ marks the optimal solution.}
\label{fig:rwpo_pipeline}
\end{figure*}

Policy-based Reinforcement Learning (RL) methods train such policies by directly optimizing expected rewards, but their stochastic gradient estimates can have high variance \citep{nazari2018reinforcement,hottung2020neural,kool2019attention,kwon2020pomo,wu2021learning}. Preference Optimization (PO) offers an alternative by learning from comparisons between sampled solutions \citep{rafailov2023direct} and has been applied to TSP and CVRP \citep{pan2025preference}. 
However, standard PO reduces each pair of solutions to a strictly binary preference: one solution is preferred to the other, regardless of whether their rewards differ marginally or substantially. This discards information that is directly available in combinatorial optimization, where exact objective values can be evaluated for every sampled solution. This limitation is particularly relevant to SCOPs, where different subset selections can produce widely varying solution quality. We therefore ask whether preference learning can retain its pairwise stability while exploiting the magnitude of observed reward differences.

To address these challenges, we introduce \emph{Reward-Weighted Preference Optimization} (RWPO), 
for SCOP-solving RL agents.
Rather than relying solely on binary preferences, RWPO directly embeds batch-normalized reward gaps as external weights on the pairwise log-likelihood differences.  Unlike BOPO \citep{liao2025bopo}, 
which uses best-anchored comparisons and reward-dependent scaling inside the sigmoid, RWPO retains all sampled rollout pairs and applies reward-dependent weights to the loss while maintaining a shared inverse temperature parameter.

Second, we develop a \emph{confidence-regularized all-pair} formulation that combines weighted-reward-gap preference fitting with a symmetric pairwise penalty over sampled rollout pairs.
We use all sampled rollout pairs and regularize pairwise confidence, so that small reward differences do not induce unnecessarily strong separation while larger improvements receive stronger supervision.

Third, beyond the training objective, we develop a scale adaptation pipeline that combines RWPO with scale-specific fine-tuning and tilted sampling. These components enable transfer to larger and shifted instance distributions without retraining from scratch or relying on local-search (LS) post-processing during inference, providing a scalable approach for improving generalizability  of neural solver across SCOPs.

Summarizing, our contributions are as follows: 1) We introduce RWPO for SCOPs, a preference learning objective that explicitly incorporates solution-quality differences by weighting pairwise comparisons using normalized reward gaps.  2) We develop a scalable training and adaptation pipeline combining RWPO, scale-specific fine-tuning, and tilted sampling.  Across SCOP benchmarks, the resulting neural solver achieves state-of-the-art (SOTA) solution quality and generalizes robustly to larger instance sizes.  3) We evaluate solution quality and cross-scale generalization, separating the zero-shot performance of the trained policy from the gains obtained through adaptation.  Finally, 4) we benchmark against standard preference and policy-gradient baselines, characterizing the learned policies through information-theoretic analyses.

Due to space limitations, we provide a more comprehensive discussion of related work in Appendix~\ref{app:related_work}.


\section{Preliminaries}
\label{sec:prelim}\label{sec:bg}

\subsection{Constructive Neural Solvers}

Constructive neural solvers generate a feasible solution sequentially by learning a stochastic policy over the next decision, typically trained via RL.  Following the attention-based architecture of \citet{kool2019attention}, a problem instance $x \sim \mathcal{D}$ is first encoded into node embeddings using a transformer-style encoder with multi-head self-attention, which allows each node to aggregate information from all other nodes and capture the global structure of the instance.  A decoder then constructs a solution autoregressively: at each step $t$, the policy $\pi_\theta(a_t \mid s_t, x)$ assigns probabilities to feasible actions, where the state $s_t$ represents the partial solution constructed until step $t$ and the action $a_t$ corresponds to selecting the next node, item, or operation. Feasibility constraints are enforced through masking, so invalid actions receive zero probability during decoding.  This sequential construction process can be naturally formulated as a Markov Decision Process (MDP), where the initial state is an empty solution, the transition appends the selected action to the partial solution, and the episode terminates when a complete solution is obtained or when no further feasible action can be selected.  The final reward $r(x,\tau)$ is computed from the completed trajectory $\tau=(a_1,\ldots,a_T)$, such as the negative tour length or the collected prize.  We employ this `standard' constructive policy as our base architecture, focusing specifically on how to effectively train and scale neural solvers for SCOPs.

\subsection{Reinforcement Learning for COPs}
For solving COPs,
a standard reinforcement learning (RL) objective is to maximize the expected reward
\begin{equation}
    \max_{\theta} \; 
    \mathbb{E}_{x \sim \mathcal{D}}
    \mathbb{E}_{\tau \sim \pi_\theta(\cdot \mid x)}
    [r(x,\tau)] ,
\end{equation}
whose policy-gradient estimator takes the REINFORCE form
\begin{equation}
    \nabla_\theta J(\theta)
    =
    \mathbb{E}_{x \sim \mathcal{D}, \tau \sim \pi_\theta(\cdot \mid x)}
    \left[
    r(x,\tau) \nabla_\theta \log \pi_\theta(\tau \mid x)
    \right].
\end{equation}

Although this objective is directly aligned with reward maximization, it often suffers from high gradient variance, poor trajectory-level credit assignment, leading to unstable policy updates and inefficient learning in large combinatorial decision spaces \citep{sutton2000policy,greensmith2004variance,schulman2015high,rafailov2023direct}.  These issues become more pronounced in SCOPs, where feasible solutions may differ substantially in reward and informative high-quality trajectories are difficult to sample.

\paragraph{POMO for SCOPs.}
The RL-based framework POMO \citep{kwon2020pomo} relies on the symmetry-equivalence of rollouts initialized from different starting nodes, a property inherent to sequence-invariant routing problems such as the TSP.  This symmetry assumption is less appropriate for SCOPs, as the first selected node can affect the remaining feasible region, achievable reward, and final solution quality. Nevertheless, POMO-style multi-start decoding remains an effective strategy when reinterpreted as learning a top-$k$ branch-selection policy \citep{zhang2025deep}.  Specifically, for instance $x$, the model selects a set $\mathcal{K}_\theta(x)$ of promising first-step actions greedily according to the policy distribution and performs parallel rollouts conditioned on these starts.  The resulting objective can be interpreted as maximizing the expected reward, conditioned on the trajectory's initial action belonging to the selected top-$k$ branches:
\begin{equation}
    \max_\theta 
    \mathbb{E}_{x\sim\mathcal{D}}
    \left[
    \mathbb{E}_{\tau\sim\pi_\theta(\cdot|x)}
    \left[
    r(x,\tau)
    \mid
    a_1(\tau)\in \mathcal{K}_\theta(x)
    \right]
    \right],
\end{equation}
where $a_1(\tau)$ denotes the first action of trajectory $\tau$. 
Top-$k$ branch decoding encourages exploration across promising initial actions, which is critical for SCOPs where early decisions strongly shape the remaining feasible region. However, the resulting branch heterogeneity makes reward-based policy gradients highly noisy. As detailed next, preference-based training mitigates this instability by comparing candidate branches within the same instance, yielding a much more stable learning signal.

We use POMO-style multi-start decoding to obtain multiple candidate trajectories for each instance. In SCOPs, these trajectories need not be equivalent because different initial selections can induce different feasible regions and rewards. This provides a natural within-instance comparison set for preference optimization.

\subsection{Preference Optimization for COPs}

Preference Optimization (PO) provides an alternative training paradigm for RL by replacing direct reward maximization with preference-based MLE \citep{rafailov2023direct,pan2025preference}.  
Its connection to RL can be motivated from the entropy-regularized objective. For a fixed instance $x$, consider
\begin{equation}
    \max_{\pi}
    \;
    \mathbb{E}_{\tau \sim \pi(\cdot \mid x)}
    [r(x,\tau)]
    +
    \alpha \mathcal{H}(\pi(\cdot \mid x)),
\end{equation}
where $\alpha>0$ controls the strength of entropy regularization. 
The latent reward can be reparameterized in relation to the corresponding policy $\pi^\star(\tau|x)$, which is written as
\begin{equation}
    \hat{r}(x,\tau)
    =
    \alpha \log \pi^\star(\tau \mid x)
    +
    \alpha \log Z(x),
\end{equation}
where $Z(x)$ is the partition function over all feasible trajectories.
For any pair $(\tau_i,\tau_j)$,
\begin{equation}
    \hat{r}(x,\tau_i)-\hat{r}(x,\tau_j)
    =
    \alpha
    \left[
    \log \pi^\star(\tau_i \mid x)
    -
    \log \pi^\star(\tau_j \mid x)
    \right].
\end{equation}
This allows reward comparison to be transformed into preference classification over trajectory pairs.
Given a pair of sampled trajectories $(\tau_i,\tau_j)$, we define the binary preference label as
\begin{equation}
    y_{ij} = \mathbb{I}\big(r(x,\tau_i) > r(x,\tau_j)\big).
\end{equation}
Since the true optimal policy $\pi^\star$ is unknown, PO uses the parameterized policy $\pi_\theta$ to model the preference probability. The resulting objective maximizes the expected log-likelihood of the observed preference pairs:
\begin{equation}
    \max_\theta
    \;
    \mathbb{E}_{x \sim \mathcal{D},(\tau_i,\tau_j)}
    \left[
    y_{ij}\log p_\theta(\tau_i \succ \tau_j \mid x)
    +
    y_{ji}\log p_\theta(\tau_j \succ \tau_i \mid x)
    \right].
\end{equation}
Here, we use \(y_{ji}\) rather than \(1-y_{ij}\) in standard binary cross-entropy maximization because trajectory pairs with equal rewards satisfy \(y_{ij}=y_{ji}=0\) and should be excluded from preference learning.

Under a Bradley--Terry (BT) model, which represents pairwise preference probabilities as a sigmoid of the difference between two latent utilities, the probability that $\tau_i$ is preferred to $\tau_j$ is defined as
\begin{equation}
    p_\theta(\tau_i \succ \tau_j \mid x)
    =
    \sigma
    \left(
    \alpha
    \left[
    \log \pi_\theta(\tau_i \mid x)
    -
    \log \pi_\theta(\tau_j \mid x)
    \right]
    \right),
\end{equation}
where $\sigma(\cdot)$ denotes the sigmoid function and the inverse-temperature parameter $\alpha>0$ controls the preference sharpness. 

In this way, PO converts reward-based policy learning into a pairwise MLE problem. Instead of estimating high-variance policy gradients from scalar trajectory rewards, the model is trained to assign higher likelihood to preferred trajectories than to inferior ones. Despite this stability, standard PO treats all correctly ordered pairs equally, stripping away the rich, exact reward differences that are readily available in combinatorial optimization.

For combinatorial optimization, however, the binary preference label is unnecessarily restrictive: the exact reward difference between two sampled trajectories is available at negligible additional cost. For instance, a pair whose rewards differ by \(0.01\) and a pair whose rewards differ by \(10\) receive the same preference signal under standard PO. RWPO addresses this mismatch by weighting the preference loss according to the normalized reward gap.


\section{Methodology}
\label{sec:method}

\subsection{Reward-Weighted Preference Optimization}
To overcome the limitations of standard preference learning and explicitly leverage the exact reward signals available in SCOPs, we introduce the RWPO framework.  The core idea is to dynamically scale the training signal based on the magnitude of the reward improvement, rather than treating all correctly ordered trajectory pairs equally. 
This explicitly accounts for the massive variance in subset-selection quality while retaining a shared inverse-temperature parameter across pairs.

Given an instance $x$ and a pair of sampled trajectories $(\tau_i,\tau_j)$, RWPO weights the preference log-likelihood by a reward-dependent coefficient. The RWPO objective is defined as:
\begin{equation}
    \max_{\theta}\quad
    \mathbb{E}_{x\sim\mathcal{D},(\tau_i,\tau_j)}
    \Big[
    \omega(x,\tau_i,\tau_j)\, y_{ij}
    \log p_\theta(\tau_i \succ \tau_j \mid x)
    +
    \omega(x,\tau_j,\tau_i)\, y_{ji}
    \log p_\theta(\tau_j \succ \tau_i \mid x)
    \Big].
\end{equation}
where $\omega(x,\tau_i,\tau_j)$ determines the importance of the preference pair.

We instantiate the weight using the normalized reward gap:
\begin{equation}
    \omega(x,\tau_i,\tau_j)
    :=
    1 + \lambda 
    \frac{\Delta(x,\tau_i,\tau_j)}
    {\bar{\Delta}^{(b)} + \epsilon},
    \ 
    \Delta(x,\tau_i,\tau_j)
    =
    r(x,\tau_i) - r(x,\tau_j),
    \label{eq:rwpo}
\end{equation}
where $\bar{\Delta}^{(b)}$ denotes the average reward gap within the batch, $\lambda \geq 0$ controls the reward reweighting strength, and $\epsilon$ is a small constant for numerical stability. This normalization makes the reward weight comparable across instances and reduces sensitivity to the scale of the objective values.  In contrast to standard PO, which assigns equal weight to all ordered preference pairs, RWPO amplifies comparisons with large reward gaps while relatively suppressing near ties.

\paragraph{Theoretical connection to policy gradients.} At the level of pairwise update coefficients, this weighting mechanism provides a connection between standard PO and reward-weighted policy gradients. For a fixed instance $x$, both REINFORCE and BT preference updates can be decomposed into pairwise score directions of the form:
\[
\nabla_\theta \log \pi_\theta(\tau_i|x)
-
\nabla_\theta \log \pi_\theta(\tau_j|x), 
\quad \text{where } r(x,\tau_i)-r(x,\tau_j)>0.
\]
The fundamental difference lies in the scalar coefficient attached to each direction: policy gradients scale naturally with the reward gap $r(x,\tau_i)-r(x,\tau_j)$, whereas unweighted PO assigns the same explicit constant to every preferred pair.
Under a bounded-gradient assumption, and when the pair-sampling distribution matches the current policy, the discrepancy between the RWPO preferred-pair update and the corresponding policy-gradient update is bounded by the mismatch between these effective scalar coefficients
(see Appendix~\ref{app:rwpo_gradient_matching} for the full proof). Consequently, incorporating reward-gap information provides a local motivation for making the preference-learning signal more closely resemble reward-weighted policy-gradient ascent, but it does not establish strict superiority, improved expected return, or convergence.

\paragraph{Confidence-regularized all-pair RWPO loss.} To further stabilize training, we introduce a \emph{confidence-regularized all-pair} formulation.
For each unordered rollout pair $\{\tau_i,\tau_j\}$, both preference directions are evaluated, but the direction corresponding to the better trajectory receives an additional reward-gap-dependent weight. 
Specifically, we define the bidirectional weight as:
\begin{equation}
    \tilde{\omega}_{ij}
    =
    1
    +
    \lambda
    \frac{\Delta(x,\tau_i,\tau_j)}{\bar{\Delta}^{(b)}+\epsilon}
    y_{ij} .
\end{equation}
The implemented objective is
\begin{equation}
\begin{aligned}
    \max_\theta
    \;
    \mathbb{E}_{x\sim\mathcal{D}, (\tau_i,\tau_j)}
    \Big[
    \tilde{\omega}_{ij}
    \log p_\theta(\tau_i \succ \tau_j \mid x) 
    +
    \tilde{\omega}_{ji}
    \log p_\theta(\tau_j \succ \tau_i \mid x)
    \Big].
\end{aligned}
\end{equation}
This formulation differs from the strictly ordered RWPO objective (Eq.~\ref{eq:rwpo}) by retaining the log-likelihoods for all rollout directions rather than discarding the non-preferred pairs. The constant unit baseline acts as a symmetric regularizer, providing a smoothing effect over the POMO rollout set. Simultaneously, the reward-gap-dependent term asymmetrically weights this all-pair objective toward the valid preference directions, proportionally scaling with the reward improvement. This design is particularly advantageous for POMO-style training, where multiple rollouts from the same instance provide a rich local comparison set: the symmetric all-pair component stabilizes preference learning, while the reward-weighted component prevents the loss function from treating near-ties and substantial reward improvements as equally important.

\begin{corollary}[Reward-aware preference separation]
Under the confidence-regularized all-pair RWPO objective, the induced
optimal preference margin $\log \pi_\theta(\tau_i \mid x) - \log \pi_\theta(\tau_j \mid x)$ is monotonically increasing with the reward gap.
Therefore, RWPO encourages stronger separation between trajectories with
larger reward improvements while avoiding excessive separation for
near-tie solutions.
\end{corollary}
The gradient of the two objectives are provided in Appendix~\ref{app:rwpo_gradient}.

\subsection{Training and Adaptation Pipeline for SCOPs}
\label{sec:training_adaptation_pipeline}

Building on the RWPO objective, we design a complete training and adaptation pipeline for SCOPs to enhance both in-distribution performance and zero-shot performance on much larger problem sizes. 
While a trained RWPO policy generalizes well, applying it to significantly larger target scales requires distribution-specific adaptation to account for expanding decision spaces. To achieve this efficiently, our pipeline (Algorithm~\ref{alg:rwpo_pipeline}) introduces two lightweight post-training stages: low-learning-rate scale-specific fine-tuning and tilted sampling adaptation. Crucially, this design explicitly bypasses computationally expensive local-search (LS) post-processing or LS-guided fine-tuning \citep{pan2025preference}. Instead, our pipeline shifts the computational burden entirely to a one-time, offline adaptation phase, ensuring inference on new instances remains fast and search-free.

\paragraph{Scale-specific fine-tuning.}
A straightforward yet effective adaptation strategy is continued training with a reduced learning rate \citep{mosbach2020stability,lin2026sft}. 
Empirically, it acts as conservative policy refinement: once RWPO has established a strong base policy, low-learning-rate adaptation further concentrates probability mass on high-quality actions without substantially altering the learned solution structure. While full backpropagation through the Transformer scales with instance size, this is strictly a one-time, offline sunk cost for a target distribution; a direct contrast to LS heuristics, which impose a substantial recurring computational penalty on every single instance during inference. To keep this offline adaptation efficient on massive instances, we purposefully employ small batch sizes, reduced multi-start sizes, and aggressive early stopping (e.g., 30 epochs).

\paragraph{Tilted sampling adaptation.}
To further concentrate probability mass on high-quality actions without destabilizing the underlying solution structure, we introduce a \emph{tilted sampling} strategy for target-scale adaptation. 
Rather than relying on computationally expensive LS oracles to generate supervisory targets \citep{pan2025preference}, this method efficiently updates the policy using self-generated rollouts.
Building on the KL-regularized policy optimization framework \citep{mittal2026beyond}, we define a reward-anchored sharpening objective that re-weights each rollout according to both its task reward and its likelihood under a frozen RWPO reference policy. Maximizing the weighted log-likelihood of these rollouts effectively performs weighted behavioral cloning, allowing us to exploit target-scale rewards while the frozen reference acts as an implicit KL-regularizer to preserve structural priors. 

For each trajectory $\tau$, we compute
\begin{equation}
\varsigma(x,\tau)
=
\lambda_r z(r(x, \tau))
+
\lambda_p z\!\left(\log \pi_{\mathrm{RWPO}}(\tau\mid x)\right),
\end{equation}
where $z(\cdot)$ denotes within-instance normalization, $\lambda_r$ controls the contribution of task reward, and $\lambda_p$ controls the strength of the base-policy likelihood anchor. Given a candidate set $\mathcal{C}(x)=\{\tau_1,\ldots,\tau_K\}$, the scores are converted into normalized weights using a softmax temperature $T_w$:
\begin{equation}
w_k
=
\frac{
\exp(\varsigma(x,\tau_k)/T_w)
}{
\sum_{j=1}^{K}\exp(\varsigma(x,\tau_j)/T_w)
}.
\end{equation}
The policy is then adapted by weighted behavioural cloning:
\begin{equation}
\mathcal{L}_{\mathrm{tilt}}(\theta)
=
-
\mathbb{E}_{x}
\left[
\sum_{k=1}^{K}
w_k
\log \pi_\theta(\tau_k \mid x)
\right].
\end{equation}

This objective can be interpreted as an approximation of a reward-tilted distribution specific for COPs.  High-reward trajectories receive larger weights, while the likelihood anchor prevents the adapted policy from moving too far away from the original RWPO policy.  Therefore, tilted sampling performs reward-guided but conservative probability redistribution: it is more reward-aware than pure distribution sharpening, yet more stable than unconstrained reward-only fine-tuning.  
More details are provided in Appendix~\ref{app:tilted_sampling}.


\begin{algorithm}[t]
\caption{RWPO Training and Scale Adaptation}
\label{alg:rwpo_pipeline}
\begin{algorithmic}[1]
\STATE \textbf{Input:} Training distribution $\mathcal{D}_{\mathrm{train}}$;
target-scale distribution $\mathcal{D}_{\mathrm{adapt}}$;
policy $\pi_\theta$;
RWPO parameters $(\alpha,\beta,\lambda)$;
tilting parameters $(\lambda_r,\lambda_p,T_w)$

\STATE \textbf{Phase I: Base RWPO training}
\FOR{each batch $x_b \sim \mathcal{D}_{\mathrm{train}}$}
    \STATE Sample $M$ multi-start trajectories
    $\{\tau_{b,m}\}_{m=1}^{M} \sim \pi_\theta(\cdot \mid x_b)$
    \STATE Evaluate rewards $r_{b,m}$ and compute normalized trajectory log-likelihoods
    \[
    \ell_{b,m}
    =
    T_{b,m}^{-\beta}
    \sum_{t=1}^{T_{b,m}}
    \log \pi_\theta(a_t \mid s_t,x_b).
    \]
    \FOR{each ordered pair $(i,j)$}
        \STATE \[\Delta_{ij} \leftarrow r_{b,i}-r_{b,j}, 
        \quad
        \tilde{\omega}_{ij}
        \leftarrow
        1+\lambda 
        \frac{[\Delta_{ij}]_+}{\bar{\Delta}^{(b)}+\epsilon} ,
        \]
        where $\bar{\Delta}^{(b)}$ is the mean absolute
    reward gap in the batch.
    \ENDFOR
    \STATE Compute the confidence-regularized all-pair RWPO loss\\
    \[
    \mathcal{L}_{\mathrm{RWPO}}
    =-\frac{1}{|\mathcal{P}_b|}
    \sum_{(i,j)\in\mathcal{P}_b}
    \Big[
    \tilde{\omega}_{ij}
    \log \sigma\!\big(
        \alpha(\ell_{b,i}-\ell_{b,j})
    \big)
    \Big].
    \]
    \STATE Update $\theta \leftarrow
    \theta-\eta\nabla_\theta\mathcal{L}_{\mathrm{RWPO}}$.
\ENDFOR

\STATE \textbf{Phase II: Scale-specific fine-tuning}
\STATE Continue Phase-I RWPO training on
$\mathcal{D}_{\mathrm{adapt}}$ using
$\eta_{\mathrm{FT}} \ll \eta$.
\STATE Set the frozen reference policy
$\pi_{\mathrm{ref}} \leftarrow \pi_\theta$.

\STATE \textbf{Phase III: Tilted-sampling adaptation}
\STATE Freeze the pre-tilt policy:
$\pi_{\mathrm{ref}} \leftarrow \pi_\theta$.

\FOR{each batch $\{x_b\}_{b=1}^{B}
      \sim \mathcal{D}_{\mathrm{adapt}}$}
    \STATE Sample $K$ trajectories
    $\mathcal{C}(x_b)
    =\{\tau_{b,k}\}_{k=1}^{K}
    \sim \pi_\theta(\cdot\mid x_b)$.

    \STATE Standardize rewards and reference
    log-likelihood: \\
    \[
    \hat r_{b,k}=z_b(r(x_b,\tau_{b,k})),
    \qquad
    \hat q_{b,k}=z_b(\log\pi_{\mathrm{ref}}
    (\tau_{b,k}\mid x_b)).
    \]

    \STATE Compute tilted scores and weights:\\
    \[
    \varsigma_{b,k}
    =
    \lambda_r\hat r_{b,k}
    +
    \lambda_p\hat q_{b,k},
    \qquad
    w_{b,k}
    =
    \frac{
    \exp(\varsigma_{b,k}/T_w)
    }{
    \sum_{j=1}^{K}
    \exp(\varsigma_{b,j}/T_w)
    }.
    \]

    \STATE Compute the tilted-sampling loss \\
    \[
    \mathcal{L}_{\mathrm{tilt}}
    =
    -\frac{1}{B}
    \sum_{b=1}^{B}\sum_{k=1}^{K}
    \operatorname{sg}(w_{b,k})
    \log\pi_\theta(\tau_{b,k}\mid x_b),
    \] \\
    where $\operatorname{sg}(\cdot)$ stops gradients through
    the sampling weights.

    \STATE Update
    $\theta\leftarrow
    \theta-\eta_{\mathrm{TS}}
    \nabla_\theta\mathcal{L}_{\mathrm{tilt}}$,
    
\ENDFOR

\end{algorithmic}
\end{algorithm}


\section{Experiments}
\label{sec:results}

We
present the main results of our experiments, in which we observe superior performance of the proposed RWPO for SCOPs compared to approaches in the literature.  We answer the following questions: 
1.~\textit{How does RWPO compare to existing algorithms on standard benchmarks of the 
OP, PCTSP, SPCTSP, and KP?}
~2.~\textit{How does the out-of-distribution (OOD) generalization of the RWPO compare to existing neural combinatorial optimization (NCO) algorithms?}
and
3.~\textit{What structural information is encoded within in the learned representation, and does that explain generalization?}

\paragraph{Evaluation Protocol.}
To ensure a fair comparison with prior studies, we follow standard benchmark protocols for both mid- and large-scale instances. For mid-scale instances ($n=100$), we use test sets sampled from the same uniform distributions as in \citet{kool2019attention} and \citet{kwon2020pomo}. For large-scale instances ($n \in \{500, 1000, 2000\}$), we adopt the test sets from \citet{zheng2024udc}. To further evaluate generalization, we conduct zero-shot testing on instances generated with different random seeds and on the standard OPLib benchmark dataset \cite{joshi2019efficient}. We evaluate performance using the average accumulated reward or cost; for minimization problems, the optimality gap is computed as $\mathrm{Gap}(\%) = (\mathrm{Obj}_{\mathrm{alg}}-\mathrm{Obj}_{\mathrm{ref}}) / \mathrm{Obj}_{\mathrm{ref}} \times 100$. Finally, to reflect practical usage, solver runtimes are measured out-of-the-box—evaluating non-neural methods on CPU and neural methods on GPU—using an Intel i5-14600KF CPU and an NVIDIA RTX 5060 Ti (16GB) GPU.

\paragraph{Implementation.}
For same-distribution evaluation, we apply continued low-learning-rate training followed by tilted sampling (TS). For large-scale or distribution-shifted instances, we employ scale-specific fine-tuning (40 epochs for $N=500$; 30 epochs for $N \in \{1000, 2000\}$), paired with TS for 100 epochs.
Regarding hyperparameters, we follow \citet{pan2025preference} by setting the inverse temperature to $\alpha=0.05$ for both RWPO and PO. The reward-weighting coefficient is set to $\lambda=0.5$ for the OP and KP, and $\lambda=1.0$ for the remaining problems. For the TS stage, we use $\lambda_r=1$, $\lambda_s=0.25$, and $T_w=1$. During training, our multi-start rollout size is $M=\min\{N,100\}$ for the KP and $M=\min\{N/2,100\}$ for all other problems, whereas for inference, we report multi-start decoding performance using $M=N$ across all SCOPs. Lastly, we apply a \emph{length-normalized log-probability} adjustment (see Appendix~\ref{app:length_normalization} for details), setting the normalization coefficient to $\beta=0.3$ for the OP and KP, and $\beta=0.1$ for the remaining SCOPs.

\paragraph{Baselines.} This paper compares the proposed RWPO (confidence-regularized all pair version) to advanced (meta-)heuristic algorithms and existing neural constructive solvers for COPs, namely non-neural solvers (LKH \citep{helsgaun2017extension}, Compass \citep{joshi2019efficient}, OR-Tools \citep{perron2019or}), neural constructive solvers (AM \citep{kool2019attention}, POMO \citep{kwon2020pomo}, MDAM \citep{xin2021multi}, Sym-NCO \citep{kim2022sym}, PO \citep{pan2025preference}) and large-scale neural solvers (BQ \citep{drakulic2023bq}, GLOP \citep{ye2024glop},  UDC \citep{zheng2024udc}). The implementation details and settings of all methods compared are described in Appendix \ref{app:baselines}.


\subsection{Comparison with Existing Algorithms on Standard Benchmarks}

\begin{table*}[t]
\centering
\caption{Comparison on OP, PCTSP, SPCTSP, and KP instances across different scales.}
\setlength{\tabcolsep}{2.6pt}
\renewcommand{\arraystretch}{1.02}
\small
\begin{tabular}{llccc | l | ccc ccc ccc}
\toprule
 & \multirow{2}{*}{Methods} 
 & \multicolumn{3}{c|}{$N=100$} 
 &
 & \multicolumn{3}{c}{$N=500$} 
 & \multicolumn{3}{c}{$N=1{,}000$} 
 & \multicolumn{3}{c}{$N=2{,}000$} \\
\cmidrule(lr){3-5} 
\cmidrule(lr){6-6} 
\cmidrule(lr){7-9} 
\cmidrule(lr){10-12} 
\cmidrule(lr){13-15}
 &  & Obj. & Gap & Time
 &  & Obj. & Gap & Time 
    & Obj. & Gap & Time 
    & Obj. & Gap & Time \\
\midrule
\midrule

\multirow{11}{*}{\rotatebox{90}{\textbf{OP}}}
& Compass              & 33.133 & 0.00\% & 12m & & 81.123 & 0.00\% & 56s  & 115.690 & 0.00\% & 4.1m & 166.749 & 0.00\% & 1.3m \\
\cmidrule(lr){2-15}
& AM-bs1024          & 32.686 & 1.35\% & 20m &  \multirow{8}{*}{\rotatebox{90}{OOD}} & 68.592 & 15.45\% & 2.5m & 79.655  & 31.15\% & 7m    & 95.484  & 42.74\% & 3.4m \\
& MDAM-bs50          & 32.975 & 0.48\% & 22m & & 65.842 & 18.84\% & 3.5m  & 81.114  & 29.89\% & 33m   & 102.983 & 38.24\% & 12m \\
& Sym-NCO            & 33.039 & 0.28\% & 6m  & & 73.225 & 9.74\% & 52s  & 95.358  & 17.57\% & 2.7m  & 122.832 & 26.34\% & 40s \\
& BQ           & 33.074 & 0.18\% & 1.5m & & 78.140 & 3.68\% & 48s  & 107.446  & 7.13\% & 4.3m   & 144.007 & 13.64\% & 3.3m \\
& POMO               & 33.150 & -0.05\% & 24s & & 65.089 & 19.77\% & 7s  & 79.329  & 31.43\% & 37s  & 99.403  & 40.39\% & 18s \\
& PO                 & 33.259 & -0.38\% & 24s & & 35.661 & 56.04\% & 6s  & 41.233 & 64.36\% & 33s  & 49.028  & 70.60\% & 16s \\
& PO+LS            & 33.281 & -0.45\% & 25s &  & 33.701 & 58.46\% & 6s & 38.010 & 67.14\% & 33s  & 44.452  & 73.34\% & 15s \\
& RWPO               & 33.285 & -0.46\% & 25s & & 74.879 & 7.70\% & 7s 
                     & 99.577 & 13.93\% & 46s
                     & 134.607 & 19.28\% & 23s \\
& RWPO+LS          & \textbf{33.302} & -0.51\% & 25s 
&
                     & 75.763 & 6.61\% & 8s 
                     & 99.907 & 13.64\% & 45s
                     & 134.408 & 19.40\% & 24s \\
\cmidrule(lr){2-15}
& UDC                & --    & --      & -- 
& \multirow{2}{*}{\rotatebox{90}{ID}}
                      & 79.248 & 2.31\% & 57s 
                      & 113.173 & 2.18\% & 19m
                      & 145.001 & 13.04\% & 25s \\
& RWPO+TS          & \cellcolor{gray!20}\textbf{33.302} & \cellcolor{gray!20}-0.51\% & 25s
                    &
                      & \cellcolor{gray!20}\textbf{83.493} & \cellcolor{gray!20}-2.92\% & 8s 
                      & \cellcolor{gray!20}\textbf{121.273} & \cellcolor{gray!20}-4.83\% & 56s
                      & \cellcolor{gray!20}\textbf{175.372} & \cellcolor{gray!20}-5.17\% & 29s \\
\midrule
\midrule

\multirow{9}{*}{\rotatebox{90}{\textbf{PCTSP}}}
& OR-Tools$^*$     
& 6.070 & 1.61\% & 5h 
& 
& 14.000 & 7.63\% & 16h   
& 20.600 & 14.53\% & 16h   
& --    & -- & --   \\
\cmidrule(lr){2-15}

& AM-bs1024          
& 6.094 & 2.01\% & 22m 
& \multirow{5}{*}{\rotatebox{90}{OOD}}
& 19.127 & 47.04\% & 8m   
& 30.812 & 71.30\% & 35m   
& 47.420 & 94.31\% & 23m \\

& MDAM-bs50          
& 6.063 & 1.49\% & 28m 
& 
& 14.829 & 14.00\% & 9m  
& 22.510 & 25.15\% & 1.4h   
& 33.015 & 35.29\% & 1.2h \\

& Sym-NCO            
& 6.054 & 1.34\% & 6m 
& 
& 19.433 & 49.39\% & 1.8m  
& 29.443 & 63.69\% & 8m    
& 45.329 & 85.74\% & 3m \\

& PO                 
& 5.982 & 0.13\% & 22s 
& 
& 16.132 & 24.02\% & 22s  
& 26.926 & 49.70\% & 4m    
& 41.755 & 71.10\% & 4m \\

& RWPO               
& 5.977 & 0.05\% & 19s 
& 
& 15.758 & 21.14\% & 22s  
& 24.487 & 36.14\% & 4m    
& 37.934 & 55.44\% & 3m \\

\cmidrule(lr){2-15}

& GLOP               
& -- & -- & -- 
& \multirow{3}{*}{\rotatebox{90}{ID}}
& 14.432 & 10.95\% & 1m  
& 19.869 & 10.46\% & 1.8m  
& 28.418 & 16.45\% & 36s \\

& UDC                 
& -- & -- & -- 
&
& 13.158 & 1.15\% & 1.4m
& 18.283 & 1.65\% & 3.6m
& 25.045 & 2.63\% & 56s \\

& RWPO+TS             
& \cellcolor{gray!20}\textbf{5.974} & \cellcolor{gray!20}0.00\% & 21s 
&
& \cellcolor{gray!20}\textbf{13.008} 
& \cellcolor{gray!20}0.00\% & 14s
& \cellcolor{gray!20}\textbf{17.987} 
& \cellcolor{gray!20}0.00\% & 2m
& \cellcolor{gray!20}\textbf{24.404} 
& \cellcolor{gray!20}0.00\% & 1.5m \\

\midrule
\midrule

\multirow{6}{*}{\rotatebox{90}{\textbf{SPCTSP}}}

& AM-bs1024          
& 6.115 & 2.36\% & 23m 
& \multirow{4}{*}{\rotatebox{90}{OOD}}
& 17.285 & 32.96\% & 6.5m  
& 26.646 & 48.35\% & 29m    
& 37.421 & 53.59\% & 20m \\

& MDAM-bs50               
& 6.071 & 1.62\% & 30m 
&
& 14.826 & 14.05\% & 8.3m  
& 22.135 & 23.24\% & 1.3h   
& 32.102 & 31.75\% & 1.2h \\

& PO                 
& 5.983 & 0.15\% & 25s 
&
& 16.562 & 27.40\% & 26s  
& 28.732 & 59.97\% & 3m    
& 46.397 & 90.42\% & 2.8m \\

& RWPO               
& 5.977 & 0.05\% & 19s 
&
& 16.601 & 27.70\% & 26s  
& 25.567 & 42.35\% & 3m    
& 37.538 & 54.07\% & 2.7m \\

\cmidrule(lr){2-15}

& UDC                 
& -- & -- & -- 
& \multirow{2}{*}{\rotatebox{90}{ID}}
& 13.233 & 1.79\% & 1.5m
& 18.523 & 3.13\% & 3.7m
& 25.551 & 4.87\% & 54s \\

& RWPO+TS             
& \cellcolor{gray!20}\textbf{5.974} & \cellcolor{gray!20}0.00\% & 20s 
&
& \cellcolor{gray!20}\textbf{13.000} 
& \cellcolor{gray!20}0.00\% & 16s
& \cellcolor{gray!20}\textbf{17.961} 
& \cellcolor{gray!20}0.00\% & 1.7m
& \cellcolor{gray!20}\textbf{24.365} 
& \cellcolor{gray!20}0.00\% & 1.5m \\
\midrule
\midrule
 &
 & \multicolumn{3}{c|}{$N=200$} 
 &
 & \multicolumn{3}{c}{$N=500$} 
 & \multicolumn{3}{c}{$N=1{,}000$} 
 & \multicolumn{3}{c}{$N=2{,}000$} \\
 \cmidrule(lr){3-5} 
\cmidrule(lr){6-6} 
\cmidrule(lr){7-9} 
\cmidrule(lr){10-12} 
\cmidrule(lr){13-15}
\multirow{6}{*}{\rotatebox{90}{\textbf{KP}}}
& OR-Tools           &\textbf{ 57.640} & 0.00\% & 6.5m & & \textbf{128.369} & 0.00\% & 25s & \textbf{258.254} & 0.00\% & 1.8m & \textbf{518.397} & 0.00\% & 52s\\
\cmidrule(lr){2-15}
& BQ                 & 57.578 & 0.11\% & 7m 
& \multirow{4}{*}{\rotatebox{90}{OOD}} & 128.279 & 0.07\% & 1.1m & 256.409 & 0.71\% & 7.3m & 517.124 & 0.25\% & 6.1m \\
& POMO               & 57.623 & 0.03\% & 49s 
& & 128.316 & 0.04\% & 8s & 254.537 & 1.44\% & 53s & 471.631 & 9.02\% & 46s \\
& PO                 & 57.629 & 0.02\% & 1.3m & & 108.750 & 15.28\% & 10s & 147.525 & 42.88\% & 41s &  230.181 & 55.60\% & 34s \\
& RWPO               & 57.629 & 0.02\% & 1.3m & & 126.541 & 1.42\% & 12s & 253.548 & 1.82\% & 1.5m & 503.235 & 2.92\% & 1.4m \\
\cmidrule(lr){2-15}
& UDC                & -- & -- & -- 
& \multirow{2}{*}{\rotatebox{90}{ID}} & 128.356 & 0.01\% & 47s & 258.224 & 0.01\% & 1.9m & 518.226 & 0.03\% & 30s \\
& RWPO               & \cellcolor{gray!20}57.629 & \cellcolor{gray!20}0.02\% & 1.3m & & \cellcolor{gray!20}128.364 & \cellcolor{gray!20}0.00\% & 12s & \cellcolor{gray!20}258.247 & \cellcolor{gray!20}0.00\% & 1.3m & \cellcolor{gray!20}518.284 & \cellcolor{gray!20}0.02\% & 1.2m \\
\bottomrule
\end{tabular}
\footnotesize{\\For each problem, the upper block reports out-of-distribution (OOD) results, evaluating the zero-shot scale generalization of the base models, while the lower block reports in-distribution (ID) results after scale-specific adaptation. For AM, MDAM, Sym-NCO, BQ, UDC, and POMO on KP, we use the official pre-trained models from the original papers.}
\label{tab:selective_cop_results}
\end{table*}

Tab.~\ref{tab:selective_cop_results} demonstrates the effectiveness of our framework across both zero-shot scale generalization (OOD) and target-scale adaptation (ID) (for CVRP results, see Appendix~\ref{app:CVRP}). Without any scale-specific fine-tuning, the base RWPO model exhibits robust OOD generalization, significantly mitigating the severe performance degradation seen in standard PO and prior constructive methods at $N=1000$ and $N=2000$. Although this zero-shot performance trails BQ by a narrow margin on OP and KP, our approach successfully surpasses BQ in the ID setting. When combined with tilted sampling for target-scale adaptation (RWPO+TS), our approach establishes a new SOTA. It consistently outperforms the highly optimized UDC baseline across all problems, regularly matching or exceeding the performance of the rigorous OR-Tools and Compass baselines while requiring only seconds to minutes of computational time, including a 4.92\% improvement at OP2000 on Compass. Furthermore, RWPO+TS is highly computationally efficient: on the OP, it achieves performance comparable to the RWPO+LS proposed in \citep{pan2025preference}, but substantially reduces the required training time.

\subsection{Analysis of Generalization}
To evaluate our framework on standardized routing benchmarks, Tab.~\ref{tab:ea4op_gap} presents zero-shot generalization results evaluated exclusively on the Euclidean 2D instances from OPLib Generation 3. While the absolute optimality gaps reported here are larger than those on our synthetic test sets, this is expected: many OPLib instances feature severe spatial distribution shifts—such as heavily clustered or highly irregular node coordinates—that deviate drastically from our training distribution. Despite these challenging shifts, RWPO demonstrates remarkable robustness. For instance, RWPO$_{100}$ significantly outperforms standard PO$_{100}$ across all regimes, cutting the optimality gap by more than half on the massive (2000, 5934) instances (36.24\% vs. 75.99\%). The results also reveal a natural cross-scale trade-off: policies trained on smaller graphs (e.g., RWPO$_{50}$) excel on smaller instances, whereas models adapted on massive graphs (RWPO$_{2000}$) generalize much better to the largest regimes. To bridge this trade-off, we introduce RWPO$_{\mathrm{cross}}$ 
, a decoding strategy that processes each instance using all scale-specific RWPO models and simply selects the final valid trajectory with the highest reward. By combining the complementary structural priors learned across different scales, the cross-scale decoding strategy strictly dominates any individual model, achieving the lowest gaps across all problem sizes (e.g., 4.23\% on the smallest scale and 22.82\% on the largest) with only a modest increase in overall inference time.

\begin{table}[H]
    \centering
    \small
    \caption{Average gap to the Compass results.}
    \label{tab:ea4op_gap}
    \begin{tabular}{lrrrr}
        \toprule
        Method
        & $(50,200]$
        & $(200,400]$
        & $(400,2000]$
        & $(2000,5934]$ \\
        \midrule
        POMO$_{100}$ & 6.03\% & 10.98\% & 41.04\% & 61.54\% \\
        PO$_{100}$ & 13.31\% & 41.47\% & 62.79\% & 75.99\% \\
        RWPO$_{50}$
        & \textbf{5.71\%} & 15.89\% & 34.48\% & 46.74\% \\
        RWPO$_{100}$
        & 6.35\% & \textbf{10.77\%} & 26.71\% & 35.72\% \\
        RWPO$_{500}$
        & 14.68\% & 16.72\% & 25.96\% & 28.47\% \\
        RWPO$_{1000}$
        & 13.82\% & 17.02\% & \textbf{22.92\%} & \textbf{26.72\%} \\
        RWPO$_{2000}$
        & 15.61\% & 17.76\% & 23.81\% & 27.29\% \\
        \midrule
        RWPO$_{\mathrm{cross}}$
        & \textbf{3.20\%} & \textbf{6.07\%}
        & \textbf{19.11\%} & \textbf{22.78\%} \\
        \bottomrule
    \end{tabular}
\end{table}

\begin{figure*}[t]
    \centering

    \begin{subfigure}[t]{0.32\textwidth}
        \centering
        \includegraphics[width=\linewidth]{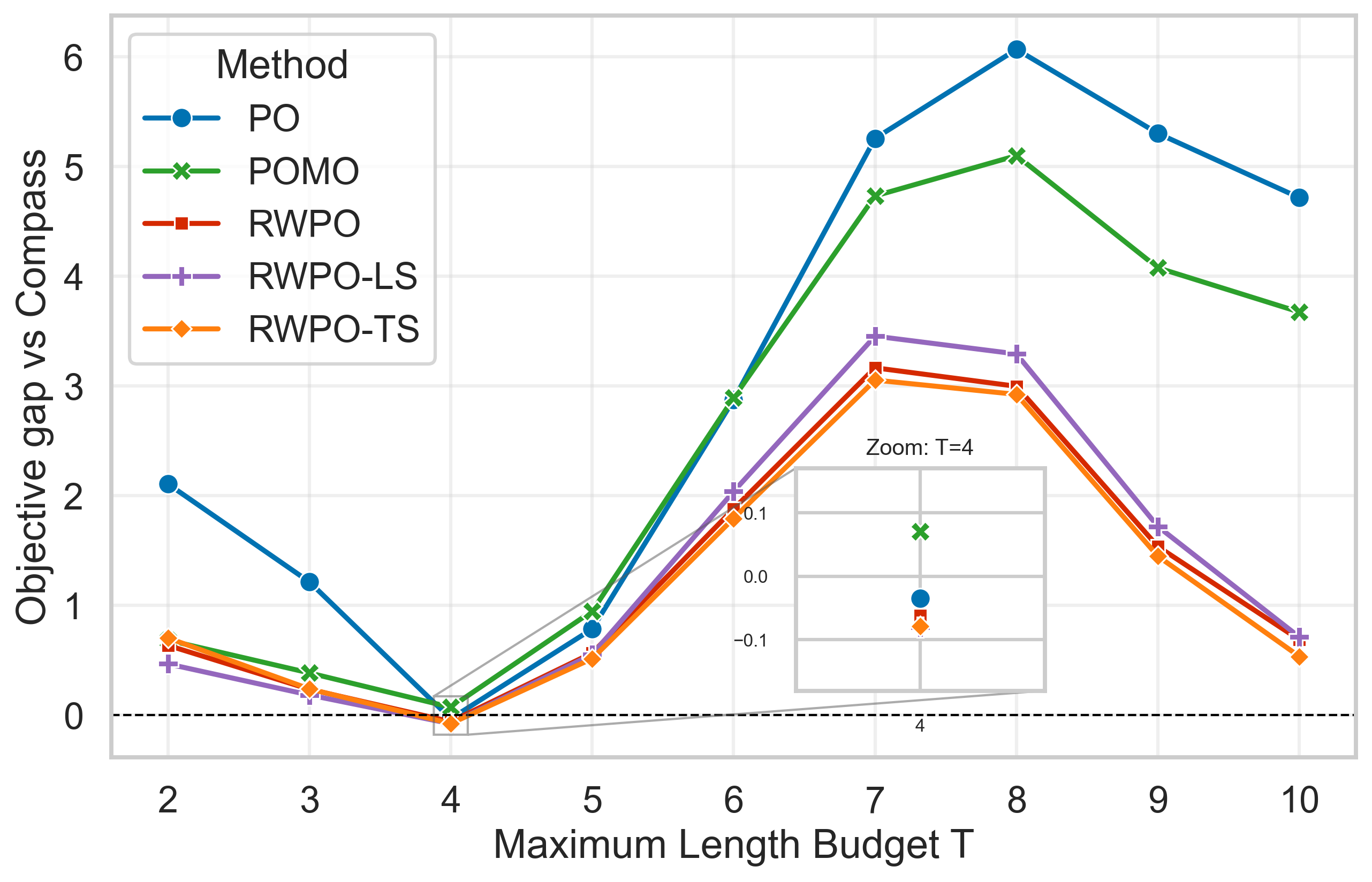}
        \vspace*{-5mm}
        \caption{$N=100$}
        \label{fig:op_gap_n100}
    \end{subfigure}
    \hfill
    \begin{subfigure}[t]{0.32\textwidth}
        \centering
        \includegraphics[width=\linewidth]{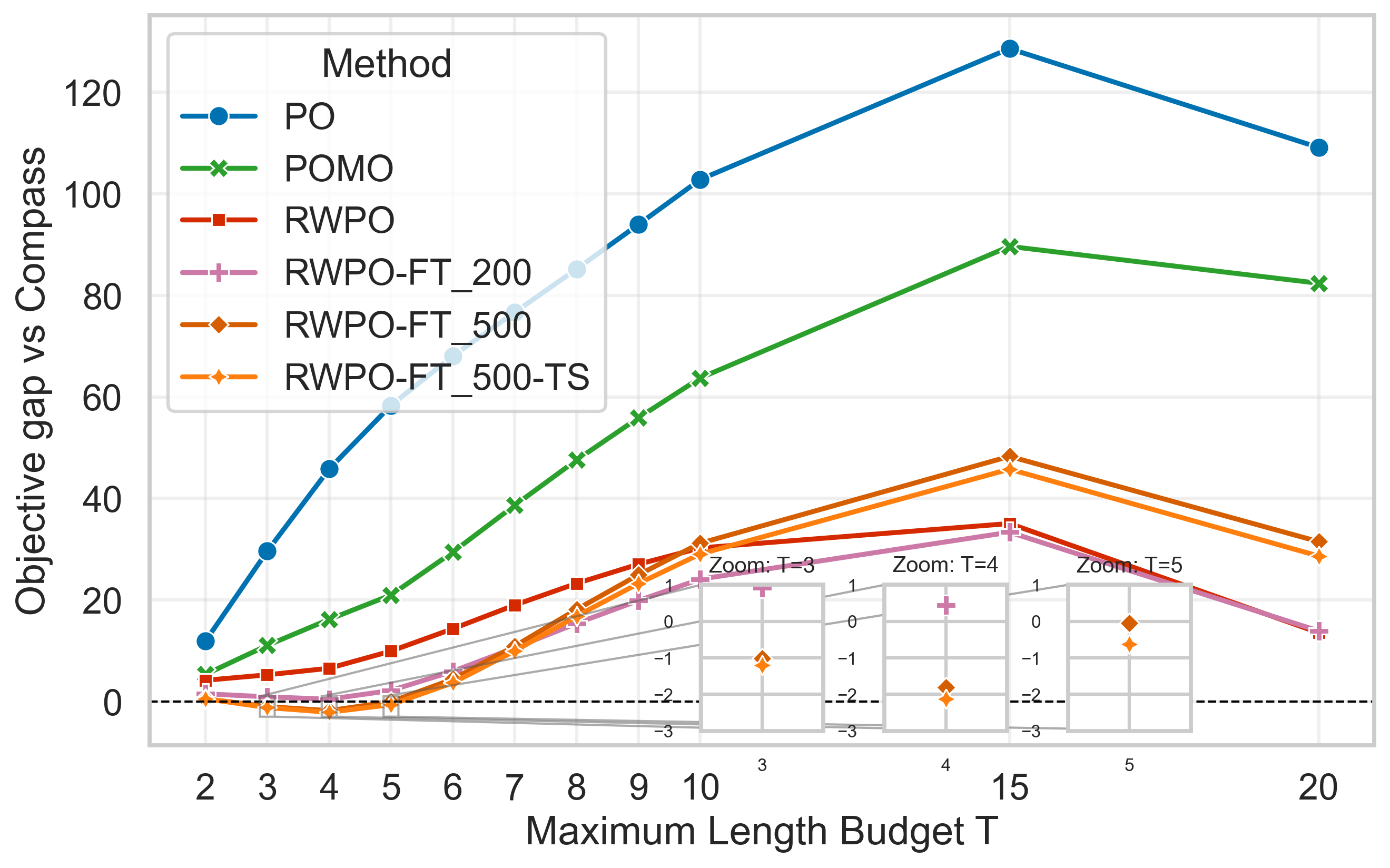}
        \vspace*{-5mm}
        \caption{$N=500$}
        \label{fig:op_gap_n500}
    \end{subfigure}
    \hfill
    \begin{subfigure}[t]{0.32\textwidth}
        \centering
        \includegraphics[width=\linewidth]{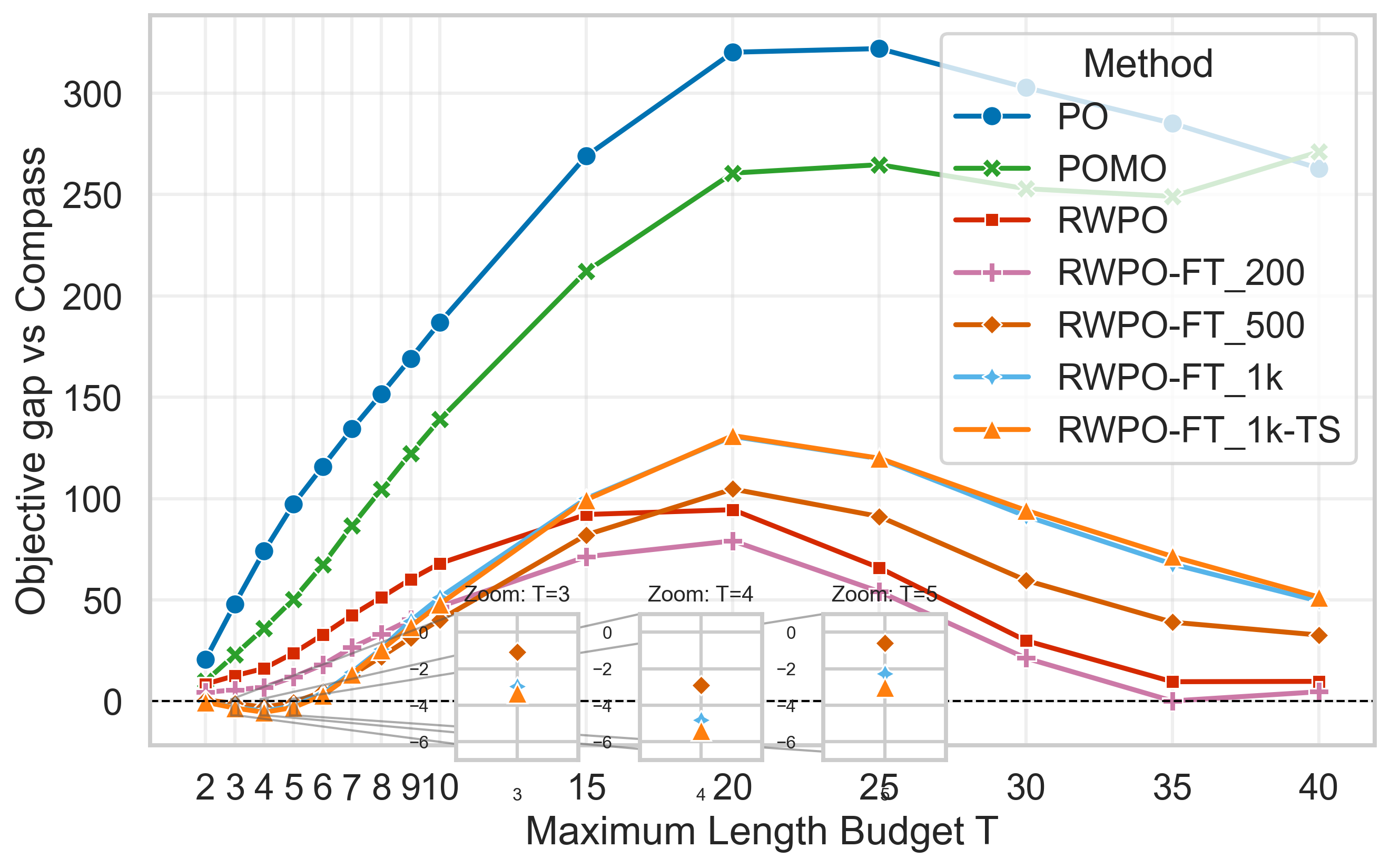}
        \vspace*{-5mm}
        \caption{$N=1{,}000$}
        \label{fig:op_gap_n1000}
    \end{subfigure}
    \caption{
    Objective gap against Compass on OP instances under different travel budgets $T$ and problem sizes.
    }
    \label{fig:op_objective_gap_compass}
\end{figure*}

Fig.~\ref{fig:op_objective_gap_compass} demonstrates that RWPO improves both average solution quality and instance-level robustness. Across all evaluated problem sizes and travel budgets, RWPO consistently achieves smaller objective gaps against Compass than both standard PO and POMO, with its advantage becoming particularly pronounced on larger instances. PO degrades severely at scale, suggesting that binary preference supervision struggles when the decision space expands and the selected subset becomes sparse. Conversely, while POMO outperforms PO in several large-scale settings—indicating the necessity of explicit reward signals—it remains limited by high-variance training and inefficient exploration. 
As the travel budget increases and the OP shifts from a sparse selection problem to a dense routing problem (similar to the TSP), RWPO maintains its strong performance, proving it captures fundamental routing structures rather than just exploiting sparse reward signals. These results confirm that preserving reward-gap information makes preference learning vastly more robust to weak comparisons and distribution shifts.

\subsection{Entropy Analysis}\label{sec:entropy_representation}

To better understand why RWPO generalizes on SCOPs, we analyze policy-level uncertainty and decoder-query representation diversity using two complementary measures. \emph{Policy entropy} is the Shannon entropy of the decoder's probability distribution over feasible next actions, measuring how concentrated its action probabilities are at each decoding step. \emph{Decoder-query soft entropy} applies the soft-entropy procedure used by \citet{conklin2026learning} to decoder queries collected across instances and rollouts at each step. Specifically, queries are $L_2$ normalized and softly assigned to randomly sampled reference directions on the unit sphere; we then compute the Shannon  entropy of the aggregated assignment distribution. This metric serves as a entropy-based proxy for angular representation diversity. 
Since query--key alignment determines relative decoder candidate scores, this measure helps identify when the query component of the decoder becomes more concentrated across different partial solutions (see Appendix~\ref{app:soft_entropy} for more details). We target decoder queries because they act as the information bottleneck for next-action scoring, uniquely aggregating critical state context: the global graph, node embeddings, and feasibility constraints. Together, the two measures characterize how action uncertainty and query-distribution diversity evolve during solution construction as feasibility constraints change. We use OP as the primary testbed because it combines reward-based subset selection with budget-constrained routing, allowing us to examine these dynamics as candidate-set size increases.

\begin{figure*}[t]
    \centering

    \begin{subfigure}[t]{0.24\textwidth}
        \centering
        \includegraphics[width=\linewidth]{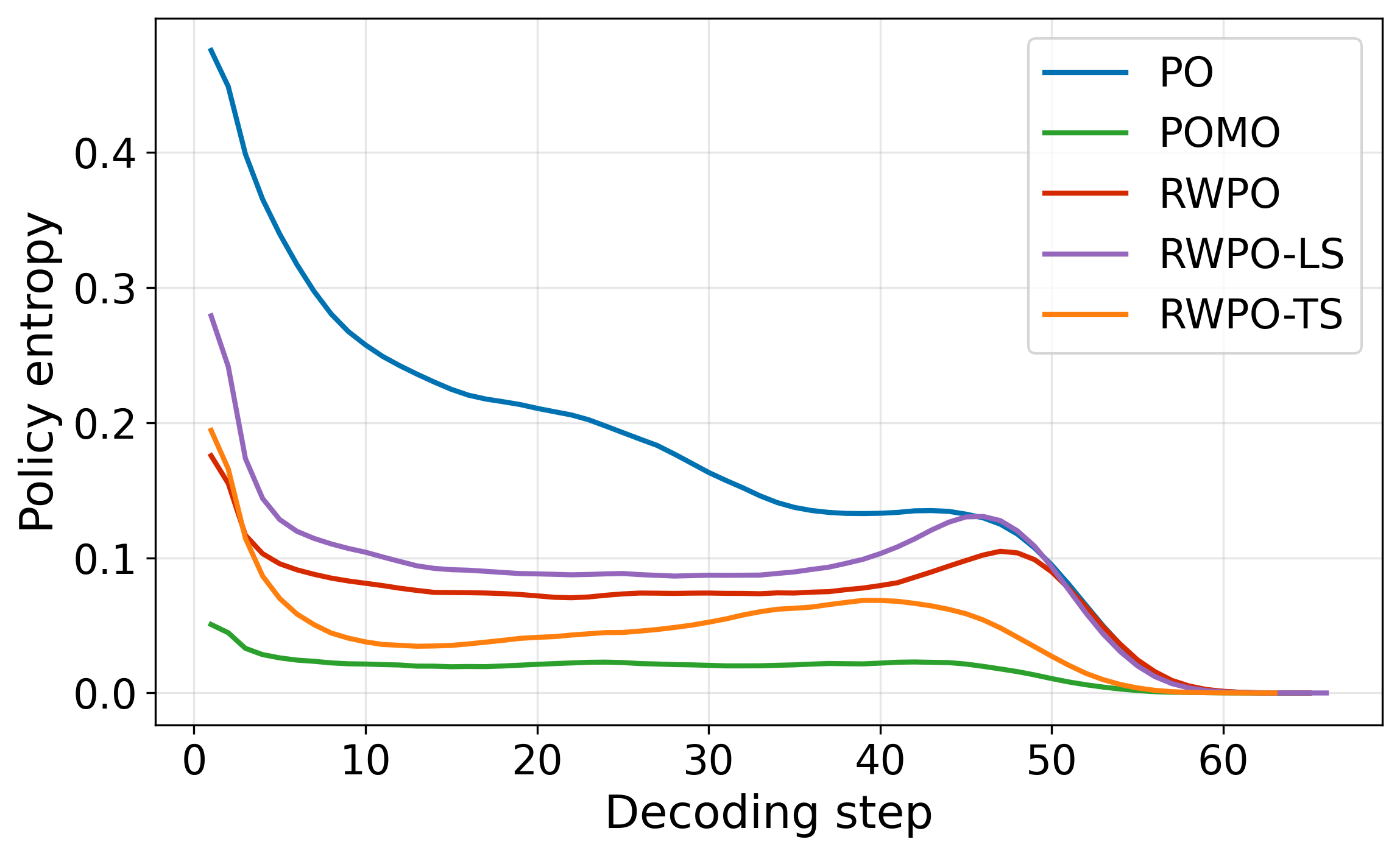}
        \vspace*{-5mm}
        \caption{$N=100$}
        \label{fig:stepwise_policy_entropy_n100}
    \end{subfigure}
    \hfill
    \begin{subfigure}[t]{0.24\textwidth}
        \centering
        \includegraphics[width=\textwidth,height=0.25\textheight,keepaspectratio]{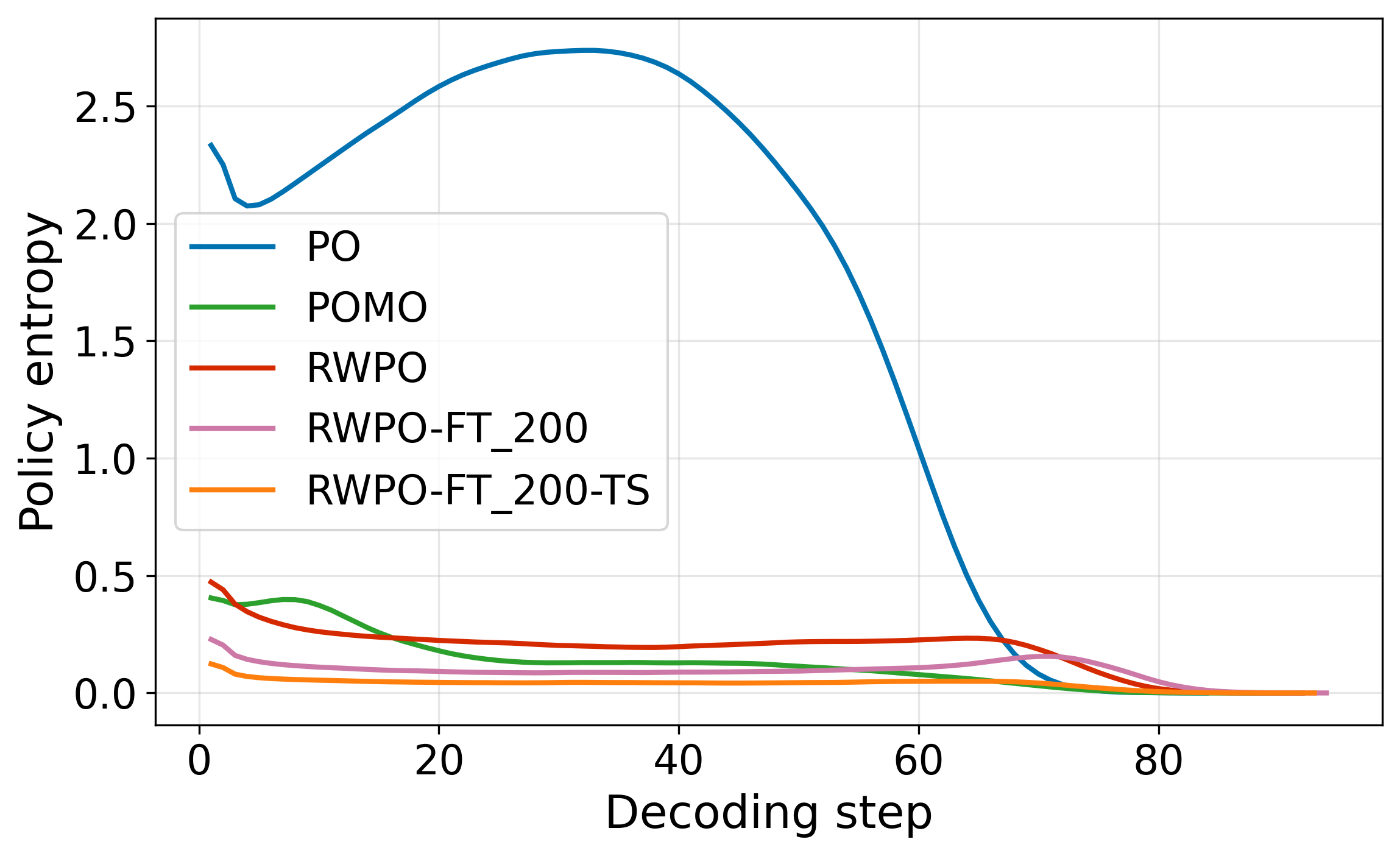}
        \vspace*{-5mm}
        \caption{$N=200$}
        \label{fig:policy_entropy_n200}
    \end{subfigure}
    \vspace{0.5em}
    \begin{subfigure}[t]{0.24\textwidth}
        \centering
        \includegraphics[width=\linewidth]{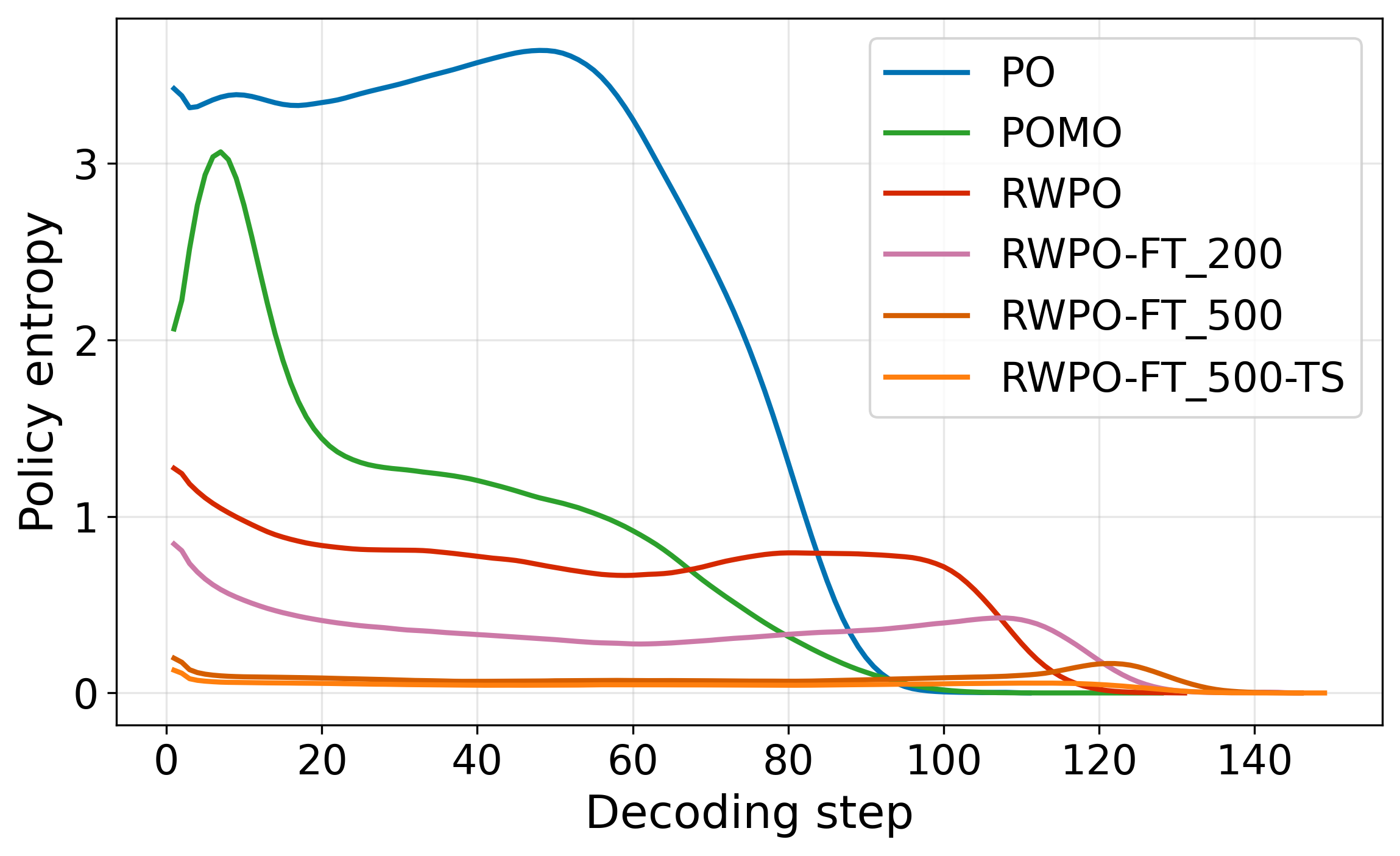}
        \vspace*{-5mm}
        \caption{$N=500$}
        \label{fig:stepwise_policy_entropy_n500}
    \end{subfigure}
    \hfill
    \begin{subfigure}[t]{0.24\textwidth}
        \centering
        \includegraphics[width=\linewidth]{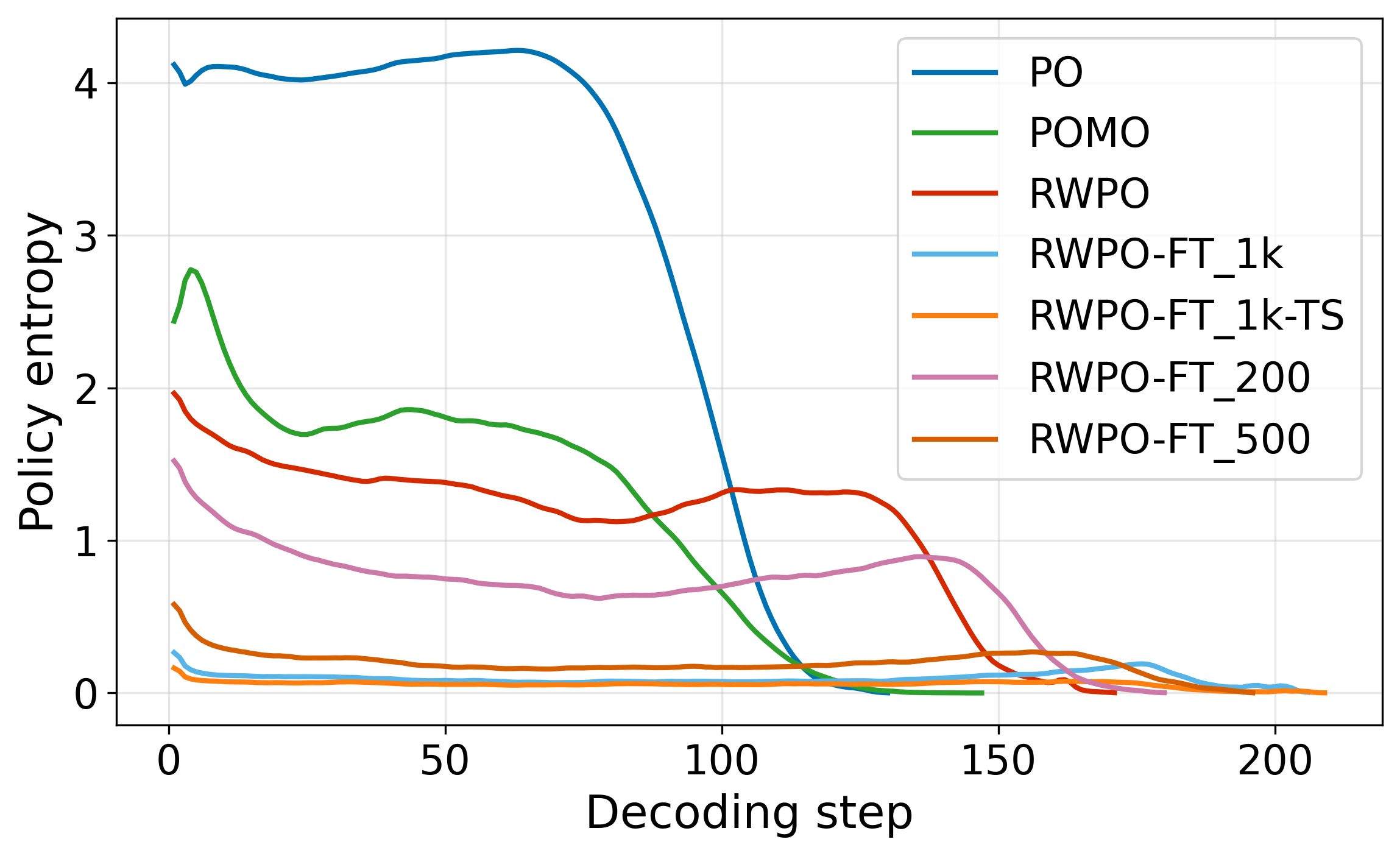}
        \vspace*{-5mm}
        \caption{$N=1{,}000$}
        \label{fig:stepwise_policy_entropy_n1000}
    \end{subfigure}
    \caption{
    Step-wise policy entropy during decoding on OP instances of different sizes. 
    }
    \label{fig:stepwise_policy_entropy}
\end{figure*}

\begin{figure*}[t]
    \centering

    \begin{subfigure}[t]{0.24\textwidth}
        \centering
        \includegraphics[width=\linewidth]{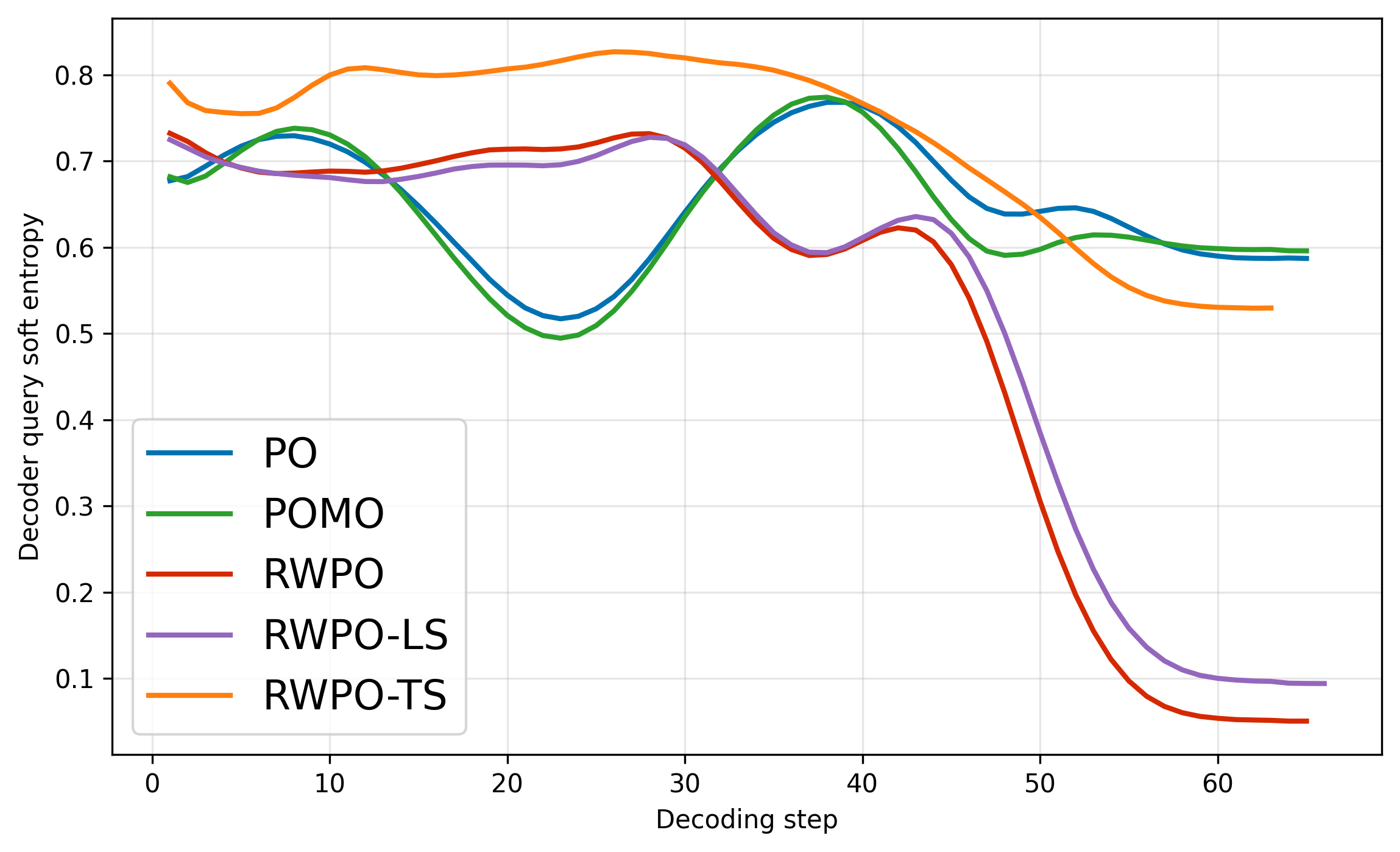}
        \vspace*{-5mm}
        \caption{$N=100$}
        \label{fig:entropy_stepwise_n100}
    \end{subfigure}
    \hfill
    \begin{subfigure}[t]{0.24\textwidth}
        \centering
        \includegraphics[width=\linewidth]{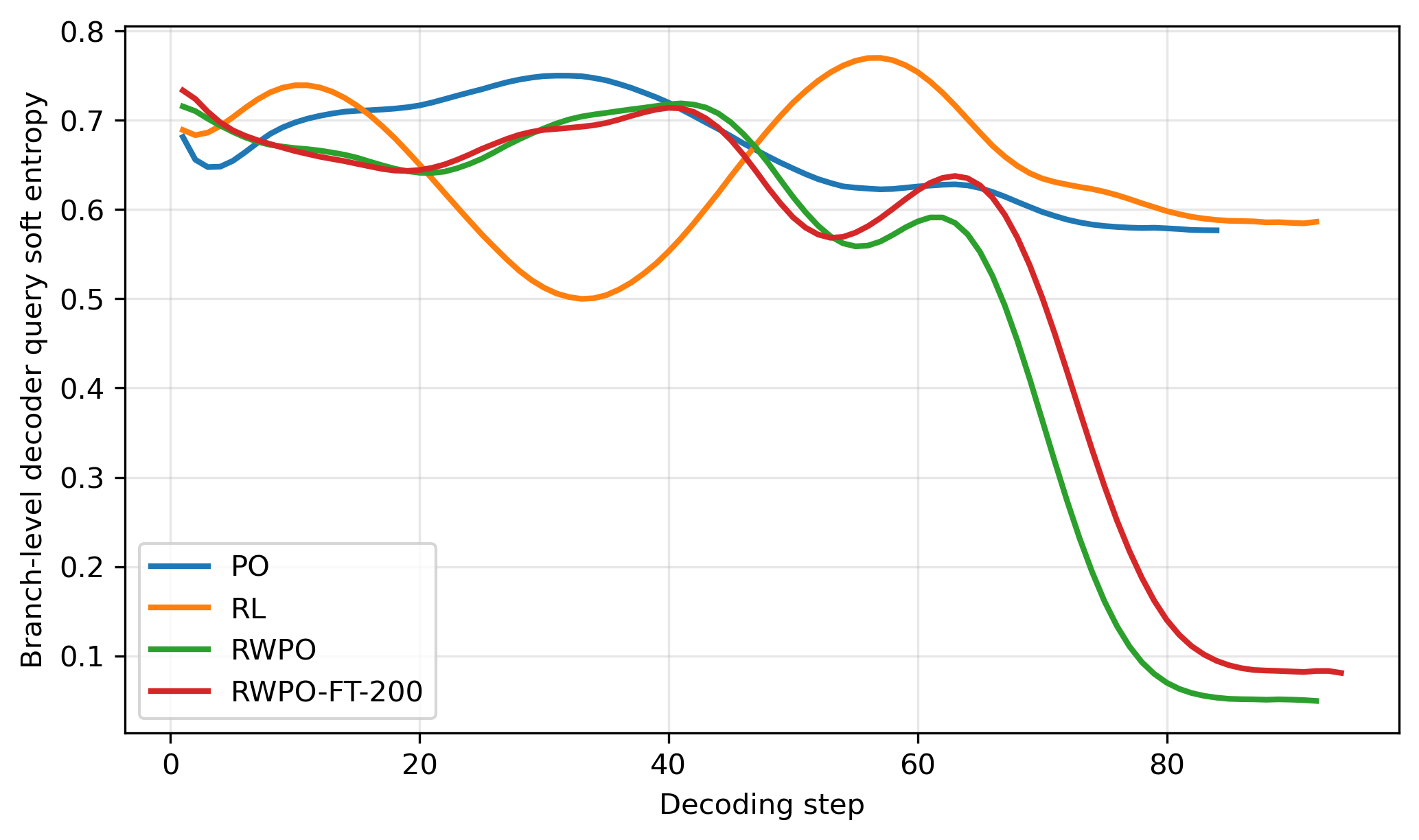}
        \vspace*{-5mm}
        \caption{$N=200$}
        \label{fig:entropy_stepwise_n200}
    \end{subfigure}
    \vspace{0.5em}
    \begin{subfigure}[t]{0.24\textwidth}
        \centering
        \includegraphics[width=\linewidth]{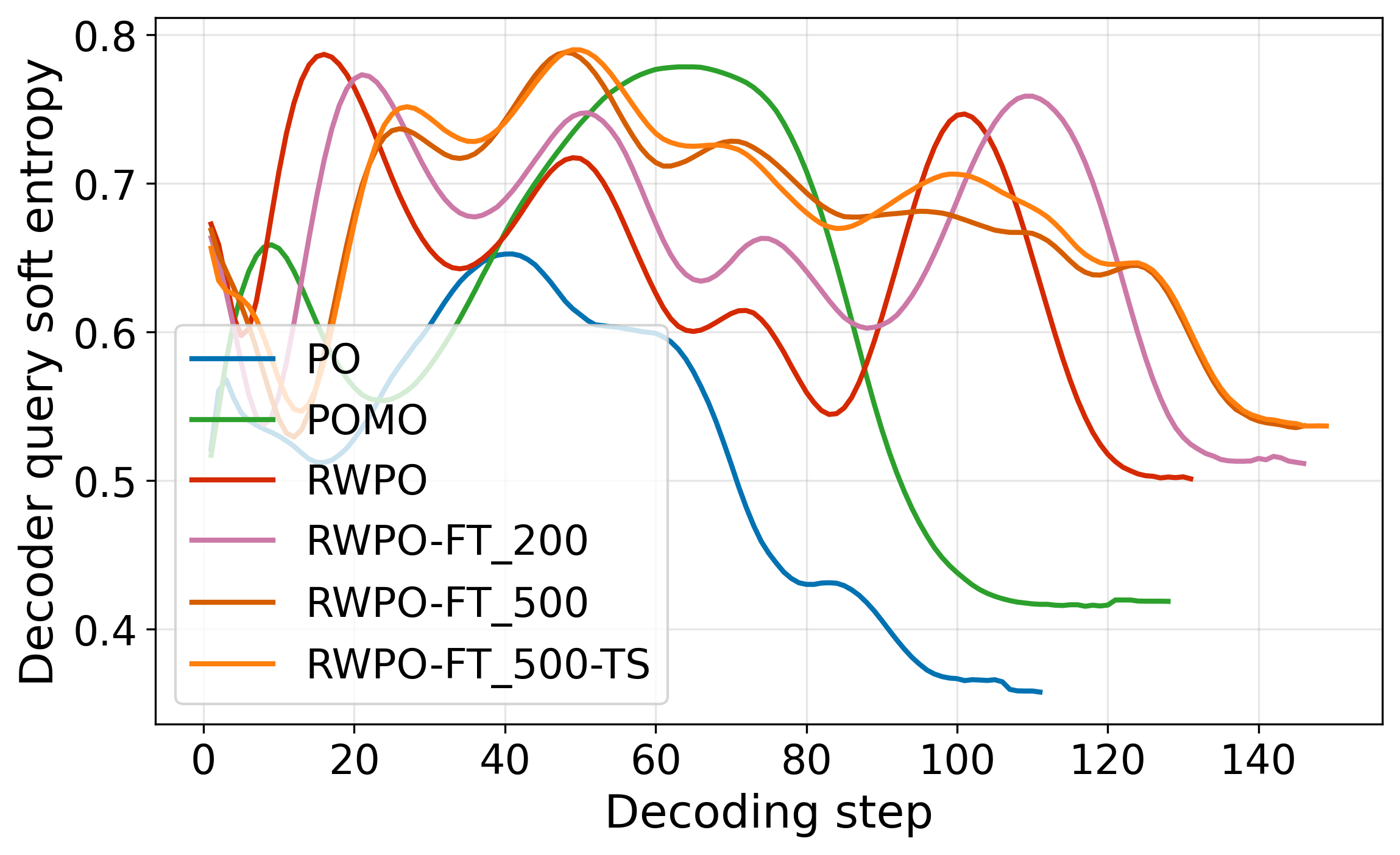}
        \vspace*{-5mm}
        \caption{$N=500$}
        \label{fig:entropy_stepwise_n500}
    \end{subfigure}
    \hfill
    \begin{subfigure}[t]{0.24\textwidth}
        \centering
        \includegraphics[width=\linewidth]{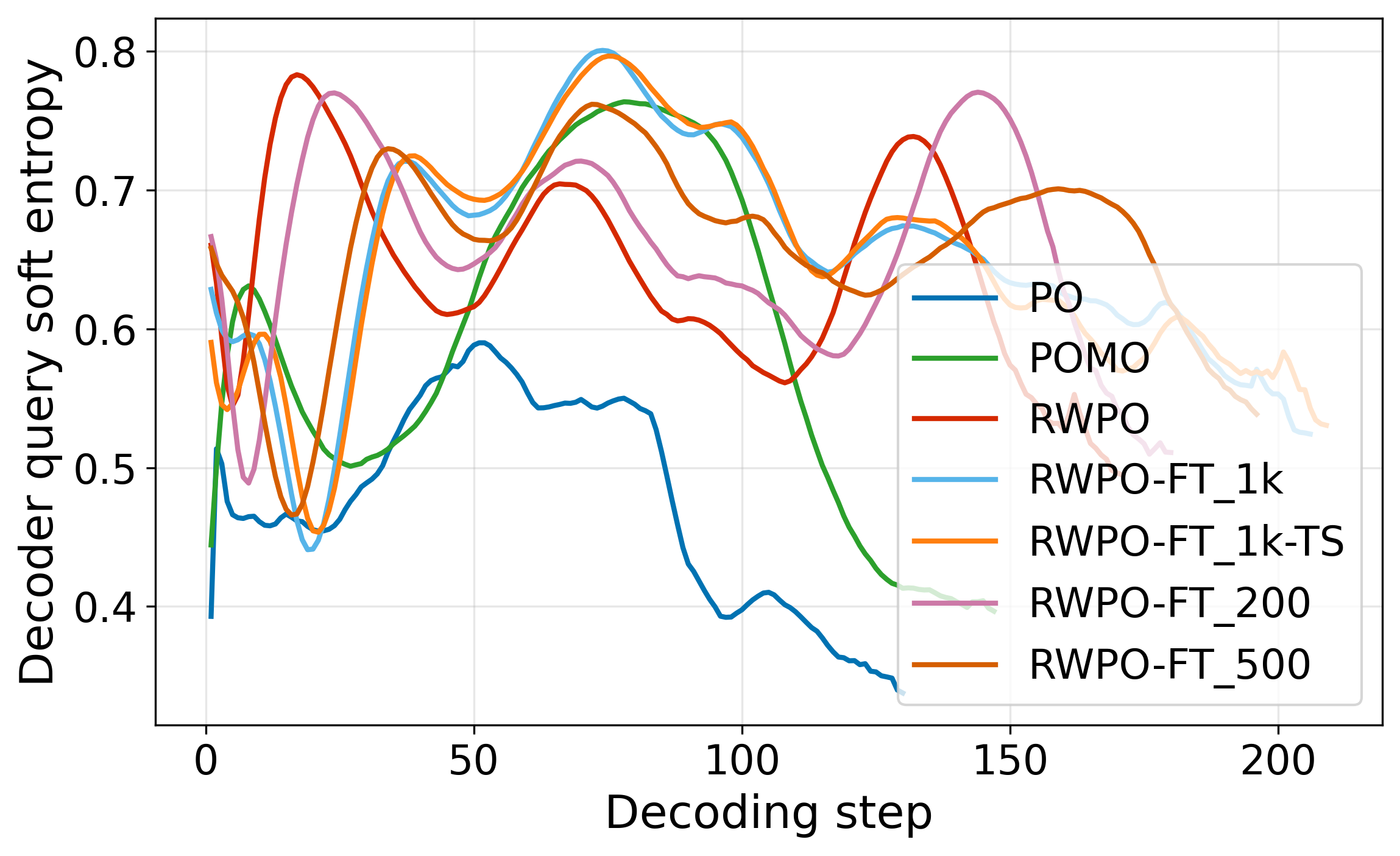}
        \vspace*{-5mm}
        \caption{$N=1{,}000$}
        \label{fig:entropy_stepwise_n1000}
    \end{subfigure}
    \caption{
Step-wise decoder-query soft entropy during decoding on OP instances of different sizes.
}
    \label{fig:stepwise_soft_entropy}
\end{figure*}

We contrast next-action uncertainty within individual rollouts (Fig.\ref{fig:stepwise_policy_entropy}) against query representation diversity across the batch (Fig.\ref{fig:stepwise_soft_entropy}).
On OP500 and OP1000, PO exhibits prolonged high policy entropy alongside comparatively lower decoder-query soft entropy, showing that uncertain action distributions need not be accompanied by diverse query representations. POMO attains high intermediate query entropy, but subsequently enters a sustained lower-entropy regime earlier than the RWPO variants. In contrast, RWPO exhibits lower early-stage policy entropy than both baselines while maintaining or recovering query diversity over substantial shared late-decoding intervals. Fine-tuned RWPO variants successfully decouple these metrics, maintaining high query diversity despite concentrated action distributions—a behavior most evident in RWPO-FT${500}$ (OP500) and RWPO-FT${1k}$ (OP1000). On OP100, RWPO-TS similarly exhibits higher query entropy but lower policy entropy than RWPO over most of decoding, although RWPO itself shows a pronounced late decline in both measures. Overall, the results show that RWPO variants can combine diverse query representations across partial solutions with concentrated next-action distributions within individual branches, rather than achieving representation diversity simply by maintaining greater action randomness.

\subsection{Ablation Study} \label{sec:ablation_study}

\begin{figure*}[t]
    \centering
    \begin{subfigure}[t]{0.24\textwidth}
        \centering
        \includegraphics[width=\linewidth]{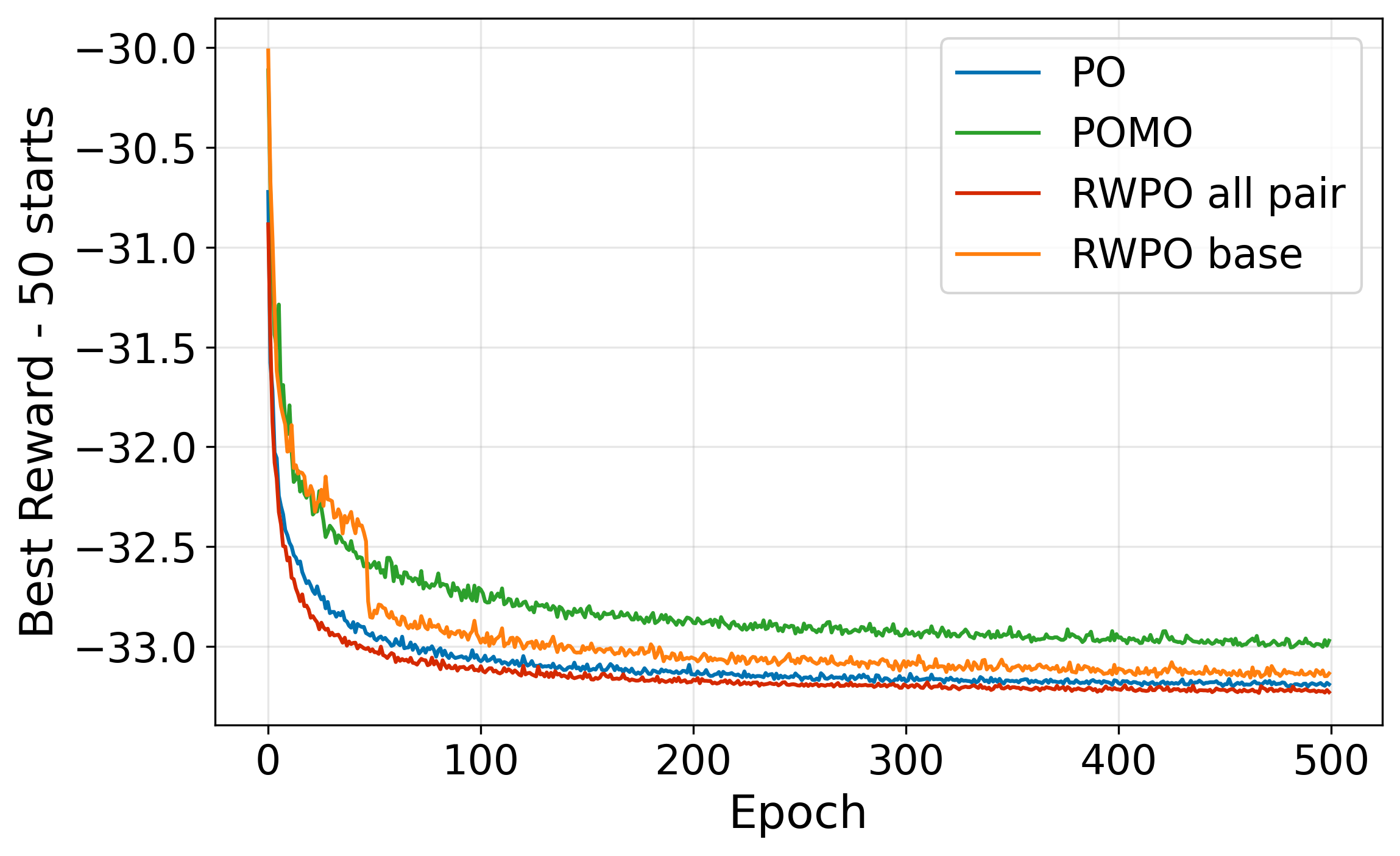}
        \vspace*{-5mm}
        \caption{Best reward on OP100}
        \label{fig:op_best_reward50}
    \end{subfigure}
    \hfill
    \begin{subfigure}[t]{0.24\textwidth}
        \centering
        \includegraphics[width=\linewidth]{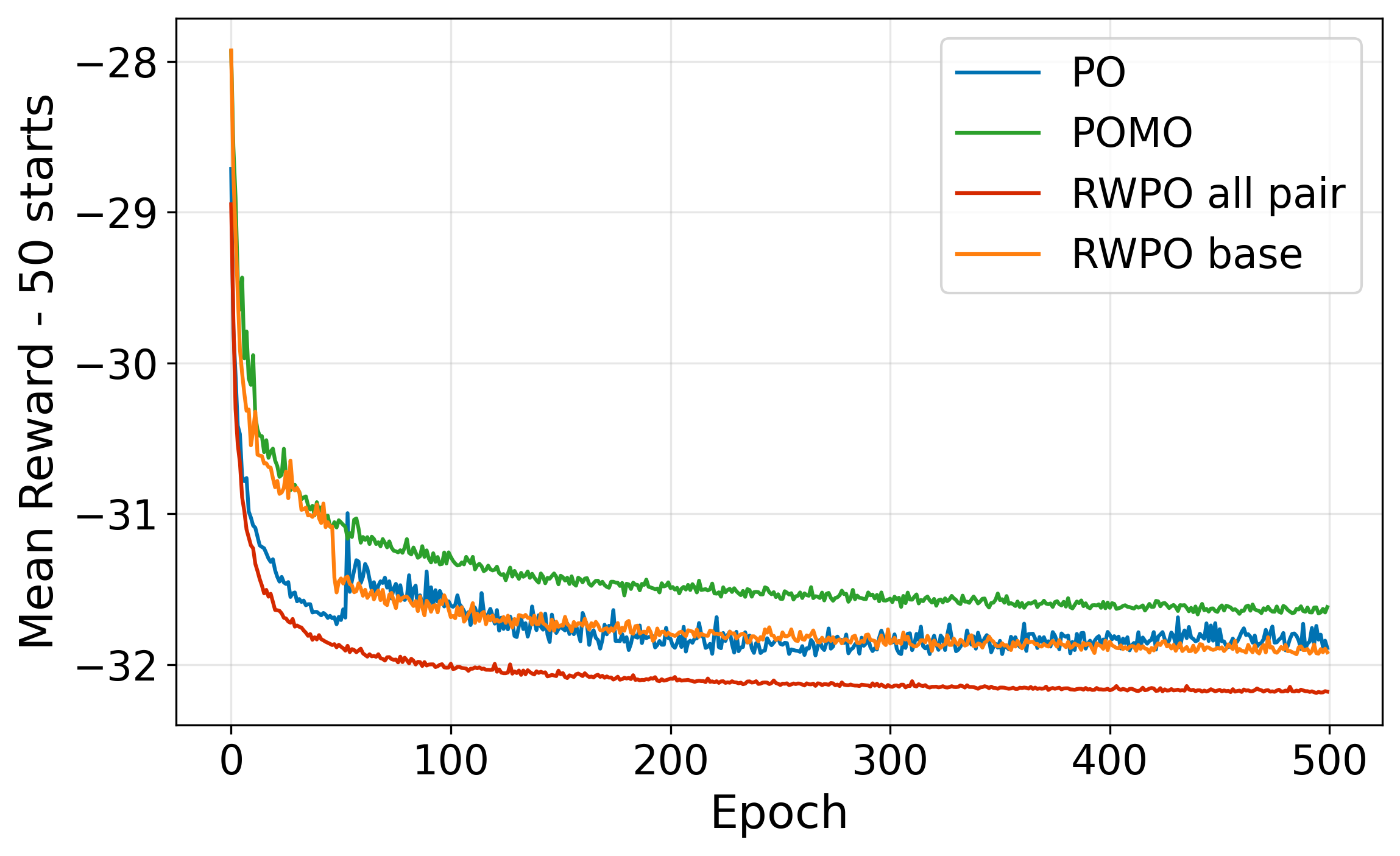}
        \vspace*{-5mm}
        \caption{Mean reward on OP100}
        \label{fig:op_mean_reward50}
    \end{subfigure}
    \hfill
    \begin{subfigure}[t]{0.24\textwidth}
        \centering
        \includegraphics[width=\linewidth]{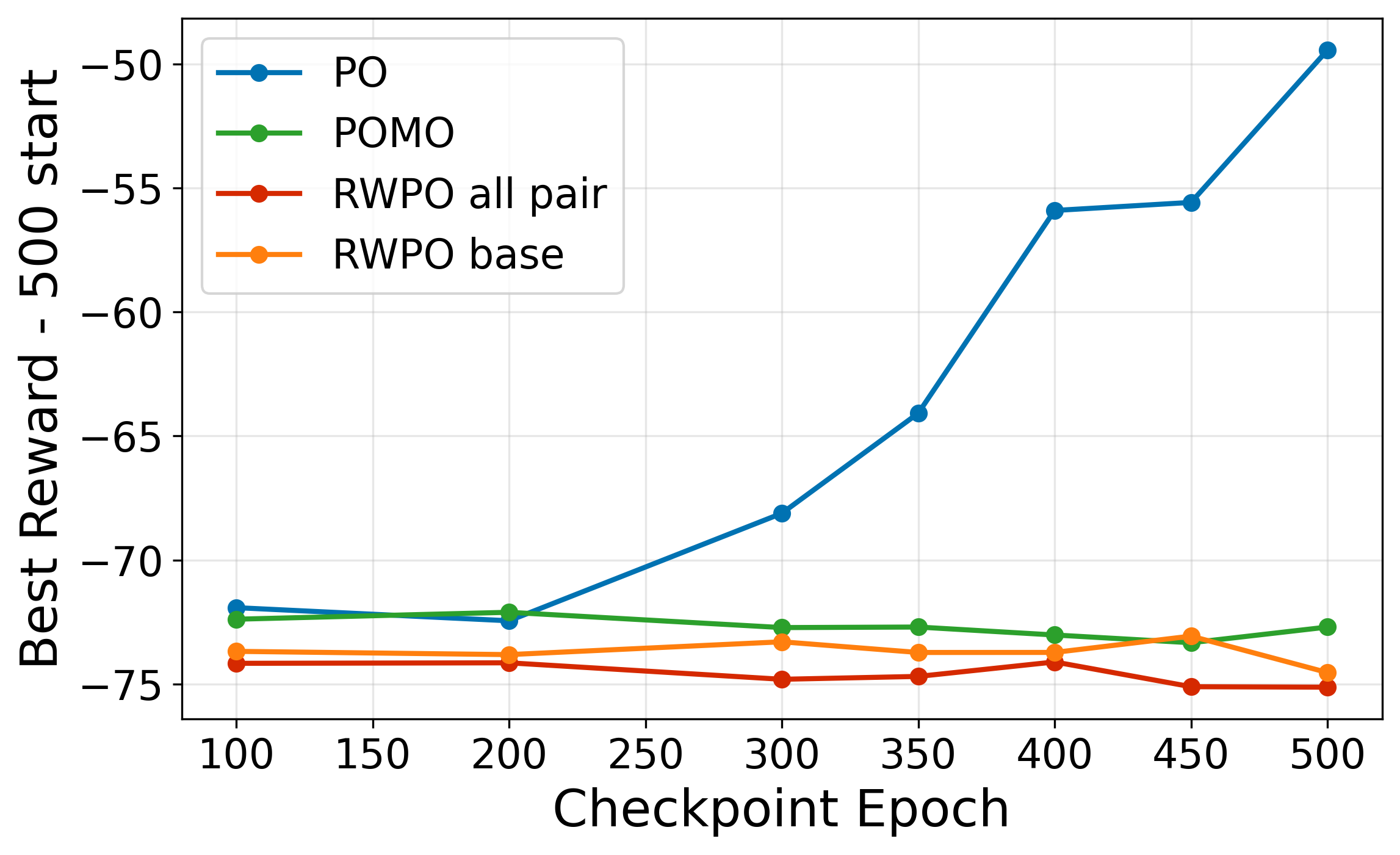}
        \vspace*{-5mm}
        \caption{Best reward on OP500}
        \label{fig:op_checkpoint_reward500}
    \end{subfigure}
    \hfill
    \begin{subfigure}[t]{0.24\textwidth}
        \centering
        \includegraphics[width=\linewidth]{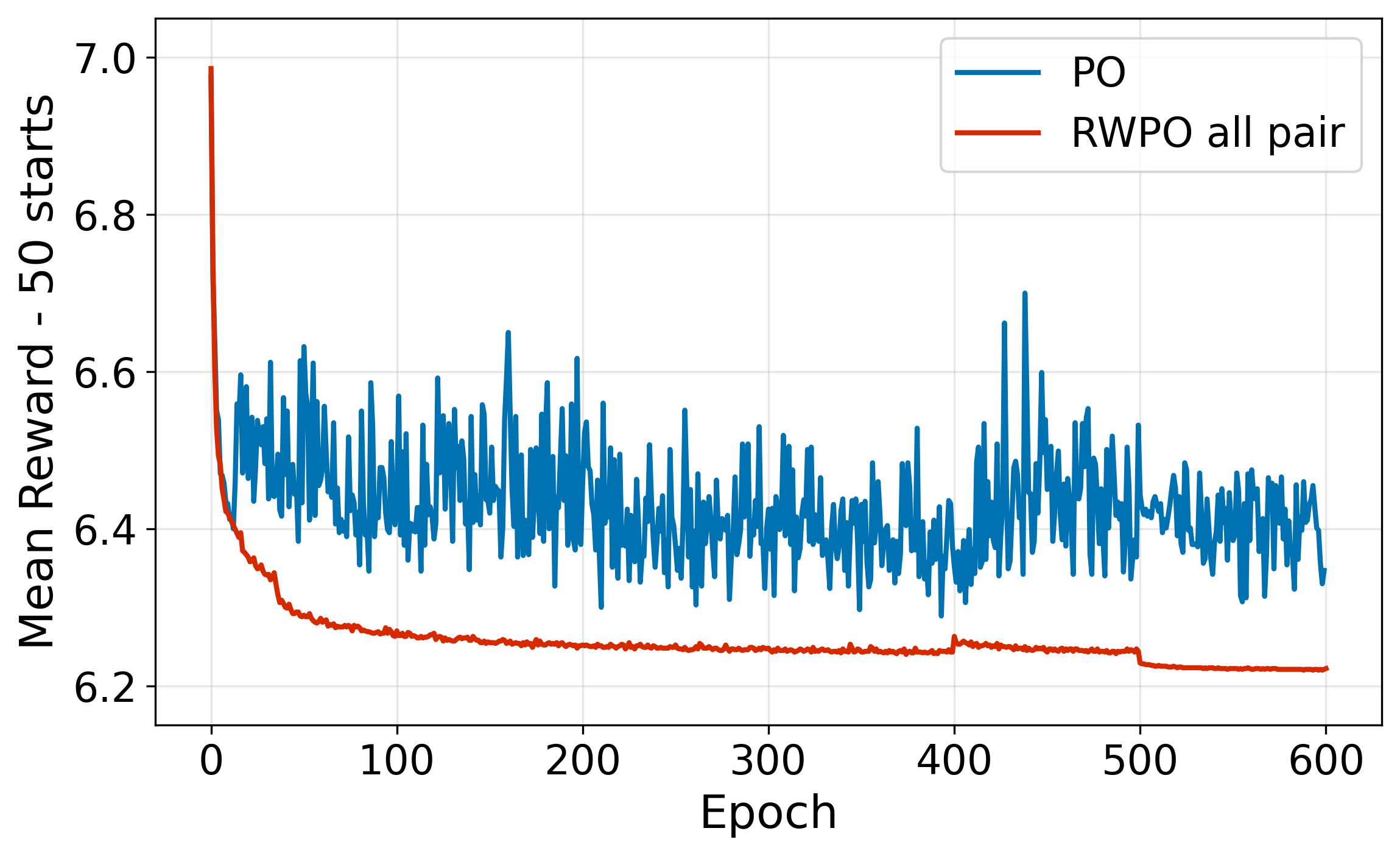}
        \vspace*{-5mm}
        \caption{Mean reward on PCTSP100}
        \label{fig:pctsp_topk_mean_reward}
    \end{subfigure}
    \caption{Training and evaluation curves on the OP and PCTSP problem.}
    \label{fig:op_training_curves}
\end{figure*}

Fig.~\ref{fig:op_best_reward50} shows that PO and both RWPO variants achieve similar best-reward values on OP100; however, this parity does not hold for the average performance across the $k$ rollouts. As shown in Fig.\ref{fig:op_mean_reward50}, the mean-reward curve of standard PO is highly oscillatory and fails to stabilize, whereas RWPO—particularly the all-pair variant—maintains a smooth, stable trajectory. A similar pattern emerges in the OP500 OOD evaluation (Fig.\ref{fig:op_checkpoint_reward500}), as well as the PCTSP (Fig.~\ref{fig:pctsp_topk_mean_reward}) and SPCTSP/KP results (Appendix~\ref{app:training_curves}). Because top-$k$ POMO aggregates the learning signal across all $k$ rollouts, the learned policy should improve the entire rollout set collectively rather than merely optimizing its single best member. The observation that PO retains a competitive best reward while exhibiting unstable mean performance supports our hypothesis: without explicit reward magnitudes, purely ordinal updates fail to properly differentiate the relative quality of the full rollout batch. In contrast, RWPO all-pair remains stable across all tested problem settings, preserving the stabilizing properties of symmetric pairwise learning while explicitly scaling its gradient updates by the reward gaps.


\section{Discussion and Conclusion}
\label{sec:discuss}\label{sec:conc}

This work presented an effective learning-based framework for selective combinatorial optimization problems (SCOPs). Central to our approach is Reward-Weighted Preference Optimization (RWPO), which improves upon standard preference learning by rigorously preserving reward-gap information. Unlike standard PO, which reduces trajectory comparisons to purely ordinal binary preferences, RWPO scales the loss of each pair by their exact reward difference, directly aligning the agent's learning signal with reward maximization. This mechanism is essential for SCOPs, where subset-selection decisions cause feasible solutions to vary drastically in actual quality even when correctly ordered. By naturally down-weighting weak comparisons, RWPO learns a highly constraint-aware policy, delivering stronger out-of-distribution (OOD) generalization than PO while maintaining much greater training stability than standard POMO. When integrated with our complete adaptation pipeline—comprising top-$k$ selective decoding, scale-specific fine-tuning, and tilted sampling—RWPO establishes state-of-the-art performance across four SCOP benchmarks.

Although RWPO shows strong performance on SCOPs, it relies on exact reward evaluations, which could be difficult to obtain in partially observable or multi-objective environments.  Second, its reward-weighting strength and length-normalization coefficient are currently fixed, suggesting future works on adaptive weighting based on reward dispersion, instance difficulty, or route-length statistics.  Finally, while our entropy analyses reveal a strong correlation between representation diversity and generalization, they are purely observational. Future works could employ representation interventions to definitively prove whether these diverse decoder queries are the direct causal factors of the model's OOD robustness.

\section*{Acknowledgments}
This publication is part of the project FMaaS - it's a cargo match (with project number NWA.1518.22.023) of the research programme Dutch Research Agenda - research along NWA routes by consortia which is (partly) financed by the Dutch Research Council (NWO).

\bibliographystyle{plainnat}  
\bibliography{references}

\appendix
\section{Related Work}
\label{app:related_work}

\subsection{RL-based Neural Network Solvers}

Reinforcement-learning-based neural combinatorial optimization (NCO) formulates solution construction as a finite-horizon decision process.  A neural policy incrementally selects feasible nodes, edges, or assignments, and the resulting solution quality provides the training signal.  Early \textbf{end-to-end methods} used pointer-style recurrent networks trained with REINFORCE for the traveling salesperson problem and knapsack problems \citep{bello2016neural}. Later work incorporated dynamic demands and capacities for vehicle routing \citep{nazari2018reinforcement}, graph-embedding-based Q-learning \citep{khalil2017learning}, and attention-based encoder--decoder architectures \citep{kool2019attention}. POMO improves exploration by constructing multiple solutions from different starting points and optimizing their returns jointly \citep{kwon2020pomo}, while Sym-NCO leverages problem symmetries to boost performance \citep{kim2022sym}. MDAM trains multiple decoders and applies a pairwise KL-divergence regularizer at the initial step to diversify their action distributions \citep{xin2021multi}. Recent neural vehicle-routing studies address generalization through adaptive multi-distribution knowledge distillation \citep{bi2022learning}, policy-space exploration with curriculum-based scale adaptation \citep{wang2024asp}, meta-learning for rapid adaptation to new tasks \citep{zhou2023towards}, ensemble RL with mixed-distribution training and diversity regularization \citep{jiang2023ensemble}, and geometry- and density-conditioned policy adaptation for large-scale generalization \citep{zhou2026instance}.
Collectively, these methods modify the policy architecture, action-space representation, search procedure, conditioning mechanism, or training distribution. Our work is complementary in that it modifies the optimization objective through reward-weighted preference optimization (RWPO), using pairwise preferences together with reward gaps to guide policy updates.

\textbf{Hybrid neural solvers} combine learned policies with classical search or optimization. NeuRewriter learns local rewriting operations for improving partial solutions \citep{chen2019learning}, while Neural Large Neighborhood Search learns repair operators within a destroy--repair framework \citep{hottung2020neural}. Neural Deconstruction Search follows a related iterative improvement paradigm by learning an RL policy to remove customers from incumbent routes and combining it with greedy insertion and augmented simulated annealing to reconstruct and improve the solutions \citep{hottung2025neural}. Expanding this learned-heuristic direction, PyVRP$^{+}$ employs LLM-driven heuristic evolution within a hybrid genetic search framework \citep{malik2026pyvrp}. Other approaches instead pair deep reinforcement learning with exact subproblem solvers or constraint programming \citep{cappart2021combining,prantare2022learning}. In summary, these hybrid methods improve the reliability of neural construction by retaining classical search, but they introduce complex algorithmic design choices concerning solver calls, neighborhood selection, and the allocation of credit between learned and hand-crafted components. 
Furthermore, extending these iterative local-search operators to selective combinatorial optimization problems (SCOPs) is notoriously difficult, as the agent must simultaneously optimize routing efficiency and search the massive combinatorial space of unvisited nodes without violating strict global budget constraints.

\subsection{Neural Divide-and-Conquer Methods}

\textbf{Neural divide-and-conquer methods} improve the scalability of data-driven approaches by partitioning a large instance, solving smaller subproblems, and combining the resulting partial solutions. Early networks in this direction learned recursive split-and-merge operations for generic combinatorial tasks \citep{nowak2018divide}. To tackle large-scale routing specifically, frameworks like Learning to Delegate, RBG, and H-TSP employ hierarchical policies to assign instance components to localized solvers \citep{li2021learning,zong2022rbg,pan2023htsp}. Building on this, several recent approaches explicitly separate global partitioning from local construction. For example, TAM learns a constraint-aware VRP partition that allows local TSP heuristics to generalize to larger instances \citep{hou2023generalize}, while GLOP and Select-and-Optimize alternate between selecting local subproblems and applying learned optimization procedures \citep{ye2024glop,cheng2023select}. To address the compounded errors that arise from these splits, Hierarchical Learning-based Graph Partitioning (HLGP) uses dual global and local policies \citep{pan2025hierarchical}, and UDC trains a graph-neural divider jointly with a subpath solver via a unified Divide--Conquer--Reunion objective \citep{zheng2024udc}. Pushing this paradigm further, DRAGON deploys LLM-driven agents to identify promising regions and reconstruct global solutions using feedback-guided experience memory \citep{chen2026dragon}. These approaches highlight a fundamental challenge in scalable neural solvers: partition decisions must be rigorously evaluated against the completed solution, as an unfavorable early split irreversibly restricts cross-partition connections and bounds the improvements available to local solvers. This vulnerability is especially pronounced in SCOPs, where localized partitions prematurely fragment the global budget and prevent the solver from discovering highly rewarding, globally dispersed subsets.

\subsection{Preference-based Reinforcement Learning Methods}

Preference-based reinforcement learning (PbRL) learns from comparisons between trajectories or candidate solutions rather than requiring a fully specified scalar reward. Early approaches inferred latent utilities from pairwise preferences and used preference-based policy iteration or Bayesian model-free policy updates \citep{akrour2011preference,furnkranz2012preference,wirth2016model,wirth2017survey}. Deep RL from Human Preferences learned a neural reward model from pairwise trajectory-segment labels and optimized the learned reward with RL \citep{christiano2017deep,ibarz2018reward}. PEBBLE improved feedback efficiency through active query selection, unsupervised exploration, and replay relabeling \citep{lee2021pebble}. These methods provide a general framework for learning objectives that are difficult to write explicitly, but their performance depends on the coverage and accuracy of the preference or reward model.

Direct preference-optimization methods avoid a separate reward-model training stage. DPO derives a pairwise objective from the KL-regularized RLHF solution \citep{rafailov2023direct}, while IPO modifies the preference objective to reduce overfitting under deterministic labels \citep{azar2024general}. P3O performs policy updates directly from comparative trajectory feedback \citep{wu2023pairwise}. WPO reweights preference examples according to their likelihood under the current policy to reduce off-policy distribution shift \citep{zhou2024wpo}; these weights reflect policy likelihood rather than the reward gap between two solutions.

Recent neural-based methods have successfully adapted preference signals to avoid scalar reward modeling. For example, \citet{pan2025preference} use statistical preference models to derive a preference-learning objective from entropy-regularized RL, with local search generating additional preference pairs for fine-tuning. BOPO similarly establishes preference optimization as a viable data-driven paradigm by constructing best-anchored pairs and scaling its pairwise loss using reward differences \citep{liao2025bopo}. Our method fundamentally differs in the scope and application of this reward signal. Rather than relying on anchor-based comparisons, the reward-dependent weight in RWPO dictates the contribution of every preference pair. Thus, the RWPO update strictly preserves both the ordinal direction of the preference and the cardinal magnitude of the reward improvement. This dual signal is mathematically critical for SCOPs, where feasible subsets are massively numerous, frequently near-tied in routing cost, but differ drastically in their accumulated global reward.

\section{Additional Methodology Details}

We provide further details to Section~3 of the main paper.

\subsection{Decoding-Length Bias of Summed Log-Probabilities in SCOPs}
\label{app:length_normalization}

In Preference Optimization (PO) and Reward-Weighted Preference Optimization (RWPO), the standard preference objective increases the trajectory
log-probability of preferred solutions. For an autoregressive constructive
policy, the trajectory log-probability is
\begin{equation}
\log \pi_\theta(\tau\mid x)
=
\sum_{t=1}^{T(\tau)}
\log \pi_\theta(a_t\mid s_t,x),
\label{eq:trajectory_logprob}
\end{equation}
where $T(\tau)$ denotes the number of decoding steps in trajectory $\tau$.

This summed log-probability can introduce a length bias in SCOPs.
Since each per-step log-probability is non-positive, longer trajectories
typically have more negative total log-probability even when their average
per-step confidence is comparable to that of shorter trajectories. This can
cause PO to favor shorter routes beyond the intended
preference for higher-reward solutions.

To mitigate this effect, we use the following length-adjusted trajectory
score:
\begin{equation}
\ell_\theta(\tau\mid x)
=
T(\tau)^{-\beta}
\sum_{t=1}^{T(\tau)}
\log \pi_\theta(a_t\mid s_t,x),
\qquad
\beta\in[0,1].
\label{eq:length_normalized_score}
\end{equation}
This quantity is a scoring function and, for $\beta\neq 0$, is not the
log-probability of a normalized trajectory distribution. In particular,
$\ell_\theta(\tau\mid x)=\log\pi_\theta(\tau\mid x)$ when $\beta=0$, whereas
$\beta=1$ gives the average per-step log-probability.

In our experiments, we use $\beta=0.3$ as a partial length correction in OP and KP, and use $\beta=0.1$ in PCTSP and SPCTSP. This hyperparameter distinction reflects the underlying structural biases of the problem classes: the OP and KP inherently encourage visiting more nodes to maximize accumulated rewards within a budget, whereas the PCTSP and SPCTSP penalize routing costs and therefore favor shorter decision sequences.
The resulting preference term is
\begin{equation}
\log \sigma
\left(
\alpha
\left[
\ell_\theta(\tau_i\mid x)
-
\ell_\theta(\tau_j\mid x)
\right]
\right).
\label{eq:length_adjusted_preference}
\end{equation}


\subsection{Motivation for Reward-Weighted Preference Optimization} \label{appendix:motivation}
\paragraph{Illustrative Example.}
Consider an extreme example with two preference pairs under the Bradley–Terry (BT) model. 
First, suppose $\tau_i \succ \tau_j$ with
\[
\log \pi_\theta(\tau_i|x) 
= \log \pi_\theta(\tau_j|x) + 0.01,
\quad 
r(x,\tau_i) = r(x,\tau_j) + 10.
\]
Second, suppose $\tau_3 \succ \tau_4$ with
\[
\log \pi_\theta(\tau_3|x) 
= \log \pi_\theta(\tau_4|x) - 50,
\quad 
r(x,\tau_3) = r(x,\tau_4) + 0.01.
\]

Under the pairwise logistic model
\[
p_\theta(\tau_i \succ \tau_j \mid x)
=
\sigma\!\big(\alpha \big(\log \pi_\theta(\tau_i|x) - \log \pi_\theta(\tau_j|x)\big)\big),
\]
with $\alpha = 0.05$, we obtain
\[
p_\theta(\tau_i \succ \tau_j) 
= \sigma(0.05 \times 0.01) 
\approx 0.50 ,
\]
\[
p_\theta(\tau_3 \succ \tau_4) 
= \sigma(0.05 \times (-50)) 
\approx 0.08.
\]

Thus, the model assigns approximately $50\%$ probability to the correct preference in the first pair and only $8\%$ in the second pair. The corresponding log-likelihood terms are
\[
\log p_\theta(\tau_i \succ \tau_j) \approx -0.69,
\qquad
\log p_\theta(\tau_3 \succ \tau_4) \approx -2.58.
\]

Since the gradient magnitude of the binary cross-entropy loss scales with 
$\sigma(z) - 1$ (for positive labels), 
the second pair produces a substantially larger gradient than the first. 
Consequently, the optimization strongly pushes the model to correct the misranking of $(\tau_3,\tau_4)$, 
even though their reward gap is negligible, 
while only weakly encouraging preference between $(\tau_i,\tau_j)$ despite their large reward difference.

\paragraph{Fundamental Issue.}
This example reveals a structural limitation of the PO: 
the update magnitude depends solely on the model's probability error, 
rather than on the reward gap between trajectories. 
As a result, pairs with small reward differences but large probability errors can dominate the gradient, 
whereas pairs with substantial reward gaps may induce only minor updates. 

To tackle this issue, we employ a weighted pairwise logistic likelihood, scaling each comparison by a function of the reward gap to emphasize more informative preferences. This is conceptually aligned with reward-weighted likelihood/regression in policy learning \citep{peters2007reinforcement,peng2019advantage} and with importance-weighted pairwise losses in learning-to-rank (e.g., LambdaRank-style reweighting) \citep{burges2006learning}.

\subsection{Gradient Analysis of RWPO Objectives}
\label{app:rwpo_gradient}

We analyze the gradient behavior of the ordered RWPO objective and the
implemented confidence-regularized all-pair RWPO objective. For a sampled
trajectory pair, define the normalized trajectory log-likelihood margin as
\begin{equation}
    m_\theta(x,\tau_i,\tau_j)
    =
    \ell_\theta(\tau_i|x)
    -
    \ell_\theta(\tau_j|x).
\end{equation}
For simplicity, let
\[
p_{ij}
=
\sigma(\alpha m_\theta(x,\tau_i,\tau_j)).
\]

\paragraph{Gradient of ordered RWPO}
For an ordered preference pair $(\tau^+,\tau^-)$ with
$r(x,\tau^+)>r(x,\tau^-)$, the RWPO loss is
\begin{equation}
    \mathcal{L}_{\mathrm{RWPO}}
    =
    -\omega_{+-}
    \log
    \sigma(\alpha m_\theta),
\end{equation}
where
\begin{equation}
    \omega_{+-}
    =
    1+
    \lambda
    \frac{
    r(x,\tau^+)-r(x,\tau^-)
    }
    {\bar{\Delta}^{(b)}+\epsilon}.
\end{equation}

Taking the gradient with respect to $\theta$ gives
\begin{equation}
\begin{aligned}
\nabla_\theta
\mathcal{L}_{\mathrm{RWPO}}
&=
-\omega_{+-}
\nabla_\theta
\log\sigma(\alpha m_\theta)
\\
&=
-\alpha
\omega_{+-}
\sigma(-\alpha m_\theta)
\nabla_\theta m_\theta .
\end{aligned}
\end{equation}

Because $\omega_{+-}$ increases monotonically with the reward gap,
larger improvements between the preferred and rejected trajectories
produce larger gradient magnitudes when the current preference margin is
identical. Therefore, RWPO assigns stronger optimization pressure to
comparisons with larger reward improvements while preserving the same
preference probability mapping $\sigma(\alpha m_\theta)$ for all pairs as PO.

\paragraph{Gradient of confidence-regularized all-pair RWPO}
Unlike the ordered RWPO objective, which increases preference-separation pressure through reward-dependent gradients, the confidence-regularized all-pair RWPO objective induces a finite reward-gap-dependent optimal preference margin.

For an unordered rollout pair $\{\tau_i,\tau_j\}$, define
\[
m_{ij}
=
\ell_\theta(\tau_i|x)
-
\ell_\theta(\tau_j|x).
\]
The implemented all-pair objective for this pair is
\begin{equation}
\mathcal{L}_{ij}
=
-
\tilde{\omega}_{ij}
\log\sigma(\alpha m_{ij})
-
\tilde{\omega}_{ji}
\log\sigma(-\alpha m_{ij}),
\end{equation}
where
\begin{equation}
\tilde{\omega}_{ij}
=
1+
\lambda
\frac{
[r(x,\tau_i)-r(x,\tau_j)]_+
}
{\bar{\Delta}^{(b)}+\epsilon},
\end{equation}
and
\begin{equation}
\tilde{\omega}_{ji}
=
1+
\lambda
\frac{
[r(x,\tau_j)-r(x,\tau_i)]_+
}
{\bar{\Delta}^{(b)}+\epsilon}.
\end{equation}

The derivative with respect to the pairwise margin is
\begin{equation}
\begin{aligned}
\frac{\partial\mathcal{L}_{ij}}
{\partial m_{ij}}
&=
-\alpha
\tilde{\omega}_{ij}
\sigma(-\alpha m_{ij})
+
\alpha
\tilde{\omega}_{ji}
\sigma(\alpha m_{ij})
\\
&=
\alpha
\left[
(\tilde{\omega}_{ij}
+
\tilde{\omega}_{ji})
\sigma(\alpha m_{ij})
-
\tilde{\omega}_{ij}
\right].
\end{aligned}
\end{equation}

For a pair where $\tau_i$ has higher reward than $\tau_j$,
\[
r(x,\tau_i)>r(x,\tau_j),
\]
we have
\[
\tilde{\omega}_{ij}
=
1+
\lambda
\frac{\Delta_{ij}}
{\bar{\Delta}^{(b)}+\epsilon},
\qquad
\tilde{\omega}_{ji}=1 .
\]

Therefore,
\begin{equation}
\frac{\partial\mathcal{L}_{ij}}
{\partial m_{ij}}
=
\alpha
\left[
\left(
2+
\lambda
\frac{\Delta_{ij}}
{\bar{\Delta}^{(b)}+\epsilon}
\right)
\sigma(\alpha m_{ij})
-
\left(
1+
\lambda
\frac{\Delta_{ij}}
{\bar{\Delta}^{(b)}+\epsilon}
\right)
\right].
\end{equation}

The gradient consists of two components. The unit terms produce the
symmetric preference regularization:
\[
\alpha(2\sigma(\alpha m_{ij})-1),
\]
which prevents unconstrained growth of the pairwise margin. The unit baseline alone does not learn preference ordering. It acts as a margin regularizer that prevents the model from pushing all pairwise differences to \(+\infty\). The
reward-gap-dependent component is
\[
\lambda
\frac{\Delta_{ij}}
{\bar{\Delta}^{(b)}+\epsilon}
\alpha
(\sigma(\alpha m_{ij})-1),
\]
which increases the preference pressure toward the higher-reward
trajectory as the reward gap grows.

\begin{proposition}[Reward-gap-dependent preference margin]
\label{prop:rwpo_margin}
For a fixed trajectory pair $(\tau_i,\tau_j)$ with
$r(x,\tau_i)>r(x,\tau_j)$, the confidence-regularized all-pair
RWPO loss induces an optimal log-likelihood margin
\begin{equation}
    m_{ij}^{*}
    =
    \frac{1}{\alpha}
    \log
    \left(
    1+
    \lambda
    \frac{
    r(x,\tau_i)-r(x,\tau_j)
    }
    {\bar{\Delta}^{(b)}+\epsilon}
    \right),
\end{equation}
where
\begin{equation}
    m_{ij}
    =
    \ell_\theta(\tau_i|x)
    -
    \ell_\theta(\tau_j|x).
\end{equation}
Therefore, RWPO assigns larger optimal preference margins to trajectory
pairs with larger normalized reward gaps, while allowing near-tie pairs
to remain weakly separated.
\end{proposition}

\begin{proof}
For a pair $(\tau_i,\tau_j)$ with
$r(x,\tau_i)>r(x,\tau_j)$, as the derivative of $\mathcal{L}_{ij}$ with respect to $m_{ij}$ gives
\begin{equation}
\frac{\partial\mathcal{L}_{ij}}
{\partial m_{ij}}
=
\alpha
\left[
(\tilde{\omega}_{ij}+1)
\sigma(\alpha m_{ij})
-
\tilde{\omega}_{ij}
\right].
\end{equation}

At the stationary point,
\begin{equation}
\sigma(\alpha m_{ij}^{*})
=
\frac{\tilde{\omega}_{ij}}
{\tilde{\omega}_{ij}+1}
=
\frac{
1+
\lambda
\frac{\Delta_{ij}}
{\bar{\Delta}^{(b)}+\epsilon}
}
{
2+
\lambda
\frac{\Delta_{ij}}
{\bar{\Delta}^{(b)}+\epsilon}
}.
\end{equation}
Applying the inverse sigmoid function gives
\begin{equation}
m_{ij}^{*}
=
\frac{1}{\alpha}
\log
\left(
1+
\lambda
\frac{\Delta_{ij}}
{\bar{\Delta}^{(b)}+\epsilon}
\right),
\end{equation}
which proves the result.
\end{proof}

\begin{corollary}[Reward-aware preference separation]
\label{cor:rwpo_reward_separation}
Under the confidence-regularized all-pair RWPO objective, the induced
optimal preference margin is monotonically increasing with the reward gap:
\[
\frac{\partial m_{ij}^{*}}
{\partial \Delta_{ij}}
=
\frac{\lambda}
{\alpha(\bar{\Delta}^{(b)}+\epsilon)
\left(
1+
\lambda
\frac{\Delta_{ij}}
{\bar{\Delta}^{(b)}+\epsilon}
\right)}
>0 .
\]
Therefore, RWPO encourages stronger separation between trajectories with
larger reward improvements while avoiding excessive separation for
near-equivalent solutions.
\end{corollary}

Consequently, the all-pair formulation preserves the stabilizing effect
of symmetric pairwise learning while selectively increasing the margin
between trajectories with larger objective improvements.

\subsection{Gradient-Matching Interpretation of Reward-Weighted Preference Optimization}
\label{app:rwpo_gradient_matching}

The analysis below focuses on the reward-gap-weighted preferred-direction component of RWPO, which provides the direct connection to reward-weighted policy-gradient updates. The implemented \emph{confidence-regularized all-pair RWPO loss} additionally includes the symmetric all-pair baseline, which serves as a confidence regularizer by controlling pairwise separation. Since this term does not correspond to a reward-weighted update direction, it is excluded from the coefficient-matching comparison with the policy-gradient formulation.

Fix an instance $x$ and a policy $\pi_\theta(\cdot|x)$. At a given training iteration, let trajectory pairs $(\tau_i,\tau_j)$ be sampled from the current policy (or another fixed sampling distribution). When
differentiating the preference objective, we hold the sampled trajectories and their reward-dependent weights fixed, as in standard PO. 
For analytical tractability, we condition on a fixed realization of the discrete sampling support during differentiation. Equivalently, the sampling policy is treated as locally constant with respect to its support mask, so gradients are computed within a region of parameter space where the selected support does not change. Notably, this condition holds for top-$k$ selective decoding provided the selected support mask remains fixed during the backward pass.

Write
\[
r_i := r(x,\tau_i),
\qquad
\ell_i := \log \pi_\theta(\tau_i|x),
\]
and define the reward and log-probability gaps as
\[
\Delta r := r_i-r_j,
\qquad
\Delta \ell := \ell_i-\ell_j .
\]
Here, $\ell_i$ and $\ell_j$ denote the standard trajectory log-probability. The
length-normalized trajectory score introduced in
Appendix~\ref{app:length_normalization} is not considered in the following
theoretical analysis; instead, we analyze the standard PO/RWPO scoring
function based on trajectory log-probabilities.

The BT preference model induced by the implicit reward
\[
\hat r_\theta(x,\tau)
=
\alpha\log\pi_\theta(\tau|x)
+
\alpha\log Z(x)
\]
defines the pairwise preference probability
\[
p_\theta(\tau_i\succ\tau_j|x)
=
\sigma(
\hat r_\theta(x,\tau_i)
-
\hat r_\theta(x,\tau_j)
)
=
\sigma(\alpha\Delta\ell),
\]
where $\sigma(u)=1/(1+e^{-u})$.

Let $\omega(\tau_i,\tau_j,x)\geq0$ be a pairwise weight that is independent
of $\theta$. For a fixed pair-sampling distribution $q$, the weighted
preference objective is
\begin{equation}
\mathcal L_\omega^q(\theta)
=
\mathbb{E}_{x;(\tau_i,\tau_j)\sim q}
\left[
\mathbb{I}(\Delta r>0)
\omega(\tau_i,\tau_j,x)
\log\sigma(\alpha\Delta\ell)
\right].
\label{eq:Lw_fixed_q}
\end{equation}
During gradient computation, $q$ is treated as fixed and corresponds to
the distribution induced by the sampled rollout pairs.

For comparison, consider the standard expected-reward objective:
\[
J(\theta)
=
\mathbb{E}_{x,\tau\sim\pi_\theta(\cdot|x)}
[r(x,\tau)] .
\]

\subsubsection{Pairwise Representation of the Policy Gradient}

We assume that the reward function is independent of $\theta$, as in
standard policy-gradient optimization.

\begin{lemma}[Pairwise representation of the policy gradient]
\label{lem:pairwise_pg}
For any fixed instance $x$,
\begin{equation}
\nabla_\theta
\mathbb{E}_{\tau\sim\pi_\theta(\cdot|x)}
[r(x,\tau)]
=
\frac{1}{2}
\mathbb{E}_{\tau_i,\tau_j\sim\pi_\theta}
\left[
\Delta r
(
\nabla_\theta\ell_i-
\nabla_\theta\ell_j
)
\right].
\label{eq:pairwise_pg}
\end{equation}
Equivalently, when ties $\Delta r=0$ have zero contribution,
\begin{equation}
\nabla_\theta
\mathbb{E}_{\tau\sim\pi_\theta(\cdot|x)}
[r(x,\tau)]
=
\mathbb{E}_{\tau_i,\tau_j\sim\pi_\theta}
\left[
\mathbb{I}(\Delta r>0)
\Delta r
(
\nabla_\theta\ell_i-
\nabla_\theta\ell_j
)
\right].
\label{eq:pairwise_pg_preferred}
\end{equation}
\end{lemma}

\begin{proof}
The likelihood-ratio identity gives
\[
\nabla_\theta
\mathbb{E}_{\tau\sim\pi_\theta(\cdot|x)}[r(x,\tau)]
=
\mathbb{E}_{\tau\sim\pi_\theta(\cdot|x)}
[
r(x,\tau)\nabla_\theta\ell(\tau)
].
\]
For two independent trajectories,
\begin{align*}
&\mathbb{E}
[
(r_i-r_j)
(
\nabla_\theta\ell_i-
\nabla_\theta\ell_j
)
]
\\
&=
\mathbb{E}[r_i\nabla_\theta\ell_i]
-
\mathbb{E}[r_i\nabla_\theta\ell_j]
-
\mathbb{E}[r_j\nabla_\theta\ell_i]
+
\mathbb{E}[r_j\nabla_\theta\ell_j].
\end{align*}
Because the two trajectories $\tau_i$ and $\tau_j$ are independent and
\[
\mathbb{E}_{\tau\sim\pi_\theta}
[
\nabla_\theta\ell(\tau)
]
=
\nabla_\theta
\sum_\tau\pi_\theta(\tau|x)
=
\nabla_\theta 1
=
0,
\]
based on the assumption that the support of \(\pi_\theta\) does not change when differentiating,
the cross terms vanish:
\[
\mathbb{E}[r_i\nabla_\theta\ell_j]
=
\mathbb{E}[r_i]
\mathbb{E}[\nabla_\theta\ell_j]
=0,
\]
and similarly for
\(\mathbb{E}[r_j\nabla_\theta\ell_i]\).
Therefore,
\[
\mathbb{E}
[
(r_i-r_j)
(
\nabla_\theta\ell_i-
\nabla_\theta\ell_j
)
]
=
2
\mathbb{E}
[
r(x,\tau)\nabla_\theta\ell(\tau)
],
\]
which proves Eq.~\eqref{eq:pairwise_pg}.

For Eq.~\eqref{eq:pairwise_pg_preferred}, define
\[
g(\tau_i,\tau_j)
=
\Delta r
(
\nabla_\theta\ell_i-
\nabla_\theta\ell_j
).
\]
Since both $\Delta r$ and the gradient difference change sign when
exchanging $\tau_i$ and $\tau_j$, we have
\[
g(\tau_i,\tau_j)=g(\tau_j,\tau_i).
\]
Therefore, positive and negative reward-gap pairs contribute equally,
and restricting the expectation to $\Delta r>0$ gives the equivalent
preferred-direction representation.
\end{proof}

\subsubsection{Policy-Score Update Induced by Weighted Preference Optimization}

\begin{lemma}[Gradient of the weighted preference loss]
\label{lem:rwpo_gradient}
For the fixed pair-sampling distribution $q$, the policy-score gradient of
\eqref{eq:Lw_fixed_q} is
\begin{equation}
\begin{aligned}
    &\nabla_\theta \mathcal L_\omega^q(\theta) = \\
    & 
    \alpha\,
    \E_{x,\,(\tau_i,\tau_j)\sim q}
    \left[
    \I(\Delta r>0)\,
    \omega(\tau_i,\tau_j,x)\,
    \sigma(-\alpha\Delta\ell)
    \left(
    \nabla_\theta\ell_i-\nabla_\theta\ell_j
    \right)
    \right].
\end{aligned}
\label{eq:grad_Lw}
\end{equation}
\end{lemma}

\begin{proof}
Let $u=\alpha\Delta\ell$. Since
\[
\frac{d}{du}\log\sigma(u)=\sigma(-u),
\]
we have
\[
\nabla_\theta\log\sigma(\alpha\Delta\ell)
=
\alpha\sigma(-\alpha\Delta\ell)
\left(
\nabla_\theta\ell_i-\nabla_\theta\ell_j
\right).
\]
\end{proof}

\subsubsection{Coefficient-Matching Bound}

The pairwise RL ascent direction in
\eqref{eq:pairwise_pg_preferred} and the preference-learning update in
\eqref{eq:grad_Lw} share the same feature direction
\[
g
:=
\nabla_\theta\ell_1-\nabla_\theta\ell_2.
\]
They differ in the scalar coefficient multiplying this direction.

When the pair-sampling distribution is the current policy,
$q=\pi_\theta\times\pi_\theta$, define
\[
c_{\mathrm{RL}}
:=
\I(\Delta r>0)\Delta r,
\qquad
c_\omega
:=
\alpha\,
\I(\Delta r>0)\,
\omega\,
\sigma(-\alpha\Delta\ell).
\]
Then
\[
\nabla_\theta J(\theta)
=
\E[c_{\mathrm{RL}}g],
\qquad
\nabla_\theta \mathcal L_\omega^q(\theta)
=
\E[c_\omega g].
\]

\begin{theorem}[Coefficient-matching bound]
\label{thm:coefficient_matching}
Assume there exists $G<\infty$ such that
$$
\|g\|\leq G
$$

almost surely. Then
\begin{equation}
\|
\nabla_\theta J(\theta)-\nabla_\theta \mathcal L_\omega^q(\theta)
\|
\leq
G \,
\E\left[
|c_{\mathrm{RL}}-c_\omega|
\right].
\label{eq:grad_bound_L1}
\end{equation}
Moreover,
\begin{equation}
\|
\nabla_\theta J(\theta)-\nabla_\theta \mathcal L_\omega^q(\theta)
\|
\leq
G
\sqrt{
\E\left[
(c_{\mathrm{RL}}-c_\omega)^2
\right]
}.
\label{eq:grad_bound_L2}
\end{equation}
\end{theorem}

\begin{proof}
By the pairwise representations above,
$$
\nabla_\theta J-\nabla_\theta \mathcal L_\omega^q
=
\E[(c_{\mathrm{RL}}-c_\omega)g].
$$

$$
\begin{aligned}
\|\nabla_\theta J-\nabla_\theta \mathcal L_\omega^q\|
&\leq
\E[
|c_{\mathrm{RL}}-c_\omega|\,\|g\|
]\\
&\leq
G\,\E[
|c_{\mathrm{RL}}-c_\omega|
],
\end{aligned}
$$
which proves \eqref{eq:grad_bound_L1}. Applying Cauchy--Schwarz,
$$
\E[|X|]\leq\sqrt{\E[X^2]},
$$
gives \eqref{eq:grad_bound_L2}.
\end{proof}

The bound shows that reward weighting can be interpreted as selecting the
scalar coefficient multiplying each pairwise policy-score direction.
The RL update gives coefficient $\Delta r$ for preferred pairs, whereas
the BT preference update gives
$
\alpha \omega\,\sigma(-\alpha\Delta\ell).
$

Thus, reward-gap weighting directly bridges preference learning and policy gradients. By incorporating $\Delta r$-dependent information into $\omega$, the preference coefficient $\alpha \omega\,\sigma(-\alpha\Delta\ell)$ adapts to the standard RL reward scaling. Consequently, reducing the error term
$$\left\vert{} \Delta r-\alpha \omega \,\sigma(-\alpha\Delta\ell) \right\vert{}$$
steers the resulting update toward the policy-gradient direction, serving as a gradient-matching motivation rather than a strict convergence guarantee."

\subsection{Tilted Sampling}
\label{app:tilted_sampling}

Let \(\pi_{\theta_{\mathrm{base}}}(\tau \mid x)\) denote a pretrained base policy, where \(x\) is an instance and \(\tau\) is a generated trajectory. Following the distinction in \citet{mittal2026beyond}, we first separate task-reward optimization from distribution sharpening at the level of their ideal objectives. Pure \textbf{task-reward} optimization directly seeks trajectories with high task reward:
\[
\pi_\theta^*
=
\arg\max_{\tau} r(x,\tau),
\]
where \(r(x,\tau)\) denotes the task-specific reward. This objective is driven only by solution quality and does not explicitly account for whether the selected trajectory is likely under the pretrained policy.

In contrast, pure \textbf{distribution-sharpening} optimization seeks trajectories that are most likely under the base policy:
\[
\pi_\theta^*
=
\arg\max_{\tau}
\pi_{\theta_{\mathrm{base}}}(\tau \mid x),
\]
or equivalently,
\[
\pi_\theta^*
=
\arg\max_{\tau}
\log \pi_{\theta_{\mathrm{base}}}(\tau \mid x).
\]
This objective does not use task reward directly. Instead, it assumes that the base policy already assigns higher likelihood to better trajectories, and therefore attempts to improve performance by concentrating probability mass on the base model's existing high-likelihood modes.

A softer distributional version of this idea is \textbf{tempered sampling} \citep{hu2024amortizing}. Rather than collapsing onto the single most likely trajectory, tempered sampling defines a sharpened target distribution
\[
q_{\mathrm{temp}}(\tau \mid x)
\propto
\pi_{\theta_{\mathrm{base}}}(\tau \mid x)^{\alpha},
\qquad \alpha > 1.
\]
Equivalently, this is a low-temperature version of the base policy with temperature \(T=1/\alpha<1\):
\[
q_{\mathrm{temp}}(\tau \mid x)
\propto
\pi_{\theta_{\mathrm{base}}}(\tau \mid x)^{1/T}.
\]
Thus, distribution-sharpening optimization and tempered sampling share the same underlying principle: both favor trajectories that are already likely under the base policy. The former corresponds to a hard maximization view, while the latter gives a softened distributional version. However, both approaches improve task performance only when base-policy likelihood is well aligned with solution quality. When this alignment is weak, sharpening may reinforce trajectories that are plausible under the base model but not necessarily high-reward.

\textbf{Tilted sampling} combines the two signals. It keeps the base-policy likelihood as a conservative anchor while also incorporating task reward. The ideal tilted target is
\[
q_{\mathrm{tilt}}(\tau \mid x)
\propto
\pi_{\theta_{\mathrm{base}}}(\tau \mid x)^{\alpha}
\exp\left(\beta^{-1}r(x,\tau)\right).
\]
Note that, $\alpha=1$ corresponds to normal KL-regularized RL fine-tuning \citep{guo2025deepseek}.

Taking logarithms, this target assigns each trajectory a combined score:
\[
\log q_{\mathrm{tilt}}(\tau \mid x)
=
\alpha \log \pi_{\theta_{\mathrm{base}}}(\tau \mid x)
+
\beta^{-1}r(x,\tau)
-
\log z(x),
\]
where \(z(x)\) is the instance-level normalizing constant. Therefore, tilted sampling can be interpreted as a reward-guided form of distribution sharpening: the reward term favors high-quality trajectories, while the base-likelihood term keeps the adapted policy close to trajectories that remain plausible under the pretrained model.
In neural combinatorial optimization, the full trajectory space is combinatorially large, making the exact normalizer \(z(x)\) intractable. We therefore approximate the tilted target over a finite candidate set
\[
\mathcal{C}(x)=\{\tau_1,\ldots,\tau_K\},
\qquad
\tau_k \sim \pi_\theta(\cdot \mid x),
\]
sampled from the current policy. To make the reward and likelihood terms comparable across candidate groups, we normalize them within each group:
\[
\widetilde{r}(x,\tau_k)
=
\frac{r(x,\tau_k)-\mu_r}{\sigma_r+\epsilon},
\]
and
\[
\widetilde{\ell}_{\mathrm{base}}(\tau_k)
=
\frac{
\log \pi_{\theta_{\mathrm{base}}}(\tau_k \mid x)-\mu_{\ell}
}{
\sigma_{\ell}+\epsilon
}.
\]
The finite-candidate tilted score is then defined as
\[
\varsigma_k
=
\lambda_r \widetilde{r}(x,\tau_k)
+
\lambda_p \widetilde{\ell}_{\mathrm{base}}(\tau_k),
\]
where \(\lambda_r\) controls the contribution of the task reward and \(\lambda_p\) controls the strength of the base-policy likelihood anchor. We convert these scores into normalized candidate weights using a softmax temperature \(T_\omega\):
\[
w_k
=
\frac{
\exp(\varsigma_k/T_\omega)
}{
\sum_{j=1}^{K}\exp(\varsigma_j/T_\omega)
}.
\]
The adapted policy is then trained by weighted behavioural cloning:
\[
\mathcal{L}_{\mathrm{tilt}}(\theta)
=
-
\mathbb{E}_{x}
\left[
\sum_{k=1}^{K}
w_k
\log \pi_\theta(\tau_k \mid x)
\right].
\]

This objective can be viewed as a sampled approximation to the ideal tilted distribution. When \(\lambda_r=0\), it reduces to base-policy sharpening or tempered imitation, concentrating probability mass on trajectories that the base policy already considers likely. When \(\lambda_s=0\), it becomes reward-only adaptation over the sampled candidate set. When both terms are active, the objective encourages task-reward improvement while retaining a conservative connection to the pretrained policy. In this sense, tilted sampling provides a controlled interpolation between distribution sharpening and reward-driven adaptation.

\section{Baseline Method Details and Hyperparameters} \label{app:baselines}

To aid reproducibility, we provide further details of the methods compared in Section~4 of the main paper, and the datasets used.

\subsection{Baselines}
\paragraph{Classical solvers.}
For LKH3, we adopt the commonly used configuration in the machine learning literature~\citep{kool2019attention} and implement the TSP and CVRP baselines based on the code from \url{https://github.com/wouterkool/attention-learn-to-route}.  For KP, we use OR-Tools and follow the implementation provided in the official Google OR-Tools documentation: \url{https://developers.google.com/optimization/pack/knapsack}.  For OP, we use Compass~\citep{kobeaga2018efficient}, implemented from \url{https://github.com/gkobeaga/op-solver}, as the main classical baseline.

\paragraph{RL-based constructive solvers.}
We compare against representative RL-based constructive solvers, including AM~\citep{kool2019attention}, POMO~\citep{kwon2020pomo}, ICAM~\citep{zhou2024instance}, MDAM~\citep{xin2021multi}, Sym-NCO~\citep{kim2022sym} and PO \cite{pan2025preference}. For POMO, PO, and ICAM, we use multi-start decoding with $M=N$ and $8\times$ augmentation on non-SCOP instances with size $N\leq 1{,}000$, and multi-start decoding without augmentation for instances with $N>1{,}000$. For SCOPs, we use multi-start decoding with $M=N$ and no augmentation. For Sym-NCO, we adopt 200 augmentations for COP instances with $N\leq 1{,}000$ and 100 augmentations for 2,000-node instances. Unless otherwise specified, we use the pretrained checkpoints released by the original papers for all neural baselines except POMO and PO, which we train under our unified experimental setup.

\paragraph{Neural divide-and-conquer methods.}
We also include neural divide-and-conquer baselines, including GLOP~\citep{ye2024glop} and UDC~\citep{zheng2024udc}, to compare against methods specifically designed for large-scale combinatorial optimization.

\subsection{Dataset} 
\paragraph{OPLib benchmark.}
To evaluate out-of-distribution generalization on real benchmark instances, we use OPLib, a public benchmark library collected from classical OP instance families and released by BCAM Data Science \url{https://github.com/bcamath-ds/OPLib}. The repository provides instance files organized by generation setting, where different generations correspond to different ways of assigning node rewards. In our experiments, we focus on \emph{Generation 3} instances because their reward-generation rule is closest to the reward distribution used to train our neural policy. This choice allows us to evaluate scale and structural generalization on standard benchmark instances while avoiding an additional reward-distribution shift that would confound the comparison. Therefore, performance on Generation 3 primarily reflects whether the learned policy can transfer from synthetic training instances to OPLib geometry and budget structures under a compatible reward distribution.

\subsection{Hyperparameters Setting}
Tab.~\ref{tab:pomo_hyperparameters} presents the parameter settings for the three training frameworks: POMO, PO, and RWPO.
\begin{table*}[t]
\centering
\caption{Hyperparameter setting for POMO, PO, and RWPO.}
\label{tab:pomo_hyperparameters}
\begin{tabular}{lcccc}
\toprule
 & \textbf{CVRP100} & \textbf{OP100} & \textbf{PCTSP100/SPCTSP100} & \textbf{KP200} \\
\midrule
Alpha & 0.03 & 0.05 & 0.05 & 0.05 \\ 
Preference Function & Exponential & BT & BT & BT \\
Length normalization & 0 & 0.3 & 0.1 & 0.3 \\
RWPO Lambda & 0.5 & 0.5 & 1 & 0.5 \\
\midrule
Epochs & 4000 & 1000 & 500 & 500\\
Epochs (FT) & 200 & 200 & 100 & 100\\
Epochs (TS) & 100 & 100 & 100 & 100\\
Epoch (FT) Size& 50000 & 1280000 & 1280000 & 1280000\\
Epoch (TS) Size& 51200 & 51200 & 51200 & 51200\\
Encoder Layer Number & 6 & 6 & 6 & 6\\
Batch Size & 64 & 64 & 64 & 64\\
Embedding Dimension & 128 & 128 & 128 & 128\\
Attention Head Number & 8 & 8 & 8 & 8 \\
Feed Forward Dimension & 512 & 512 & 512 & 512\\
Tanh Clip & 50 & 50 & 50 & 50\\
Learning Rate & 3e-4 & 3e-4 & 3e-4 & 3e-4\\
Small Learning Rate & 3e-5 & 3e-5 & 3e-5 & 3e-5\\
Norm Type & InstanceNorm & LayerNorm & LayerNorm & LayerNorm \\
\bottomrule
\end{tabular}
\end{table*}
The attention layer of RWPO is the same as for POMO and PO.

\section{Further Experimental Details}

We provide further details of the experiments of  Section~4 of the main paper.

\subsection{Soft Entropy Estimation}
\label{app:soft_entropy}
In addition to policy entropy, which is computed directly from the decoder action distribution, we use a representation-level entropy estimator to quantify how information is distributed in the hidden embedding space. Directly estimating the entropy of high-dimensional continuous neural representations is challenging: differential entropy is sensitive to scaling and coordinate transformations, while hard discretization or clustering-based estimators can be computationally expensive and difficult to apply at scale. We therefore adopt the soft entropy estimator of \citet{conklin2026learning}, which can be interpreted as a differentiable relaxation of binning-based Shannon entropy. The method softly quantizes hidden representations into a fixed set of reference directions on the unit hypersphere and then computes the Shannon entropy of the resulting categorical distribution.

Let $Z \in \mathbb{R}^{M \times d}$ denote a collection of hidden representations, where $M$ is the number of representation vectors and $d$ is the hidden dimension. In our experiments, $Z$ can correspond to encoder node embeddings, decoder query embeddings, or other intermediate representations collected over instances and decoding steps. The first step is to normalize each representation vector onto the unit sphere:
\begin{equation}
    \bar{z}_m
    =
    \frac{z_m}{\|z_m\|_2 + \epsilon_{\mathrm{norm}}},
    \qquad
    m=1,\ldots,M,
\end{equation}
where $\epsilon_{\mathrm{norm}}$ is a small constant for numerical stability. We then sample $K$ reference anchors $\{w_k\}_{k=1}^{K}$ uniformly at random from the unit sphere $\mathbb{S}^{d}$. Equivalently, this can be implemented by sampling $\tilde{w}_k \sim \mathcal{N}(0,I_d)$ and normalizing:
\begin{equation}
    w_k
    =
    \frac{\tilde{w}_k}{\|\tilde{w}_k\|_2}.
\end{equation}
Let $W=[w_1,\ldots,w_K]\in\mathbb{R}^{d\times K}$ denote the matrix of anchor directions.

For each normalized representation $\bar{z}_m$, we compute its cosine similarity to all anchors and apply a temperature-scaled softmax:
\begin{equation}
    \check{z}_{m,k}
    =
    \frac{
    \exp\left(\bar{z}_m^\top w_k / T\right)
    }{
    \sum_{j=1}^{K}
    \exp\left(\bar{z}_m^\top w_j / T\right)
    },
    \qquad
    k=1,\ldots,K,
\end{equation}
where temperature parameter $T>0$ controls the sharpness of the soft assignment. A smaller $T$ makes the assignment closer to hard nearest-anchor quantization, while a larger $T$ produces smoother assignments. Thus, each representation vector is mapped to a probability vector $\check{z}_m \in \Delta^{K-1}$ over the anchor set.

The empirical soft-quantized distribution over anchors is obtained by averaging the assignments over all representations:
\begin{equation}
    \hat{p}_k
    =
    \frac{1}{M}
    \sum_{m=1}^{M}
    \check{z}_{m,k},
    \qquad
    k=1,\ldots,K.
\end{equation}
Since $\hat{p}$ is a categorical distribution over $K$ anchors, its Shannon entropy is
\begin{equation}
    \mathcal{H}_{\mathrm{soft}}(Z)
    =
    -
    \sum_{k=1}^{K}
    \hat{p}_k \log \hat{p}_k .
\end{equation}
For easier comparison across different choices of $K$, we report the normalized soft entropy
\begin{equation}
    \bar{\mathcal{H}}_{\mathrm{soft}}(Z)
    =
    \frac{\mathcal{H}_{\mathrm{soft}}(Z)}{\log K},
\end{equation}
which lies in $[0,1]$. A value close to $1$ indicates that the representations are spread relatively evenly across anchor directions, while a smaller value indicates that the representation mass is concentrated around fewer directions.

When representations are collected layer-wise, the same computation can be applied independently to each layer. Let $Z^{(\ell)}$ denote the representations from layer $\ell=1,\ldots,L$. We compute the layer-wise soft entropy $\bar{\mathcal{H}}_{\mathrm{soft}}(Z^{(\ell)})$ and report the average:
\begin{equation}
    \bar{\mathcal{H}}_{\mathrm{soft}}^{\mathrm{model}}
    =
    \frac{1}{L}
    \sum_{\ell=1}^{L}
    \bar{\mathcal{H}}_{\mathrm{soft}}(Z^{(\ell)}).
\end{equation}
In our analysis, we apply this estimator separately to encoder and decoder representations. Encoder soft entropy measures how broadly instance-level information is distributed across representation directions, while decoder soft entropy characterizes the diversity of the decision representation used to construct the solution.

\paragraph{Algorithm.}
The full procedure is summarized below.

\begin{algorithm}[H]
\caption{Soft entropy estimation for hidden representations}
\label{alg:soft_entropy}
\begin{algorithmic}[1]
\REQUIRE Hidden representations $Z=\{z_m\}_{m=1}^{M}$, number of anchors $K$, temperature $\tau$
\STATE Normalize each representation: $\bar{z}_m \leftarrow z_m / (\|z_m\|_2+\epsilon_{\mathrm{norm}})$
\STATE Sample anchors $\tilde{w}_k \sim \mathcal{N}(0,I_d)$ for $k=1,\ldots,K$
\STATE Normalize anchors: $w_k \leftarrow \tilde{w}_k / \|\tilde{w}_k\|_2$
\FOR{$m=1,\ldots,M$}
    \STATE Compute cosine similarities $s_{m,k} \leftarrow \bar{z}_m^\top w_k$
    \STATE Compute soft assignment:
    \[
        \check{z}_{m,k}
        \leftarrow
        \frac{\exp(s_{m,k}/\tau)}
        {\sum_{j=1}^{K}\exp(s_{m,j}/\tau)}
    \]
\ENDFOR
\STATE Aggregate assignments: $\hat{p}_k \leftarrow \frac{1}{M}\sum_{m=1}^{M}\check{z}_{m,k}$
\STATE Compute entropy: $\mathcal{H}_{\mathrm{soft}}(Z) \leftarrow -\sum_{k=1}^{K}\hat{p}_k\log \hat{p}_k$
\STATE Normalize: $\bar{H}_{\mathrm{soft}}(Z) \leftarrow \mathcal{H}_{\mathrm{soft}}(Z)/\log K$
\RETURN $\bar{H}_{\mathrm{soft}}(Z)$
\end{algorithmic}
\end{algorithm}

\paragraph{Soft entropy estimator.}
We employ soft entropy as a representation-level estimator for continuous neural embeddings. Unlike dimension-wise discretization, soft entropy yields a smooth Shannon-entropy estimate without relying on hard binning, thereby circumventing the instability and memory overhead inherent to high-dimensional spaces. Furthermore, compared to clustering-based quantization, it eliminates the need for iterative centroid fitting. Anchor directions are sampled once and reused, reducing the entire estimation process to highly efficient cosine similarities followed by a softmax operation. Crucially, because vectors are $L_2$-normalized prior to estimation, this metric specifically captures the angular diversity of the latent space. This is particularly suitable for Transformer-based models, where high-dimensional embedding geometry and attention mechanisms rely heavily on relationships among continuous vector representations.

\subsection{Interpretation of Representation and Policy Entropy}
In this work, soft entropy complements policy entropy by characterizing
different aspects of the learned solver. Policy entropy measures
behavioural uncertainty in the decoder action distribution,
\[
\mathcal{H}_{\mathrm{policy}}(\pi_t)
=
-\sum_{a\in\mathcal{A}_t}
\pi(a|s_t)\log\pi(a|s_t),
\]
whereas soft entropy measures the diversity of continuous hidden
representations. For encoder representations, soft entropy reflects
whether the model preserves diverse instance-level information before
decoding. For decoder queries, it characterizes the diversity of
decision-state representations used to construct the next action. In
particular, the decoder query
\[
q_t=f(h_{\mathrm{mean}},h_{\mathrm{current}},h_{\mathrm{end}},B_t)
\]
summarizes the current partial solution, global context, depot/end
information, and remaining budget. Therefore, decoder-query soft entropy
measures how diverse or compressed the internal decision state is before
interacting with candidate actions.

This distinction separates two effects: whether the model retains rich
information about the input instance and whether the decoder maintains an
expressive representation of the evolving solution state. High
decoder-query soft entropy should not be interpreted as behavioural
uncertainty; rather, it indicates diversity in the latent decision-state
representation. Our analyses in Section~\ref{sec:entropy_representation}
and Appendix~\ref{app:entropy_representation_diversity} shows that
RWPO preserves richer encoder representations while maintaining structured
decoder-query representations, suggesting that reward-weighted
preference learning improves generalization through more informative
representations.

Decoder hidden-state entropy, as studied in
\cite{conklin2026learning}, provides a complementary measure of
representation compression. However, these hidden states are computed
after the decoder query has interacted with encoded candidate information
through attention layers. They therefore combine decision-state
information with candidate-dependent features and are more dependent on
decoder architecture. We use decoder hidden-state entropy as a secondary
representation metric rather than the primary measure of route-level
decision-state diversity.

Policy entropy remains the most direct measure of behavioural uncertainty,
but it is affected by the size of the feasible action set because its
maximum value scales with $\log|\mathcal{A}_t|$. Therefore, when comparing
action-level uncertainty across states, we additionally consider
normalized or residual entropy to reduce the effect of varying support sizes.

\subsection{Mutual-Information Analysis}
While soft entropy measures representation diversity, it does not reveal
whether this diversity corresponds to task-relevant information. We
therefore complement our entropy analysis with a mutual-information analysis
between learned representations and OP-specific labels (Tab.~\ref{tab:mi_labels}). For encoder representations, we consider static
instance properties, including reward, depot distance, reward--distance
structure, spatial region, and top-rollout membership, to evaluate whether
the model captures reward and geometric information before decoding. For
decoder-query representations, we consider dynamic decision-state
variables, including accumulated reward, current distance, remaining
budget, feasible action count, budget--feasibility interactions, reward
quantiles, and top-versus-bottom trajectory membership. 

This analysis distinguishes meaningful task-aware representations from
merely diffuse features. In particular, higher mutual information between
decoder queries and feasibility or preference-related labels indicates
that representation diversity corresponds to structured decision
information: the policy can maintain flexibility over candidate actions
while still encoding constraints, remaining resources, and solution
quality signals. Therefore, the entropy and mutual-information analyses
together characterize whether RWPO induces informative exploration rather
than unstructured randomness.

\begin{table*}[t]
\centering
\caption{
Task-relevant labels used for mutual-information analysis on OP.
The labels cover static instance properties, dynamic decoding states, and preference-level solution-quality signals.
}
\label{tab:mi_labels}
\small
\begin{tabularx}{\textwidth}{p{0.14\textwidth} p{0.2\textwidth} X}
\toprule
\textbf{Category} & \textbf{Label} & \textbf{Definition and motivation} \\
\midrule

\multirow{6}{=}{\textbf{Encoder-side static instance labels}}
& Depot distance
& $d_{\mathrm{depot}}(i)=d(i,0)$.
Measures the geometric accessibility of node $i$ from the depot. 
Nodes far from the depot is costly to visit even if they have high reward. \\[0.5em]
\cmidrule(lr){2-3}
& Depot feasible
& $\mathbb{I}\!\left(D_t+d(v_t,i)+d(i,0)\leq B\right)$.
Indicates whether node $i$ can be visited while still allowing return to the depot.
This tests whether representations encode feasibility in addition to reward attractiveness. \\[0.5em]
\cmidrule(lr){2-3}
& Node type
& A categorical label distinguishing depot, selected nodes, unvisited candidates, and infeasible nodes.
This tests whether the model separates structurally different roles. \\[0.5em]
\cmidrule(lr){2-3}
& Reward
& $r_i$.
The static prize of node $i$.
Since OP is a selective maximization problem, this is the core objective-relevant node attribute. \\[0.5em]
\cmidrule(lr){2-3}
& Reward--distance ratio
& $\rho_i=\frac{r_i}{d(v_t,i)+\epsilon}$.
Measures reward efficiency relative to travel cost, which is critical when high reward and high travel distance conflict. \\[0.5em]
\cmidrule(lr){2-3}
& Spatial quadrant
& $q_i=\mathrm{Quadrant}(x_i,y_i)$.
A coarse spatial-region label used to test whether encoder embeddings capture geometric clustering. \\

\midrule

\multirow{7}{=}{\textbf{Decoder-side dynamic state and constraint labels}}
& Accumulated reward
& $R_t=\sum_{s=1}^{t} r_{v_s}$.
Measures the objective value collected by the partial route.
The decoder should adapt future choices based on how much reward has already been obtained. \\[0.5em]
\cmidrule(lr){2-3}
& Budget--feasibility interaction
& $(B-D_t)|\mathcal{A}_t|$ or $\frac{B-D_t}{|\mathcal{A}_t|+\epsilon}$.
Combines remaining budget with the number of feasible actions, capturing how constrained the current decision state is. \\[0.5em]
\cmidrule(lr){2-3}
& Current distance
& $D_t=\sum_{s=1}^{t} d(v_{s-1},v_s)$.
Measures consumed travel budget and determines how much flexibility remains. \\[0.5em]
\cmidrule(lr){2-3}
& Decoding phase
& $\phi_t=\frac{t}{T_{\mathrm{end}}}$.
Indicates whether the policy is in the early, middle, or late construction stage.
Good OP policies should behave differently across these phases. \\[0.5em]
\cmidrule(lr){2-3}
& Feasible actions
& $|\mathcal{A}_t|$.
Measures the size of the valid action set at step $t$.
High mutual information with this label indicates awareness of the current feasible decision space. \\[0.5em]
\cmidrule(lr){2-3}
& Remaining budget
& $B_t^{\mathrm{rem}}=B-D_t$.
Captures the remaining travel resource and determines whether the policy should continue exploring, select nearby nodes, or terminate. \\[0.5em]
\cmidrule(lr){2-3}
& Selected count
& $C_t=t$ or $C_t=|\{v_1,\ldots,v_t\}|$.
Tracks route length and construction progress, which affects both accumulated reward and feasibility. \\

\midrule

\multirow{2}{=}{\textbf{Preference-level solution-quality labels}}
& Top-rollout membership
& For encoder analysis, 
$y_i^{\mathrm{top}}=\mathbb{I}(i\in\tau^{\mathrm{top}})$, where $\tau^{\mathrm{top}}$ denotes a top-performing rollout or the union of nodes visited by top rollouts.
For decoder analysis,
$y_m^{\mathrm{top/bottom}}=\mathbb{I}(R(\tau_m)\geq Q_{1-q}(R))$.
This tests whether representations encode information associated with globally high-quality solutions. \\[0.5em]
\cmidrule(lr){2-3}
& Reward quantile
& $q_m=\mathrm{QuantileBin}(R(\tau_m))$.
This assigns each sampled rollout to a discrete final-reward quantile.
Unlike binary top-versus-bottom membership, it provides a graded preference signal and tests whether decoder states encode relative solution quality. \\

\bottomrule
\end{tabularx}
\end{table*}

\section{Additional Experimental Results}

We provide further experiments, not reported in Section~4 of the main paper.

\subsection{Comparison with Existing Algorithms on CVRP} \label{app:CVRP}
\begin{table*}[t]
\centering
\caption{Comparison on instances across different scales. Best results are highlighted in bold.}
\label{tab:cvrp_results}
\resizebox{\textwidth}{!}{
\begin{tabular}{llccc | l | ccc ccc ccc}
\toprule
 & \multirow{2}{*}{Methods} 
 & \multicolumn{3}{c|}{$N=100$} 
 &
 & \multicolumn{3}{c}{$N=500$} 
 & \multicolumn{3}{c}{$N=1{,}000$} 
 & \multicolumn{3}{c}{$N=2{,}000$} \\
\cmidrule(lr){3-5} 
\cmidrule(lr){6-6} 
\cmidrule(lr){7-9} 
\cmidrule(lr){10-12} 
\cmidrule(lr){13-15}
 &  & Obj. $\downarrow$ & Gap & Time
 &  & Obj. $\downarrow$ & Gap & Time 
    & Obj. $\downarrow$ & Gap & Time 
    & Obj. $\downarrow$ & Gap & Time \\
\midrule
\midrule

\multirow{10}{*}{\rotatebox{90}{CVRP}}
& LKH3$^*$
& \textbf{15.65} & 0.00\% & --
&
& \textbf{37.23} & 0.00\% & 1.2m
& 46.40 & 7.91\% & 4m
& 64.90 & 8.15\% & 14m \\
\cmidrule(lr){2-15}

& BQ
& 16.09 & 2.79\% & 2.2m
& \multirow{6}{*}{\rotatebox{90}{OOD}}
& 38.44 & 3.25\% & 1.5m
& 44.17 & 2.72\% & 13m
& 62.59 & 4.30\% & 30m \\

& POMO
& 15.75 & 0.66\% & 2.3m
& 
& 44.64 & 19.90\% & 2.5m
& 52.97 & 23.18\% & 19m
& 92.34 & 53.87\% & 2.2h \\

& PO
& 15.80 & 0.94\% & 4.4m
&
& 43.76 & 17.53\% & 2.9m
& 62.69 & 45.78\% & 20m
& 124.87 & 108.09\% & 2.2h \\

& PO+LS
& 15.82 & 1.11\% & 4.5m
&
& 44.20 & 18.73\% & 3.0m
& 63.95 & 48.71\% & 21m
& 116.63 & 94.35\% & 2.2h \\

& RWPO
& 15.79 & 0.90\% & 4.4m
&
& 42.48 & 14.10\% & 2.9m
& 59.49 & 38.35\% & 20m
& 115.98 & 93.26\% & 2.2h \\

& RWPO+LS
& 15.81 & 1.04\% & 4.5m
&
& 42.95 & 15.37\% & 3.0m
& 62.51 & 45.36\% & 22m
& 134.67 & 124.42\% & 2.2h \\
\cmidrule(lr){2-15}

& RWPO+TS
& \cellcolor{gray!20}15.75 & 0.61\% & 4.4m
& \multirow{3}{*}{\rotatebox{90}{ID}}
& 38.14 & 2.45\% & 3m
& 44.89 & 4.40\% & 20m
& 64.34 & 7.21\% & 2.2h \\

& ICAM
& 15.87 & 1.41\% & 1.9m
&
& \cellcolor{gray!20}37.49 & \cellcolor{gray!20}0.70\% & 1.3m
& 43.07 & 0.16\% & 27m
& 61.34 & 2.22\% & 4h \\

& UDC
& -- & -- & --
&
& 37.99 & 2.04\% & 2.3h
& \cellcolor{gray!20}\textbf{43.00}
& \cellcolor{gray!20}0.00\% & 10h
& \cellcolor{gray!20}\textbf{60.01}
& \cellcolor{gray!20}0.00\% & 18h \\
\bottomrule
\end{tabular}
}
\footnotesize{\\For each problem, the upper block reports out-of-distribution (OOD) results, evaluating the zero-shot scale generalization of the base models, while the lower block reports in-distribution (ID) results after scale-specific adaptation. For BQ (larger model), POMO (training 30500 epochs), ICAM, and UDC (augmentation x200), we use the official pre-trained models from the original papers. For POMO, PO, ICAM, and RWPO, we apply the augmentation (x8) proposed in \citep{kwon2020pomo}.}
\end{table*}

Tab.~\ref{tab:cvrp_results} evaluates zero-shot scale generalization (OOD) and target-scale adaptation (ID) on CVRP instances. In the OOD setting, traditional neural solvers suffer severe performance degradation as instance size increases; specifically, the optimality gaps of PO and POMO compound rapidly at $N \ge 500$. By anchoring preference updates to the true reward gap, RWPO successfully mitigates this degradation. This advantage becomes more pronounced at larger scales: compared to standard PO, RWPO reduces the optimality gap from $17.53\%$ to $14.10\%$ at $N=500$, from $45.78\%$ to $38.35\%$ at $N=1{,}000$, and from $108.09\%$ to $93.26\%$ at $N=2{,}000$.

Following scale-specific adaptation with tilted sampling, RWPO+TS achieves performance comparable to dedicated large-scale solvers. By restricting optimality gaps to $2.45\%$, $4.40\%$, and $7.21\%$ on $N \in \{500, 1000, 2000\}$ instances, respectively, RWPO+TS drastically improves upon the directly transferred base policies. These results indicate that reward-weighted preference learning provides a highly robust initialization for scale adaptation, which TS subsequently refines. Finally, in the ID setting, RWPO+TS remains strictly competitive with specialized methods such as ICAM and UDC, yielding near-optimal performance without relying on the extensive data augmentation pipelines required by those baselines.

\subsection{Comparison Among Fine-tuning Methods}
\subsubsection{Comparison Between Local Search Fine-tuning with Scale-specific Finetuning and Tilted Sampling} \label{app:comparison_ls_scale_ft_ts}
\begin{figure*}[t]
    \centering

    \begin{subfigure}[t]{0.98\textwidth}
        \centering
        \includegraphics[width=\linewidth]{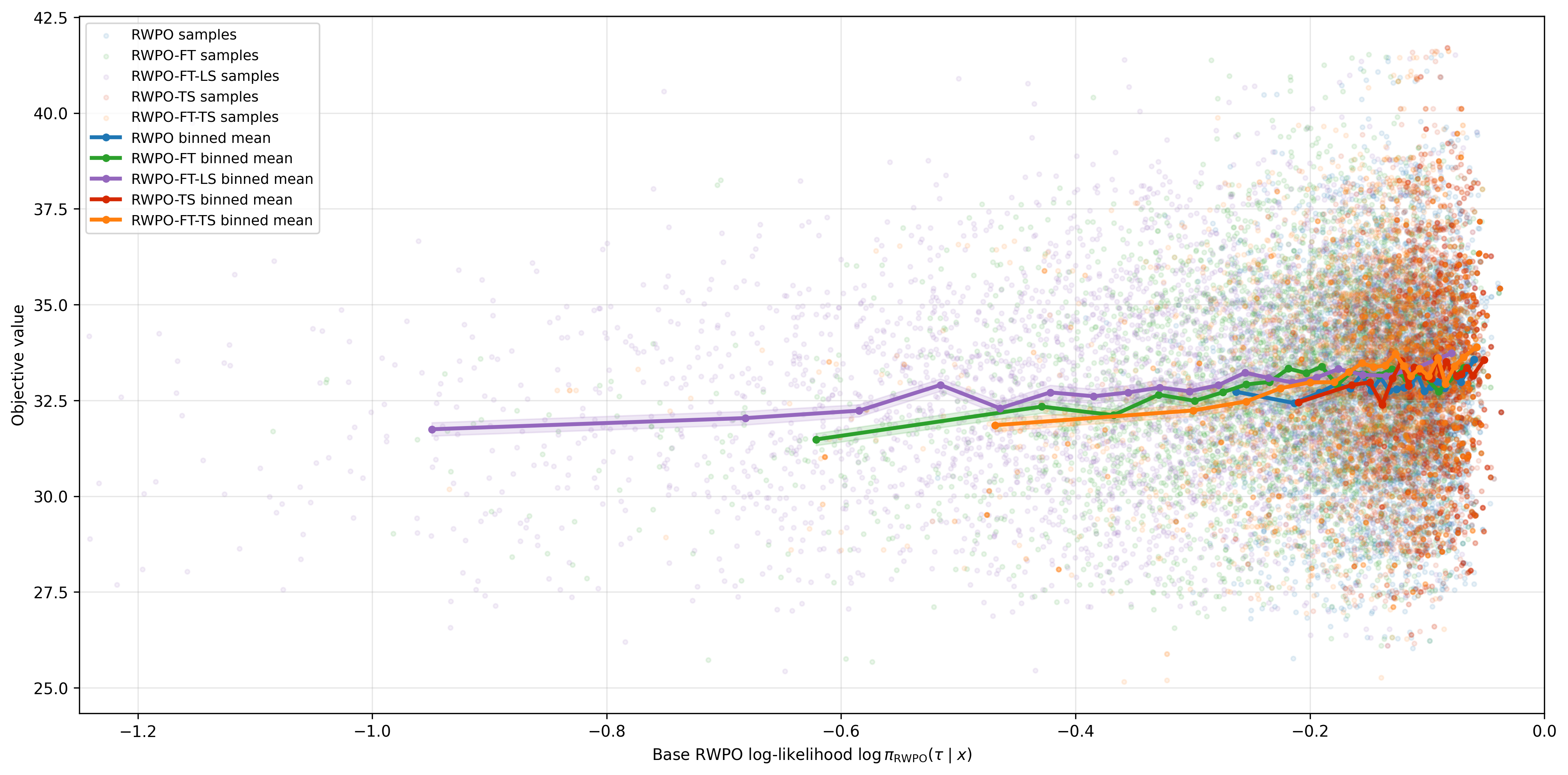}
        \caption{$N=100$: same-scale adaptation}
        \label{fig:likelihood_quality_frontier_n100}
    \end{subfigure}

    \vspace{0.5em}

    \begin{subfigure}[t]{0.98\textwidth}
        \centering
        \includegraphics[width=\linewidth]{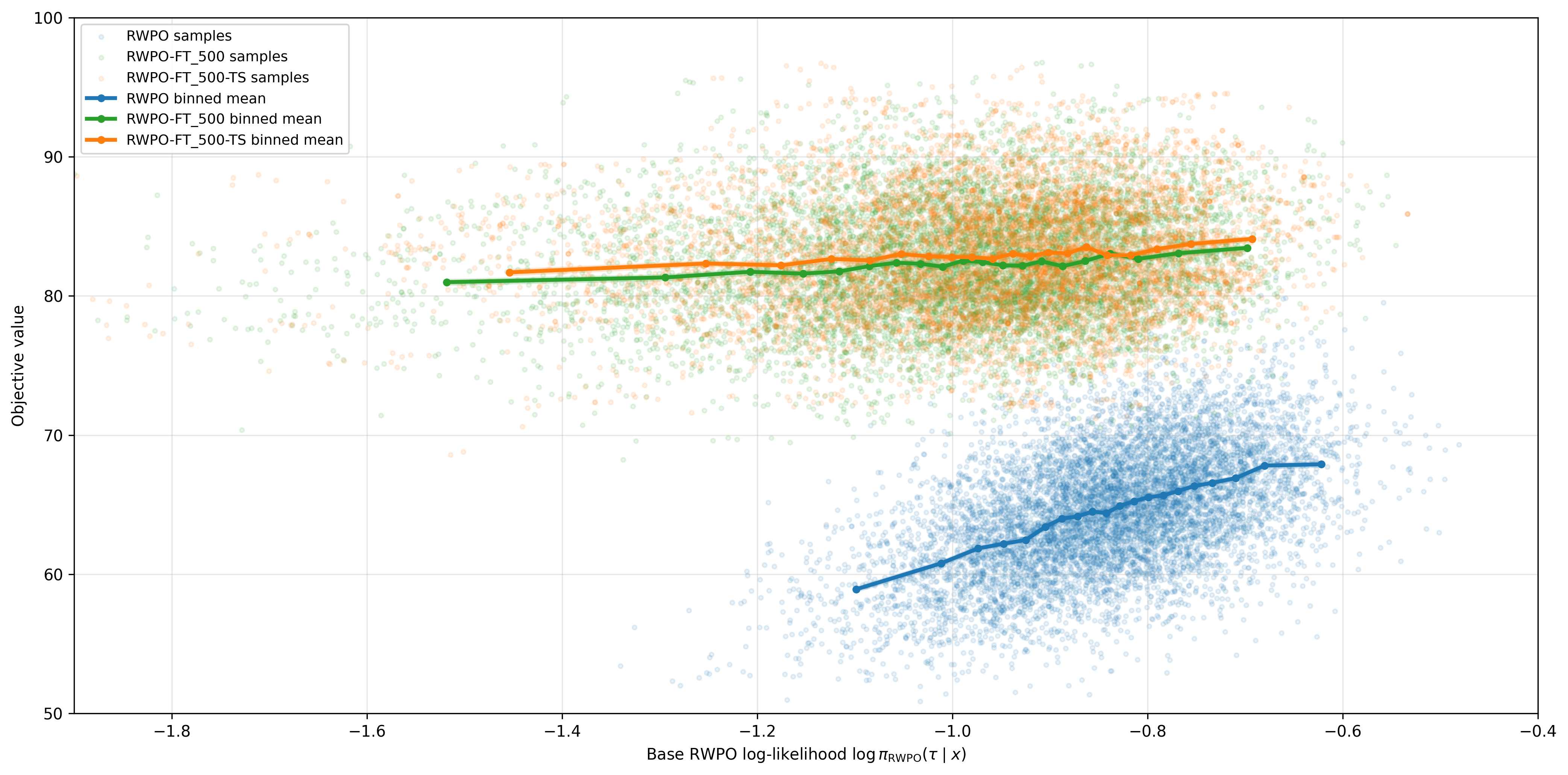}
        \caption{$N=500$: large-scale adaptation}
        \label{fig:likelihood_quality_frontier_n500}
    \end{subfigure}

    \caption{
    Length-normalized likelihood--quality frontier under the base RWPO policy (1k epochs). Each point represents a sampled trajectory, plotted by its length-normalized log-likelihood under the base policy (x-axis) versus its objective value (y-axis). Solid curves indicate binned mean objectives. On $N=100$, scale-specific fine-tuning (FT) and tilted sampling (TS) shift the distribution toward substantially higher-quality solutions while retaining high likelihood under the base policy, whereas local search (LS) fine-tuning induces a significantly larger distribution shift. On $N=500$, the frontier flattens and trajectories tightly cluster in the high-likelihood region; nevertheless, RWPO-based adaptation successfully elevates the upper quality envelope. These dynamics confirm that FT performs a conservative probability redistribution among already-plausible trajectories—avoiding drastic shifts into low-likelihood regions—while TS effectively concentrates the probability mass of the base policy.
    }
    \label{fig:likelihood_quality_frontier}
\end{figure*}
The likelihood--quality frontiers in Fig.~\ref{fig:likelihood_quality_frontier} reveal a scale-dependent adaptation pattern. Each point represents a sampled trajectory, where the horizontal axis is the length-normalized log-likelihood assigned by the base RWPO policy and the vertical axis is the objective value. A higher log-likelihood, i.e., a value closer to zero, indicates that the trajectory is more probable and more consistent with the base RWPO policy's existing preferences. A lower log-likelihood, i.e., a more negative value, indicates that the trajectory is less likely under the base RWPO policy and therefore lies farther from the policy's original high-probability region. Note that in this section, the base RWPO represents the model trained without the small learning-rate continued (scale-specific) post-training; RWPO-FT represents the model with continued post-training.  

On the \(N=100\) setting (Fig.~\ref{fig:likelihood_quality_frontier_n100}), all adapted policies remain concentrated in regions with relatively high likelihood under the base RWPO policy, suggesting that post-training improves trajectories largely within regions already covered by RWPO. It is interesting that RWPO-FT has a lower log-likelihood under the base RWPO policy, while achieving substantially higher objective values. This suggests that the performance gain is not simply due to sharpening the original RWPO distribution or concentrating probability mass on trajectories that the base policy already assigns high likelihood. Instead, RWPO-FT appears to increase probability on trajectories that are less preferred by the original RWPO policy but achieve higher objective values.
In contrast, on the \(N=500\) setting (Fig.~\ref{fig:likelihood_quality_frontier_n500}), RWPO-FT substantially shifts the sampled solutions to a much higher objective region, while the tilted variant further improves the binned quality curve slightly. 
This indicates that, under larger-scale distribution shift, high-quality trajectories are not necessarily those with the highest likelihood under the original RWPO policy. 


Comparing the two fine-tuning strategies, RWPO-FT-LS samples extend farther toward lower base RWPO likelihood values, while
RWPO-FT-TS samples are more concentrated near the high-likelihood region. In the likelihood--quality frontier, RWPO-FT-LS generates many trajectories with lower likelihood under the base RWPO policy, suggesting that LS guidance encourages exploration of solutions less represented in the original RWPO sampling distribution. This suggests that LS-guided fientuning can improve trajectories that were assigned relatively low probability by the base RWPO policy. However, this larger deviation from the base likelihood distribution may indicate a larger policy shift, which could require additional evaluation for robustness.
In contrast, RWPO-FT-TS remains closer to the high-likelihood region of the base RWPO policy. Its reward term encourages improvement, while the base-likelihood term anchors the adapted policy to trajectories that remain plausible under the original model. As a result, RWPO-FT-TS behaves as a more conservative reward-guided redistribution mechanism. Therefore, \textit{RWPO-FT-TS appears to induce a smaller shift relative to the base policy, whereas LS-guided fine-tuning explores a broader region of the trajectory space.} These observations further support the illustration in Fig.~\ref{fig:rwpo_pipeline} in the main text, which depicts the evolution of the policy probability distribution during adaptation.

\subsubsection{Fine-tuning does not degrade generalization ability.}
\begin{table*}[t]
\centering
\caption{Effect of fine-tuning on larger-scale generalization on OP. Relative differences are computed using mean tour revenue, where higher is better.}
\label{tab:finetuning_generalization}
\resizebox{\textwidth}{!}{
\begin{tabular}{ll l c c c c}
\toprule
\textbf{Fine-tuned model} 
& \textbf{Reference model} 
& \textbf{OOD scales} 
& \textbf{Mean $\Delta$ (\%)} 
& \textbf{Mean $|\Delta|$ (\%)} 
& \textbf{Max loss (\%)} 
& \textbf{Better/equal (\%)} \\
\midrule
RWPO-FT
& RWPO 
& 200, 500, 1000 
& -0.57 
& 0.66 
& 2.06 
& 22.86 \\

RWPO-FT-LS 
& RWPO-FT
& 200, 500, 1000 
& -0.28 
& 1.40 
& 2.68 
& 34.29 \\

RWPO-FT-TS 
& RWPO-FT 
& 200, 500, 1000 
& +0.63 
& 0.63 
& 0.00 
& 100.00 \\

RWPO-FT-TS 
& RWPO-FT-LS 
& 200, 500, 1000 
& +0.94            
& 1.57 
& 2.14                  
& 74.29 \\

RWPO-FT$_{200}$ 
& RWPO-FT 
& 500, 1000 
& +6.71 
& 6.72 
& 0.15 
& 96.15 \\

RWPO-FT$_{500}$ 
& RWPO-FT 
& 1000 
& +9.42 
& 12.83 
& 6.35 
& 66.67 \\

RWPO-FT$_{500}$-TS 
& RWPO-FT$_{500}$  
& 1000 
& +1.54             
& 1.54 
& 0.00
& 100.00 \\
\bottomrule
\end{tabular}
}
\footnotesize{
$\Delta$ denotes the relative change in mean tour revenue between the fine-tuned model and its reference model. 
Mean $\Delta$ captures the average signed change, while Mean $|\Delta|$ measures the average magnitude of change regardless of direction. 
Max loss reports the largest performance degradation, and Better/equal reports the percentage of cases where the fine-tuned model matches or outperforms the reference model.
}
\end{table*}

Tab.~\ref{tab:finetuning_generalization} evaluates whether post-training adaptation improves or harms larger-scale generalization. Continued post-training on the original scale (RWPO-FT) only slightly changes OOD performance relative to the base RWPO model, with a small average drop of $0.57\%$, suggesting that low-learning-rate refinement is largely conservative but does not by itself substantially improve transfer. LS-guided fine-tuning also shows limited OOD benefit and can introduce larger variation, indicating possible sensitivity to the fine-tuning distribution. In contrast, RWPO-TS consistently improves over RWPO-FT, achieving a positive mean gain of $0.63\%$ with no observed loss and a $100\%$ better/equal rate, showing that tilted sampling provides a safer adaptation mechanism. Scale-specific fine-tuning is more effective for large-scale transfer: fine-tuning on $N=200$ improves performance on $N=500$ and $N=1000$ by $6.71\%$ on average, while fine-tuning on $N=500$ improves transfer to $N=1000$ by $9.42\%$. Applying tilted sampling after $N=500$ fine-tuning further improves performance by $1.54\%$ with no degradation. Overall, the table shows that \textit{RWPO supports stable post-training adaptation, and that the most reliable pipeline combines scale-specific fine-tuning with tilted sampling}.

\subsection{Representation Diversity under Scale Generalization} \label{app:entropy_representation_diversity}

\paragraph{Route-level visualization of decoder-query representation diversity.}
To complement the aggregate entropy statistics, we visualize representative decoded OP routes with visited nodes colored by decoder-query soft entropy. Decoder-query soft entropy measures the diversity of the decoder's latent decision-state representation rather than the uncertainty of the next-action distribution. A high value indicates that the decoder maintains a richer representation of the current partial solution, potentially preserving diverse information about accumulated reward, remaining budget, feasibility constraints, and future trade-offs; a low value indicates a more compressed decision-state representation. Therefore, decoder-query soft entropy provides a representation-level view that complements policy entropy, which directly measures behavioral uncertainty in the next-action distribution. These route-level visualizations connect the numerical representation statistics to the actual spatial decision process: in OP, a policy must decide not only which high-reward regions to visit but also when to move between regions under a limited travel budget. Visualizing decoder-query soft entropy along the constructed route allows us to inspect whether the decision-state representation changes in a spatially structured manner, for example, whether it becomes more concentrated within locally coherent route segments and more diverse near transitions between candidate regions.
 
\begin{figure*}[t]
    \centering

    \begin{subfigure}[t]{0.49\textwidth}
        \centering
        \includegraphics[width=\linewidth]{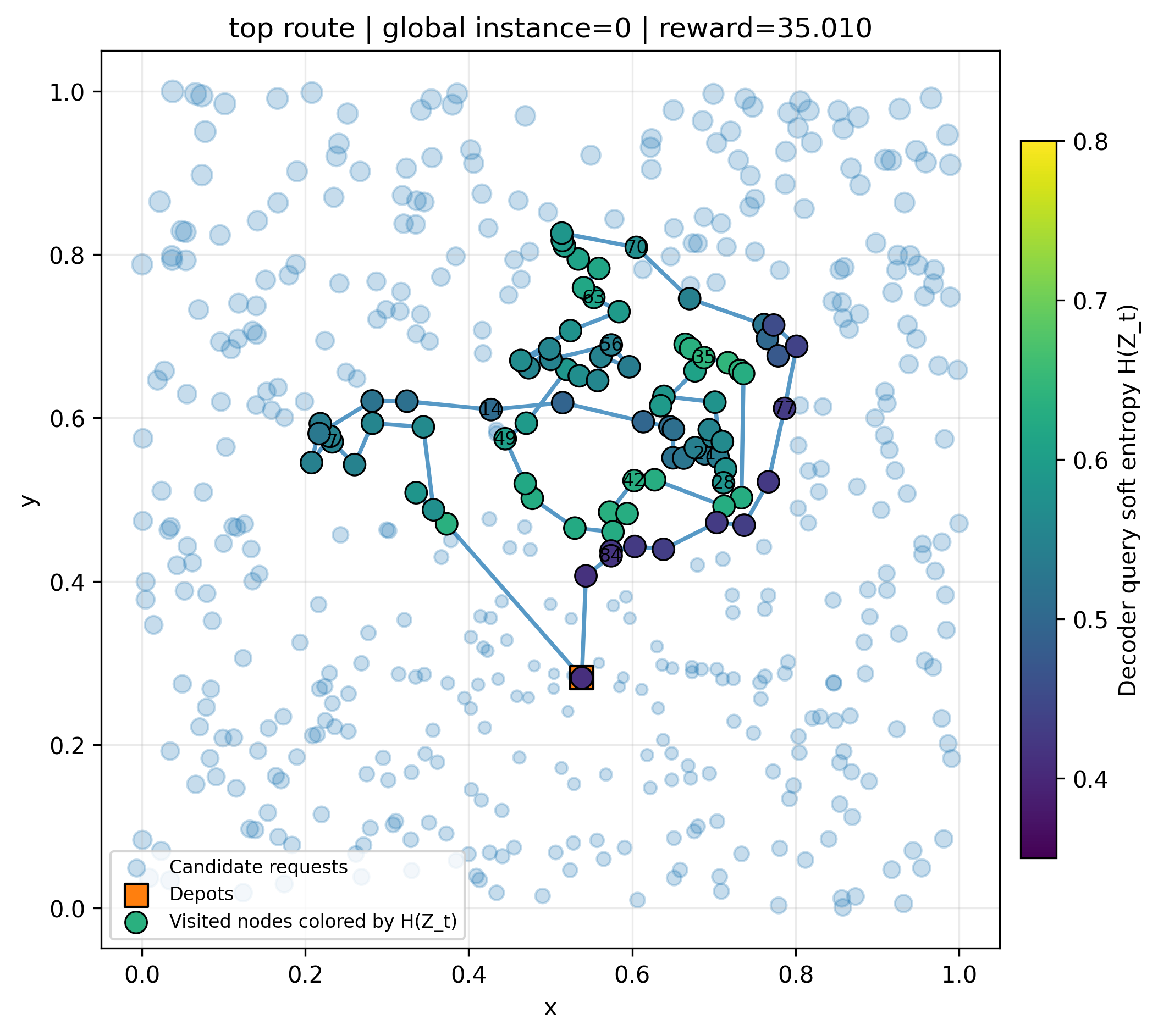}
        \caption{PO}
        \label{fig:route_entropy_inst0_po}
    \end{subfigure}
    \hfill
    \begin{subfigure}[t]{0.49\textwidth}
        \centering
        \includegraphics[width=\linewidth]{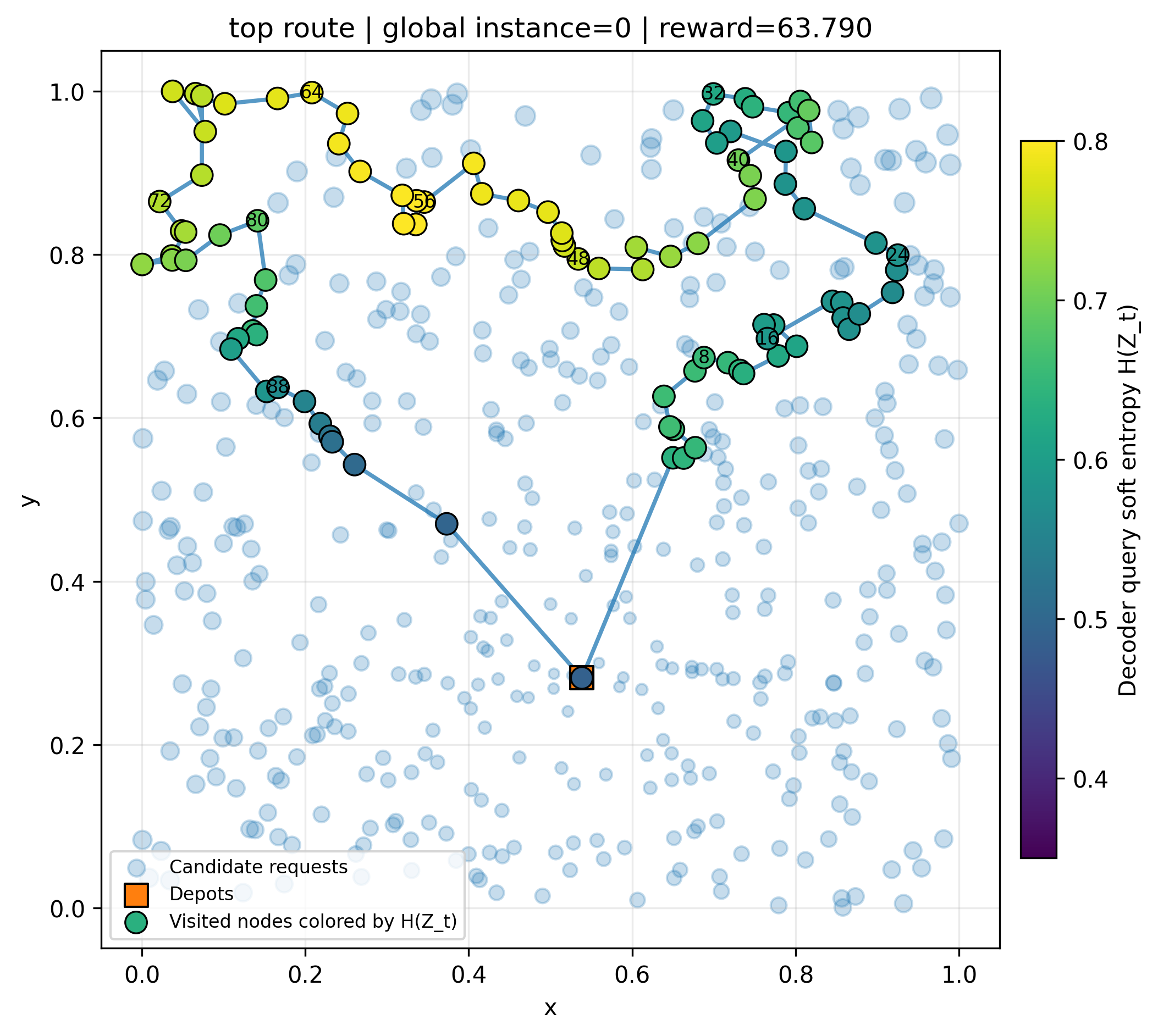}
        \caption{POMO}
        \label{fig:route_entropy_inst0_rl}
    \end{subfigure}

    \vspace{0.5em}

    \begin{subfigure}[t]{0.49\textwidth}
        \centering
        \includegraphics[width=\linewidth]{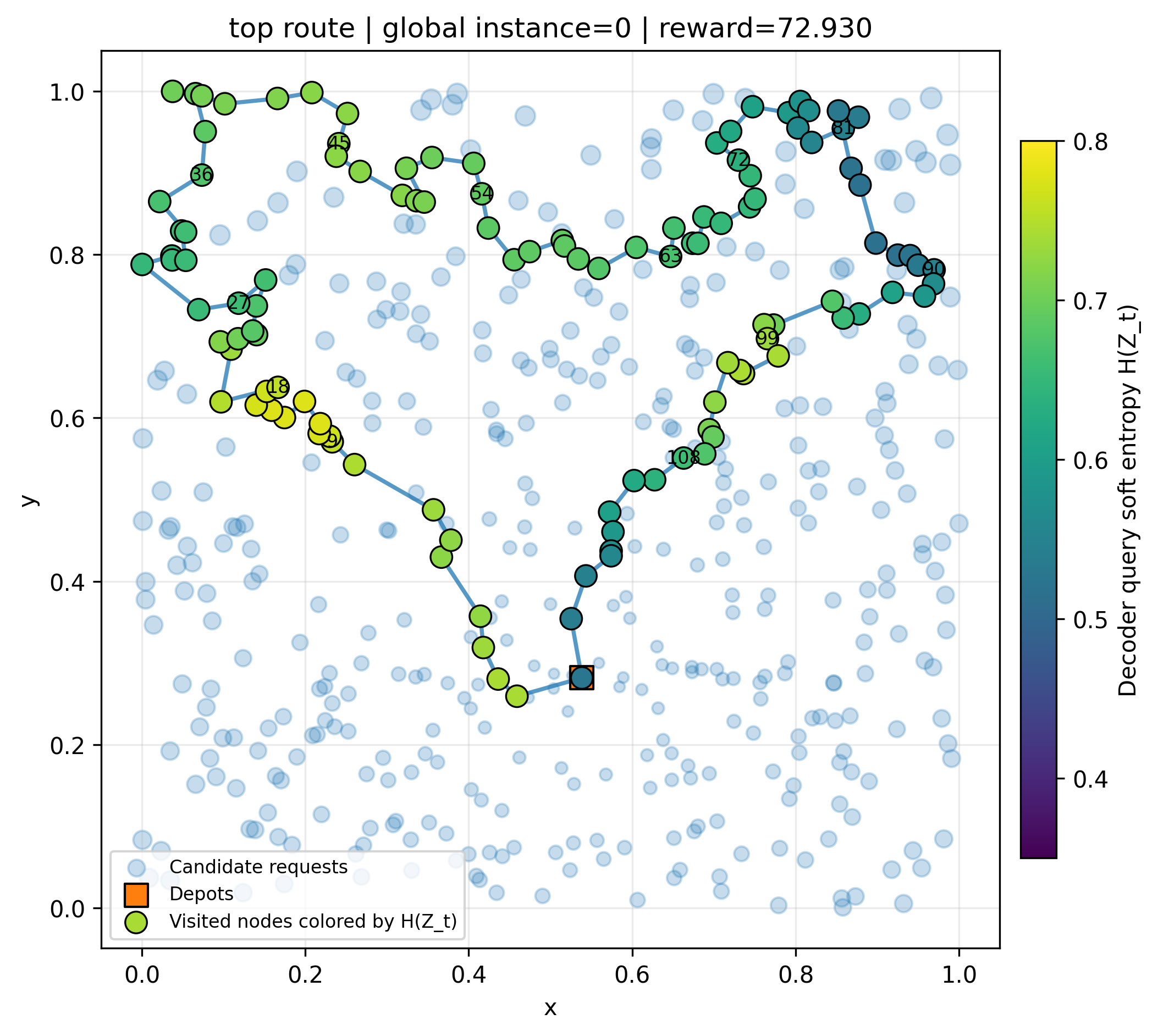}
        \caption{RWPO}
        \label{fig:route_entropy_inst0_rwpo}
    \end{subfigure}
    \hfill
    \begin{subfigure}[t]{0.49\textwidth}
        \centering
        \includegraphics[width=\linewidth]{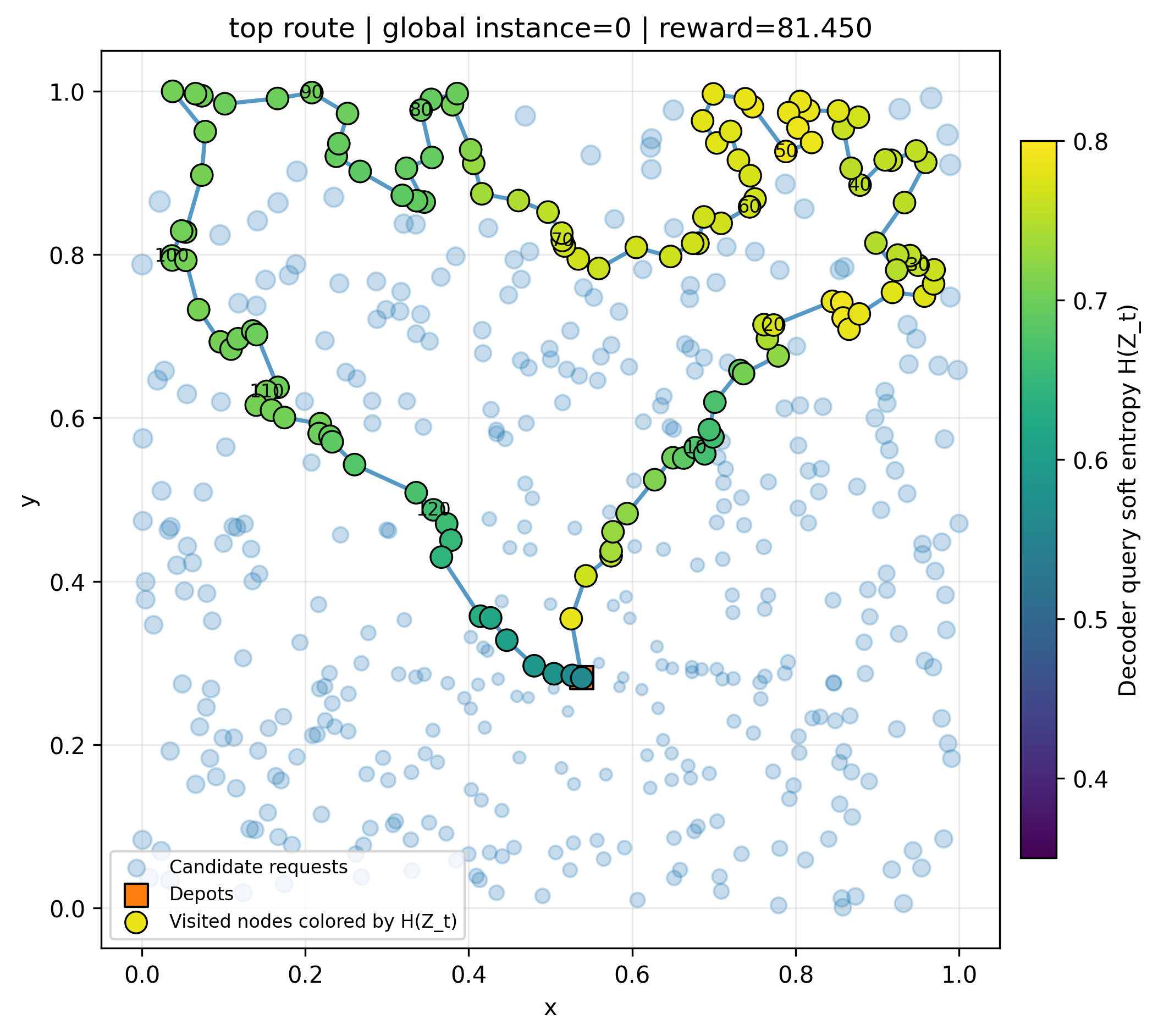}
        \caption{RWPO-FT$_{500}$}
        \label{fig:route_entropy_inst0_rwpo_ft500}
    \end{subfigure}

    \caption{
    Route-level decoder-query soft entropy visualization on a representative OP instance with $N=500$.
    Candidate nodes are shown in light blue, selected routes are shown as connected paths, and visited nodes are coloured by decoder-query soft entropy $\mathcal{H}(Z_t)$.
    These plots provide qualitative evidence of how different methods modulate uncertainty along the constructed route.
    }
    \label{fig:route_entropy_inst0}
\end{figure*}

\begin{figure*}[t]
    \centering

    \begin{subfigure}[t]{0.49\textwidth}
        \centering
        \includegraphics[width=\linewidth]{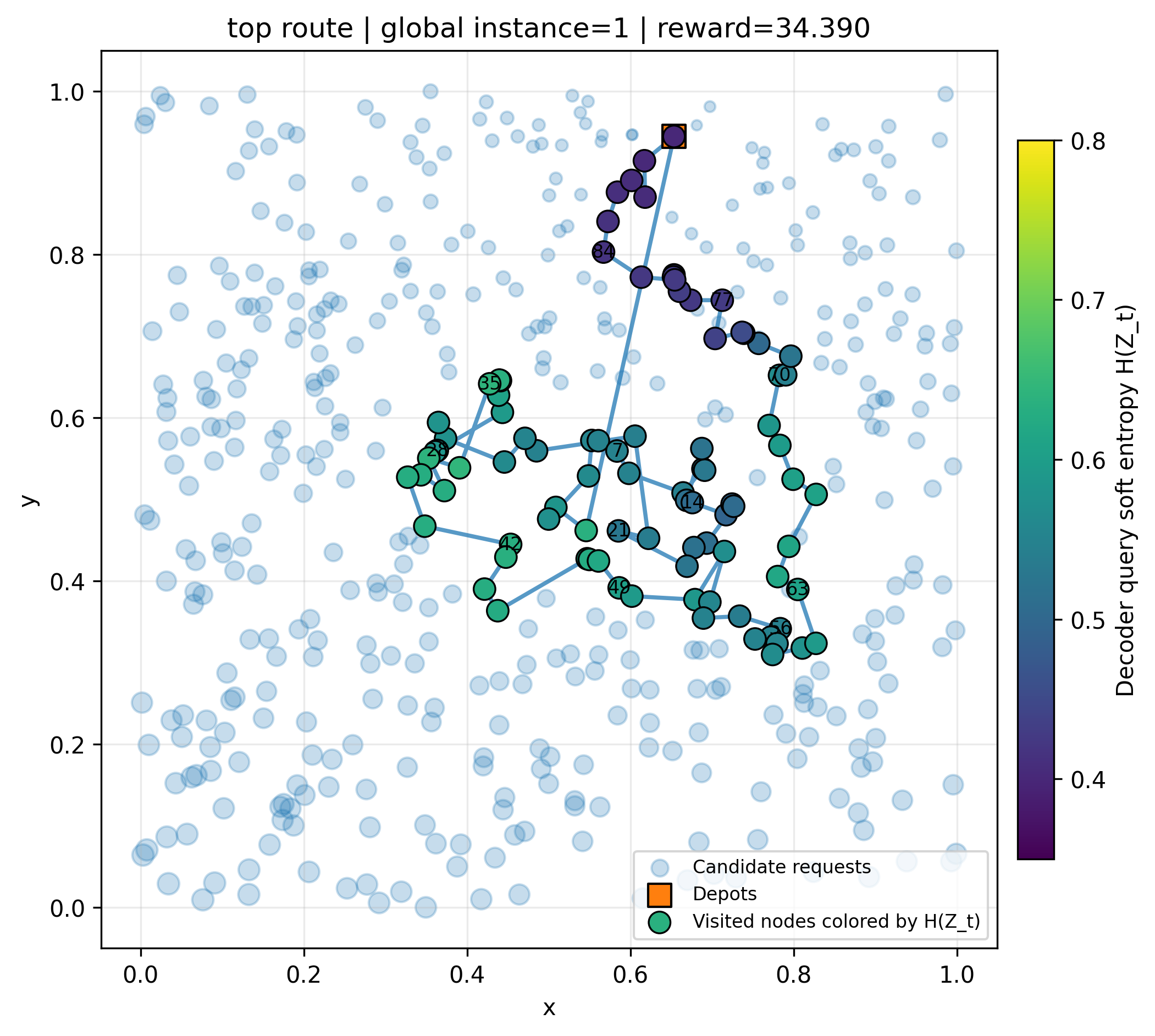}
        \caption{PO}
        \label{fig:route_entropy_inst2_po}
    \end{subfigure}
    \hfill
    \begin{subfigure}[t]{0.49\textwidth}
        \centering
        \includegraphics[width=\linewidth]{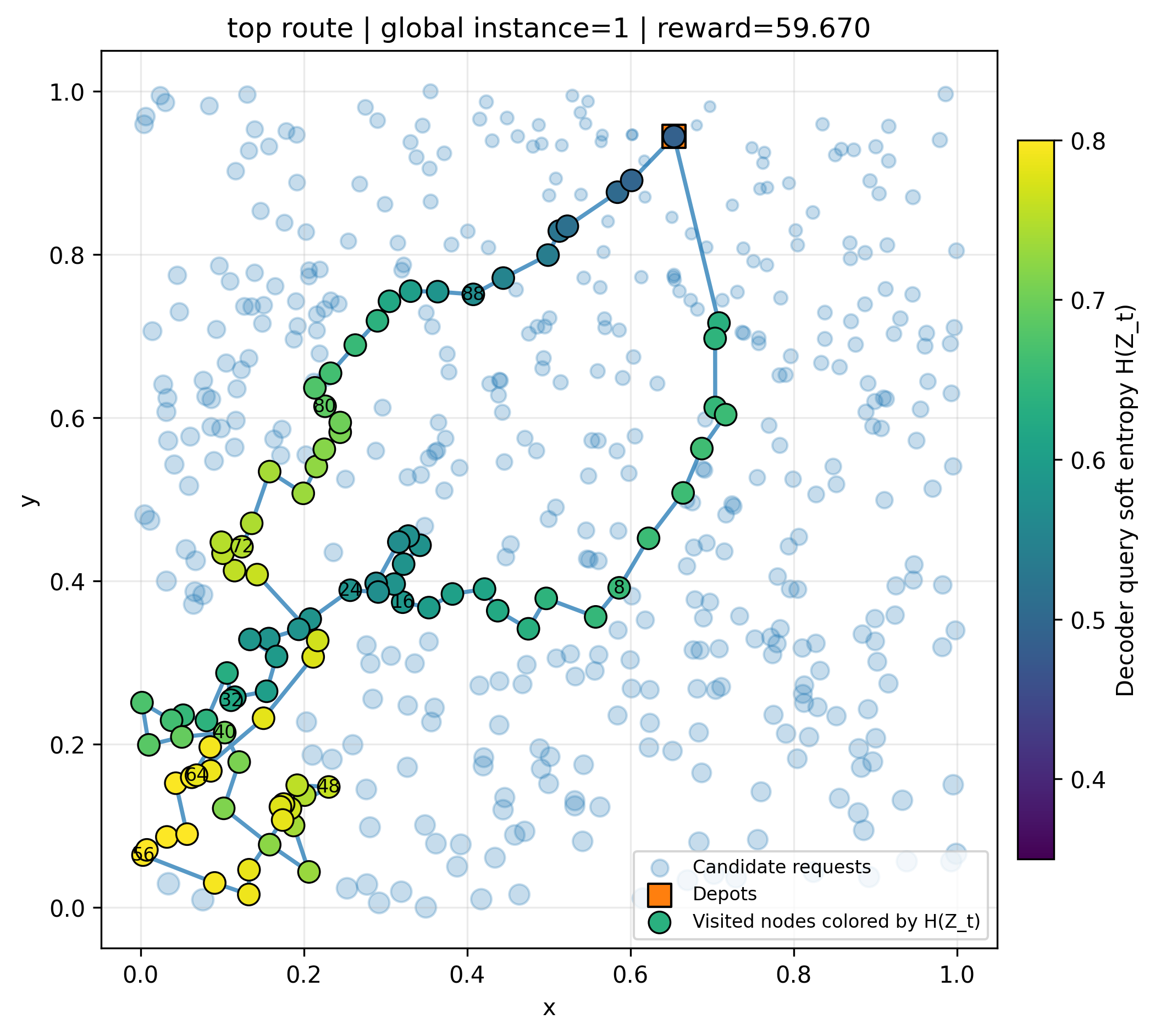}
        \caption{POMO}
        \label{fig:route_entropy_inst2_rl}
    \end{subfigure}

    \vspace{0.5em}

    \begin{subfigure}[t]{0.49\textwidth}
        \centering
        \includegraphics[width=\linewidth]{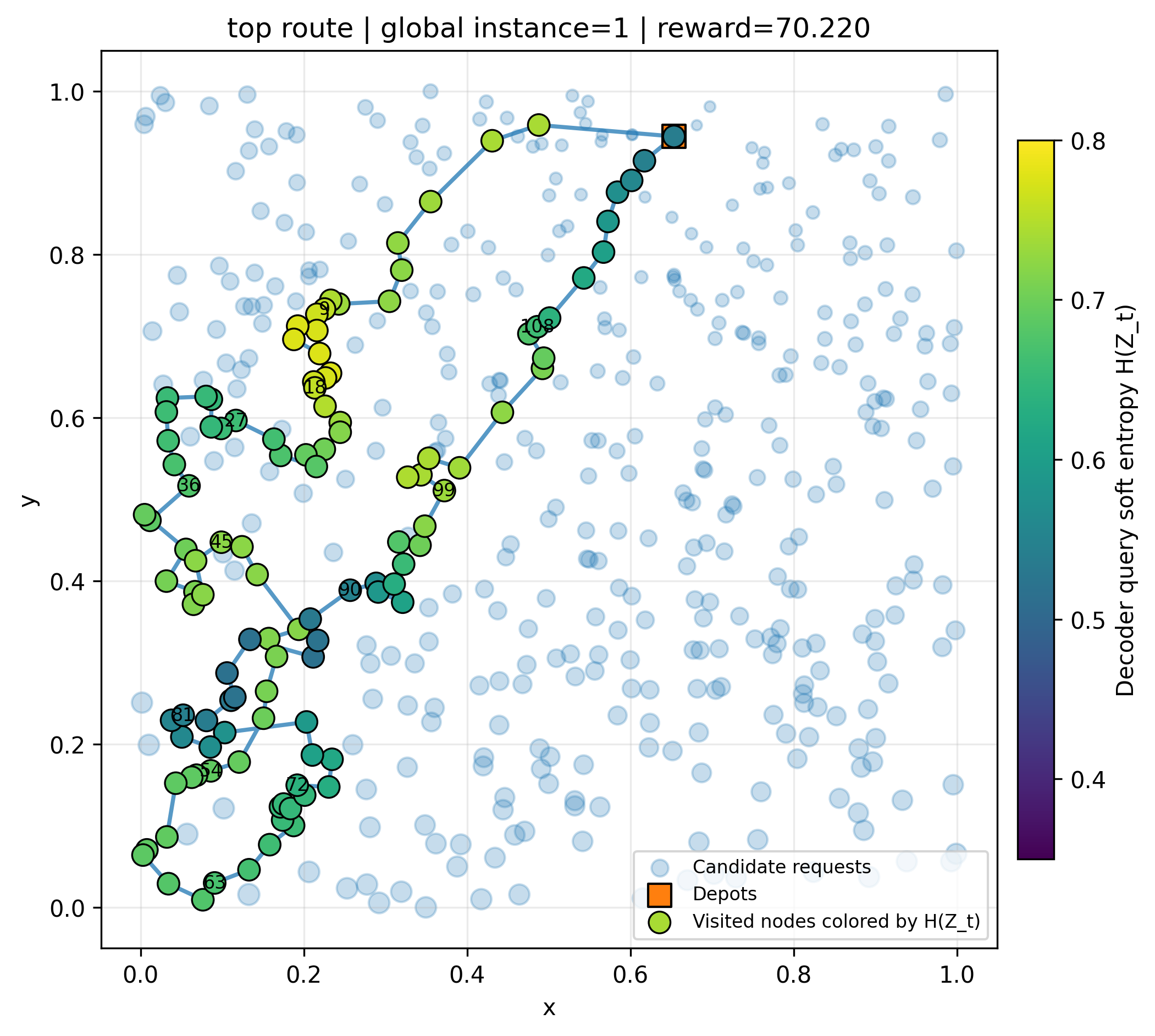}
        \caption{RWPO}
        \label{fig:route_entropy_inst2_rwpo}
    \end{subfigure}
    \hfill
    \begin{subfigure}[t]{0.49\textwidth}
        \centering
        \includegraphics[width=\linewidth]{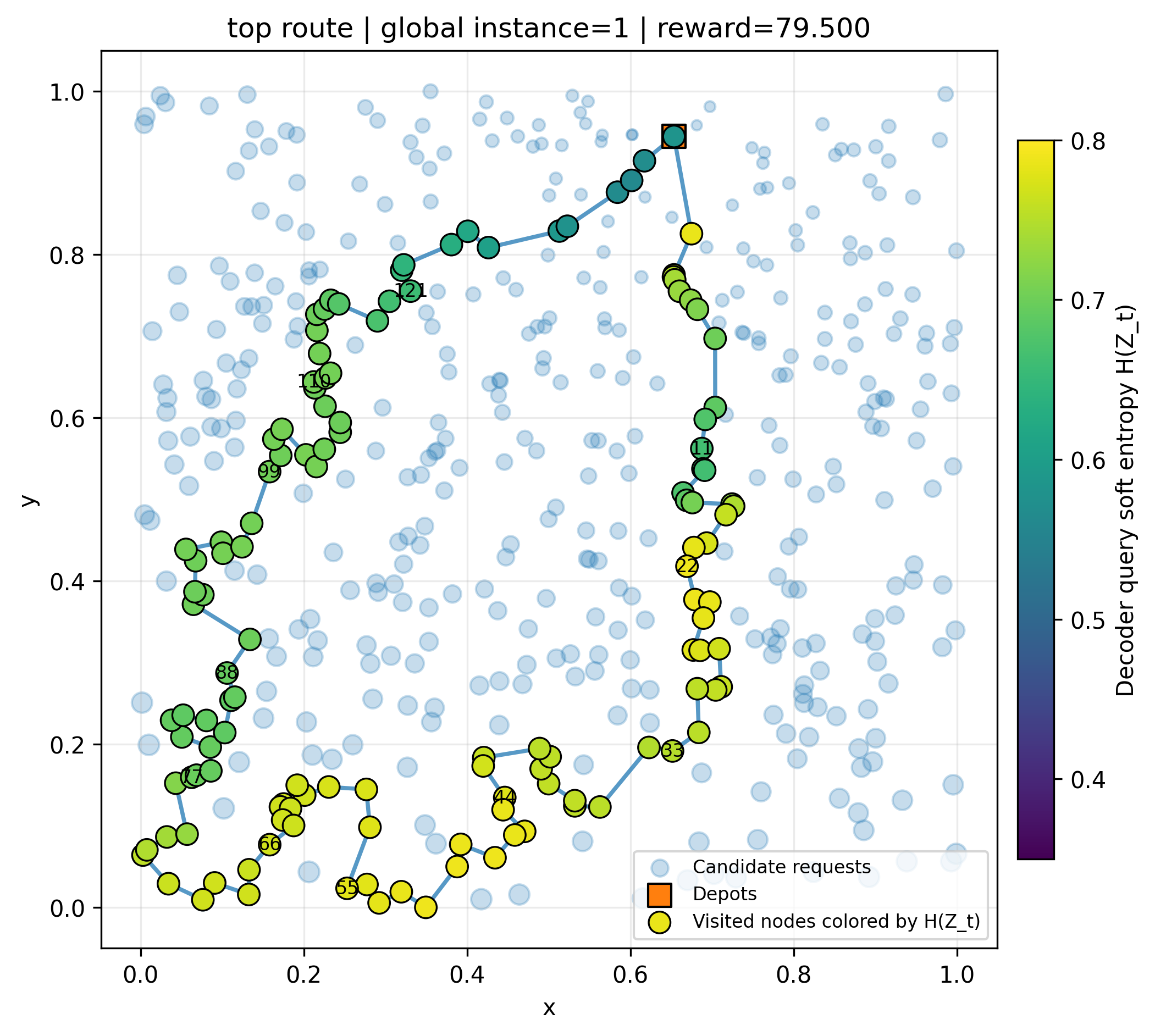}
        \caption{RWPO-FT$_{500}$}
        \label{fig:route_entropy_inst2_rwpo_ft500}
    \end{subfigure}

    \caption{
    Route-level decoder-query soft entropy visualization on a second representative OP instance with $N=500$.
    The comparison illustrates how entropy evolves spatially along the selected route and whether uncertainty is concentrated near local route segments or transitions between regions.
    }
    \label{fig:route_entropy_inst2}
\end{figure*}

Figures~\ref{fig:route_entropy_inst0} and~\ref{fig:route_entropy_inst2} show that the methods differ substantially in how decoder-query representation diversity is distributed along the route. 
The route-level entropy patterns are consistent with the inductive biases of the corresponding training objectives. PO, trained only to preserve pairwise preference order, lacks reward-gap information and therefore learns a weaker reward--geometry representation under scale shift; this is reflected by collapsed decoder-query soft entropy, confinement to low-reward regions, inefficient first moves, and crossed routes. POMO, trained by policy-gradient reward maximization, can move toward high-reward regions, but its representation diversity increases mainly after reaching those regions, suggesting that it is better at local exploitation than early global region selection. In contrast, RWPO weights preference updates by reward differences, so large objective gaps receive stronger learning signals. This encourages the decoder to preserve richer decision-state representations during early route-shaping decisions, when the policy must choose which high-reward region to approach. Scale-specific fine-tuning further adapts this representation to larger candidate sets: RWPO-FT$_{500}$ maintains high decoder-query diversity through the transition into high-reward regions and then becomes more concentrated near route completion, leading to fewer crossings and better geometric route quality. Overall, the visualization supports the view that \textit{reward-weighted preference learning improves OP generalization by aligning representation diversity with reward magnitude, feasibility, and route structure}.

\subsection{Mutual Information Between Hidden Representations and Task-Relevant Labels on OP}\label{app:mutual_infomation}
\paragraph{Scale-dependent division of labor in encoder and decoder representations.}

To examine whether RWPO's improved generalization is accompanied by a different allocation of information between the encoder and decoder, we estimate the soft mutual information between intermediate representations and labels describing the OP instance and route-construction state. The encoder labels include static instance-level attributes, such as depot distance, depot feasibility, node type, reward, reward--distance, spatial quadrant, and top-rollout membership. The decoder labels describe the evolving partial route, including accumulated reward, budget--feasibility interaction, current distance, decoding phase, feasible actions, remaining budget, and selected-node count. 

\begin{figure*}[t]
    \centering

    \begin{subfigure}[t]{0.49\textwidth}
        \centering
        \includegraphics[width=\linewidth]{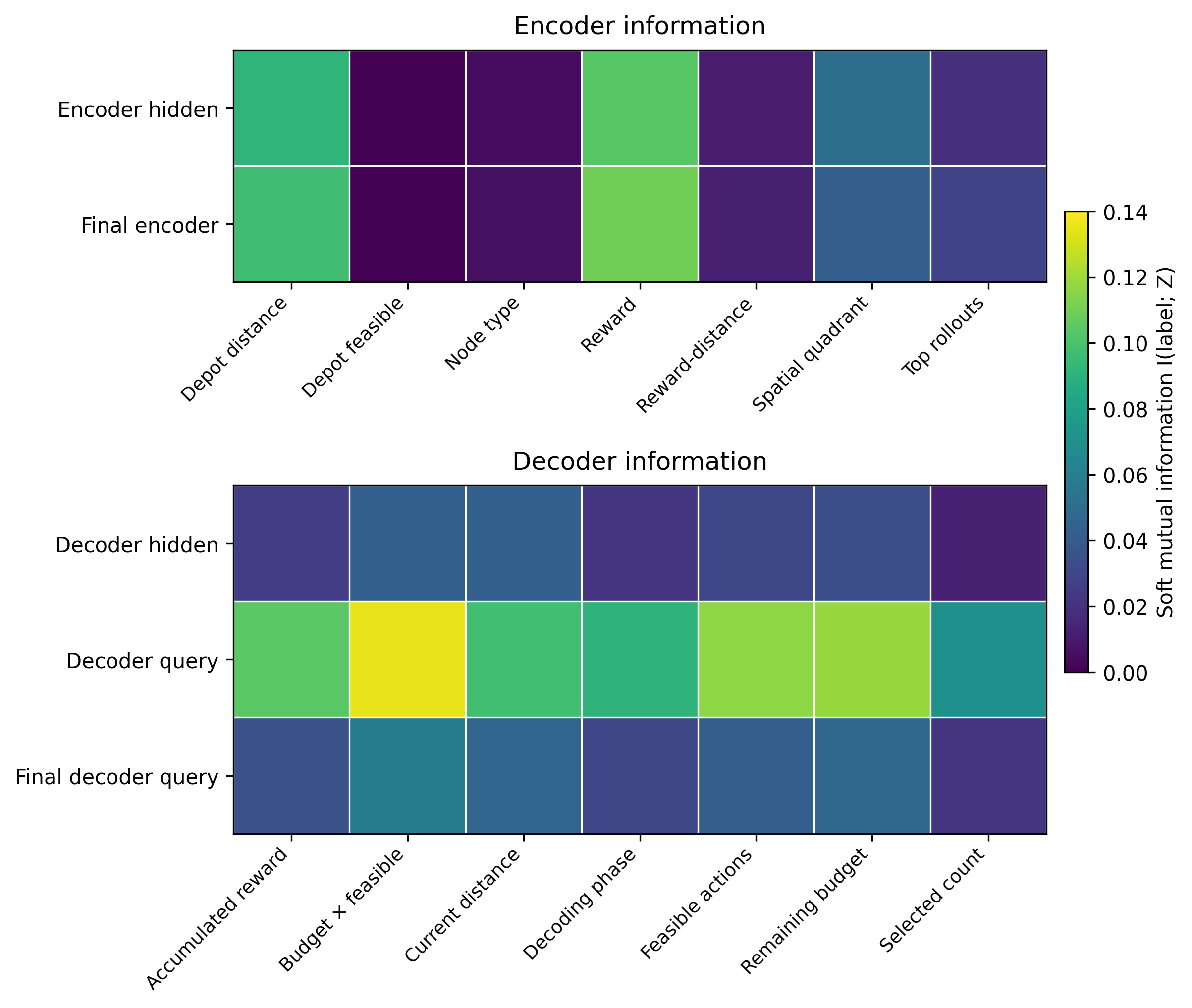}
        \caption{PO}
        \label{fig:info_heatmap_po_n100}
    \end{subfigure}
    \hfill
    \begin{subfigure}[t]{0.49\textwidth}
        \centering
        \includegraphics[width=\linewidth]{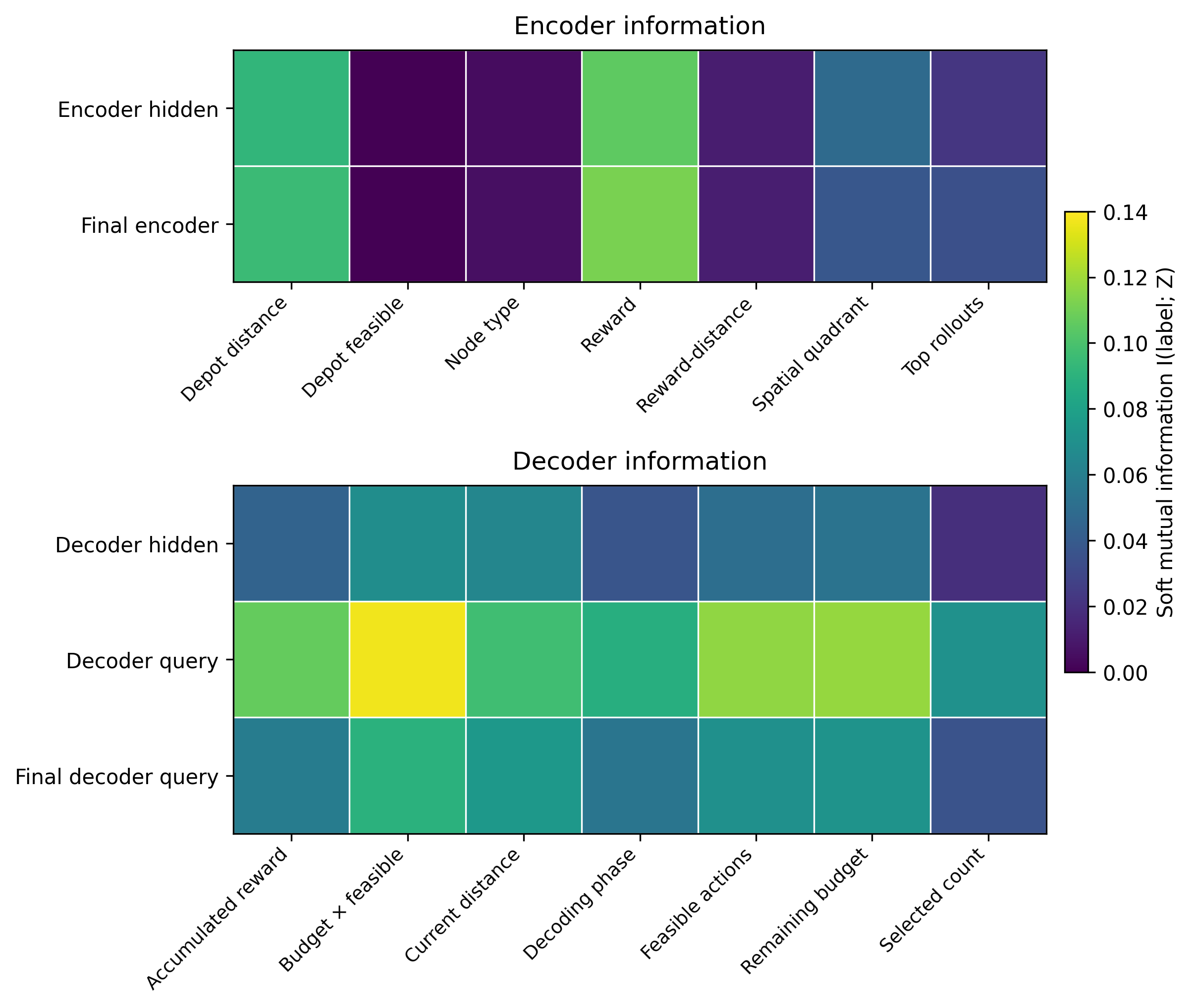}
        \caption{POMO}
        \label{fig:info_heatmap_rl_n100}
    \end{subfigure}

    \vspace{0.5em}

    \begin{subfigure}[t]{0.49\textwidth}
        \centering
        \includegraphics[width=\linewidth]{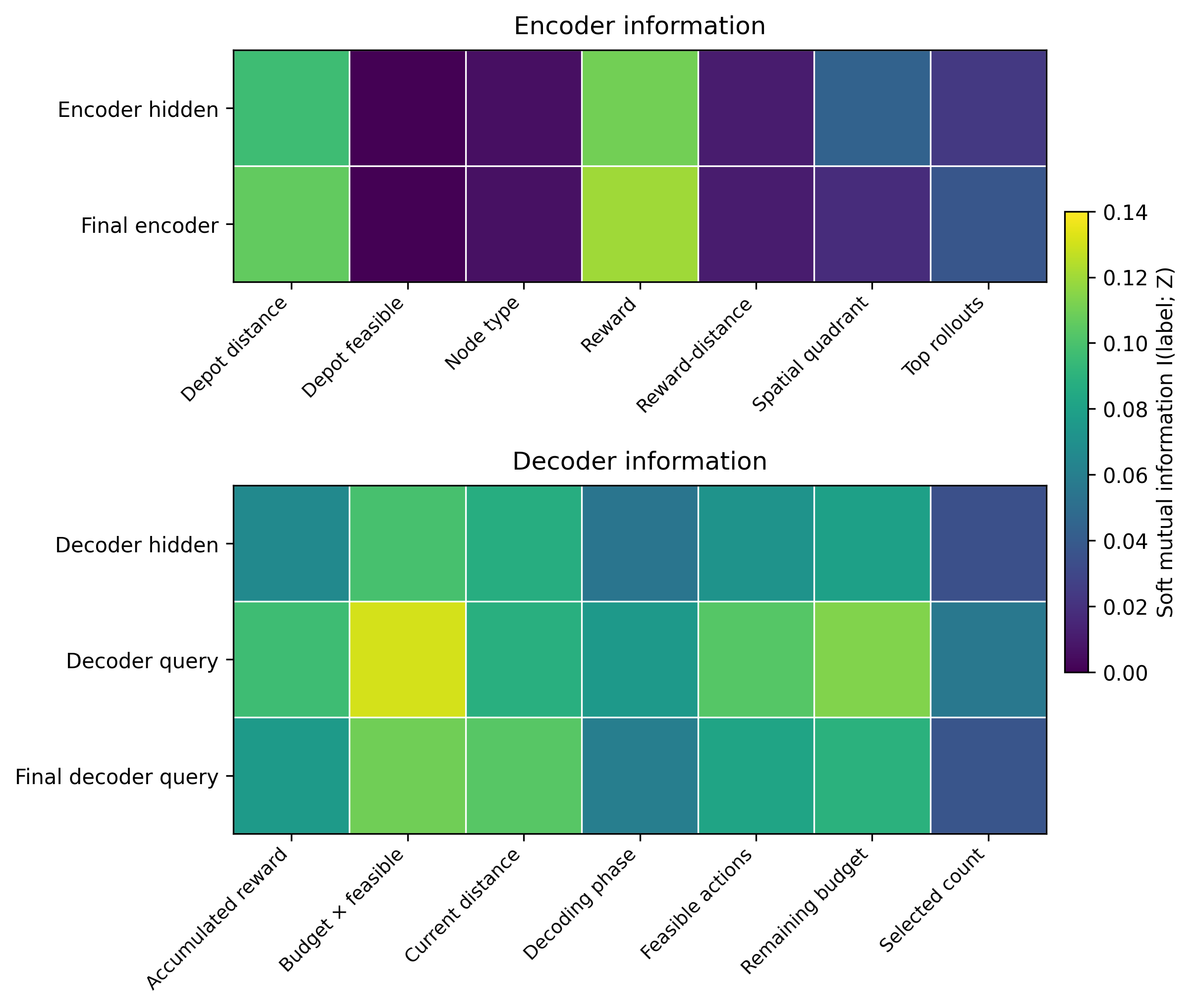}
        \caption{RWPO}
        \label{fig:info_heatmap_rwpo_n100}
    \end{subfigure}
    \hfill
    \begin{subfigure}[t]{0.49\textwidth}
        \centering
        \includegraphics[width=\linewidth]{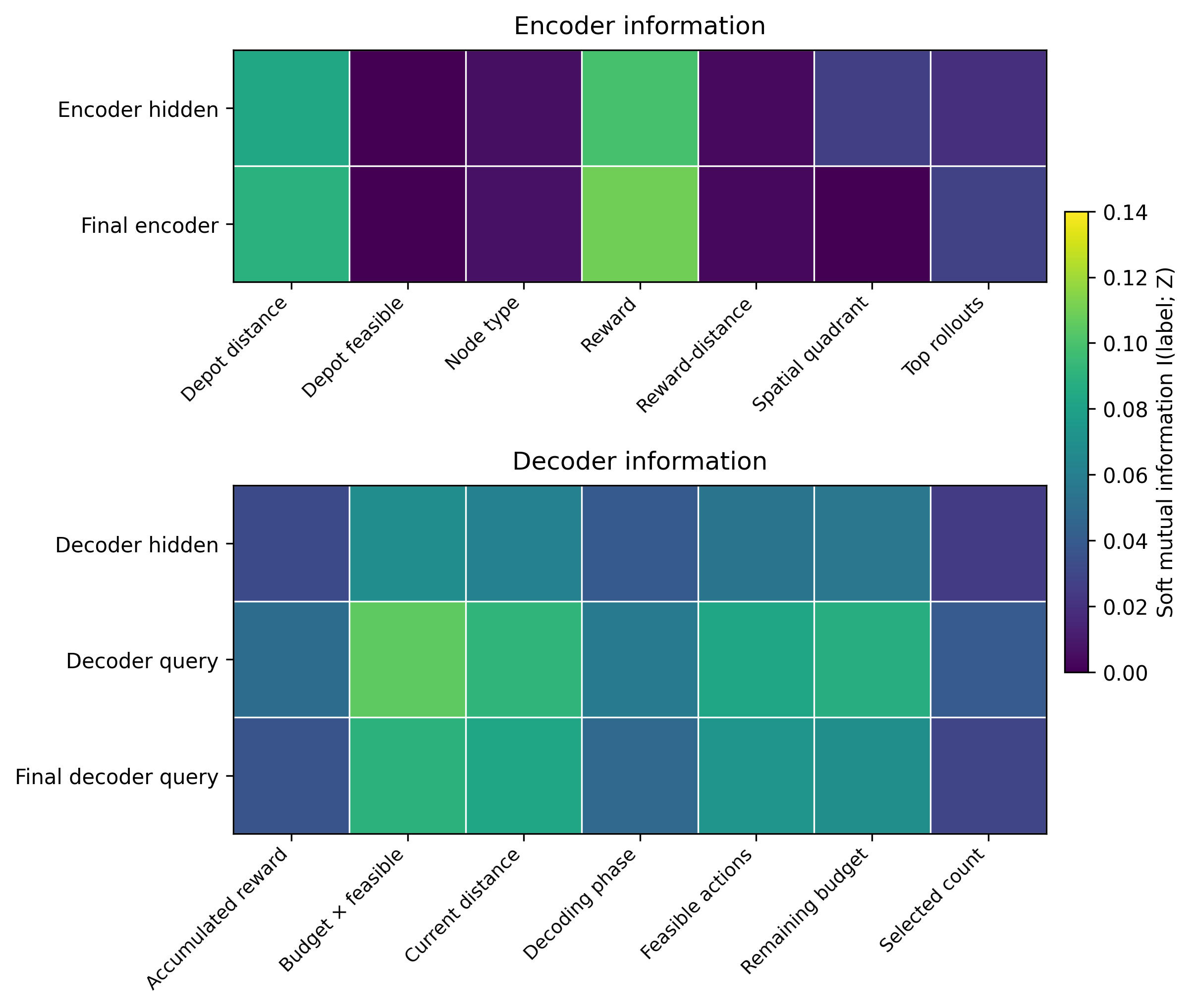}
        \caption{RWPO-TS}
        \label{fig:info_heatmap_rwpo_ls_n100}
    \end{subfigure}

    \caption{
    Mutual information between encoder/decoder representations and task-relevant labels on OP instances with $N=100$.
    The encoder labels include static instance-level information such as depot distance, node reward, spatial quadrant, and top-rollout membership.
    The decoder labels include dynamic route-construction signals such as accumulated reward, budget feasibility, current distance, decoding phase, feasible actions, remaining budget, and selected count.
    Brighter colours indicate that the corresponding representation contains more information about the label.
    }
    \label{fig:info_heatmap_n100}
\end{figure*}

\begin{figure*}[t]
    \centering

    \begin{subfigure}[t]{0.49\textwidth}
        \centering
        \includegraphics[width=\linewidth]{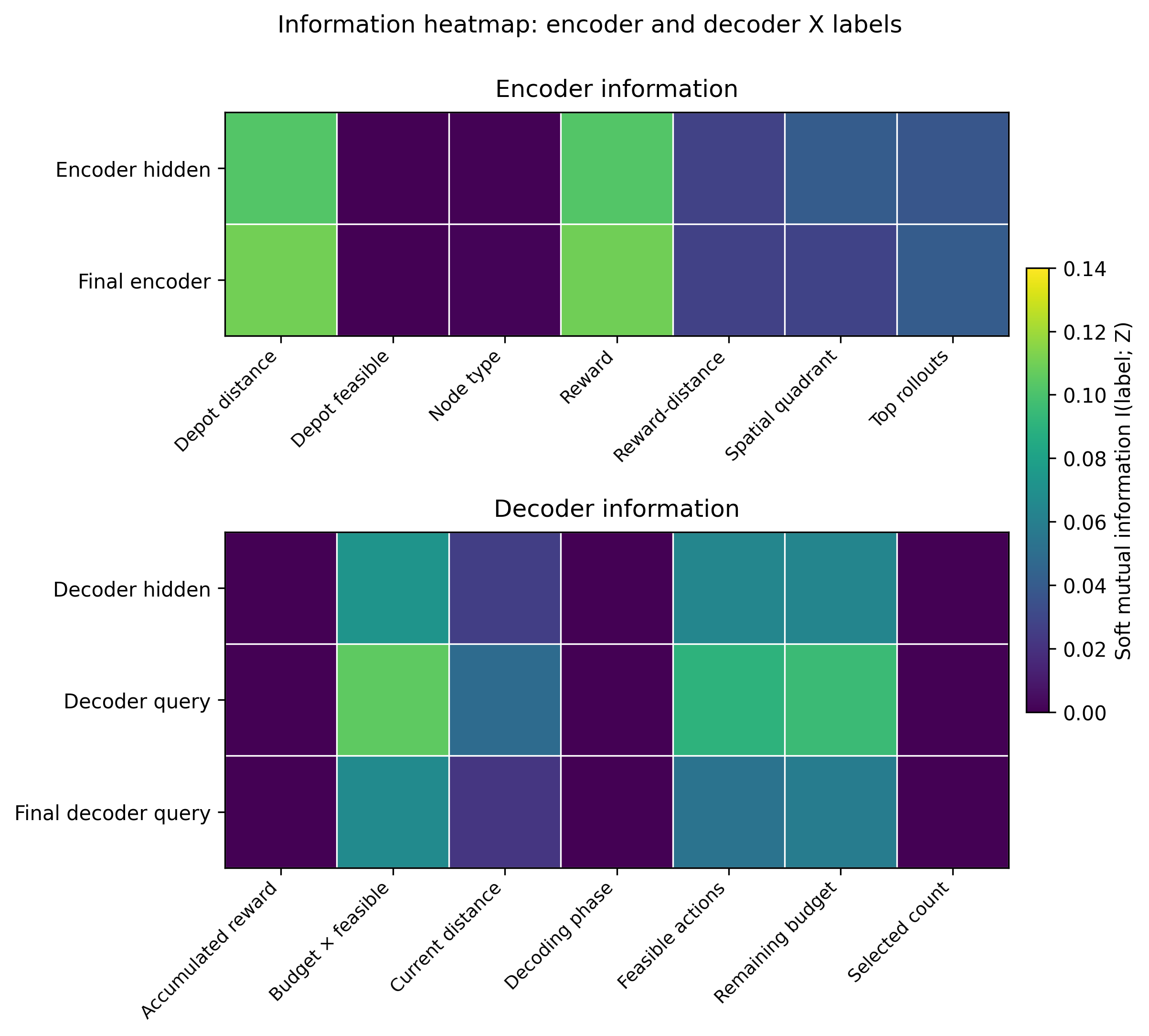}
        \caption{PO}
        \label{fig:info_heatmap_po_n500}
    \end{subfigure}
    \hfill
    \begin{subfigure}[t]{0.49\textwidth}
        \centering
        \includegraphics[width=\linewidth]{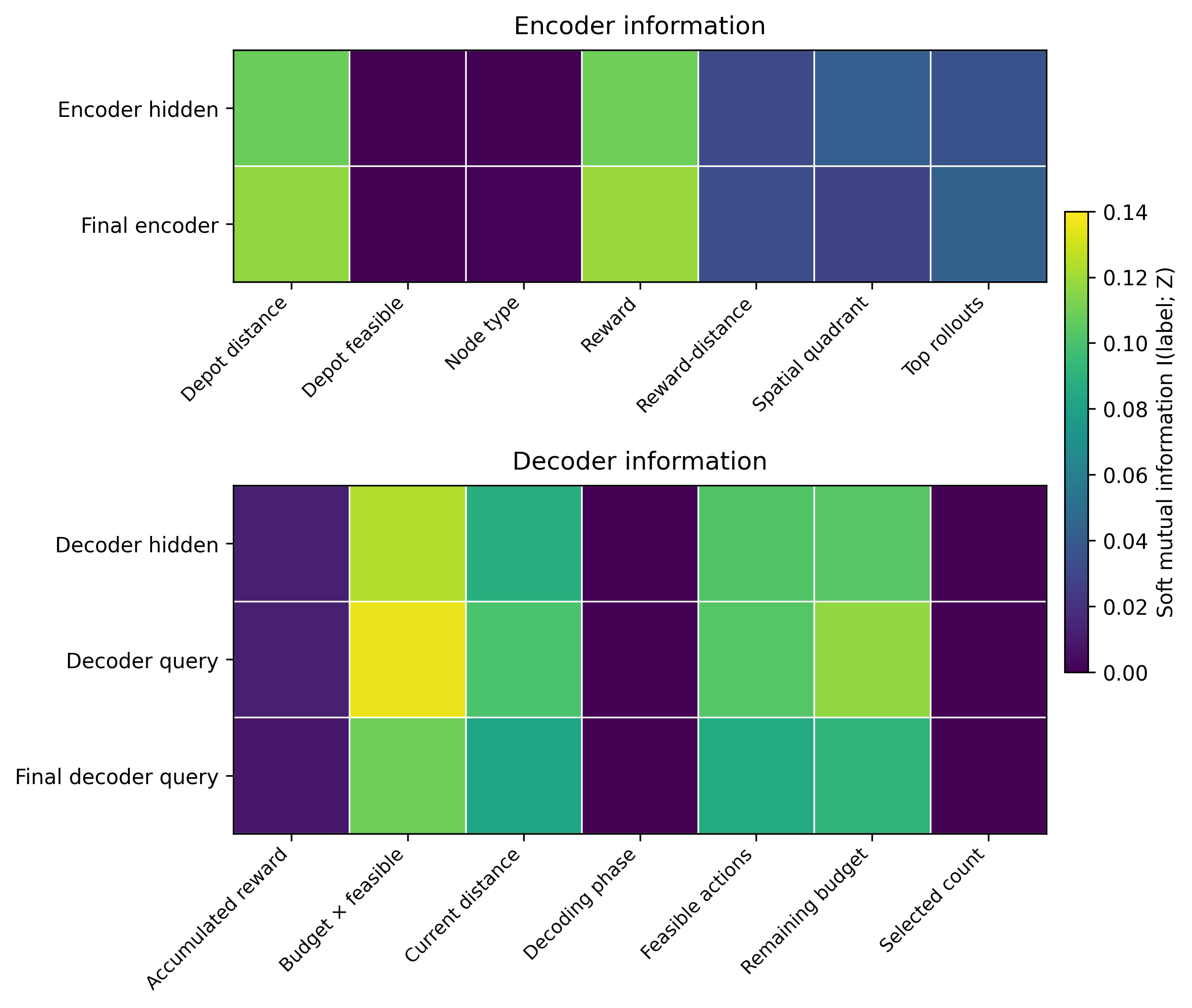}
        \caption{POMO}
        \label{fig:info_heatmap_rl_n500}
    \end{subfigure}

    \vspace{0.5em}

    \begin{subfigure}[t]{0.49\textwidth}
        \centering
        \includegraphics[width=\linewidth]{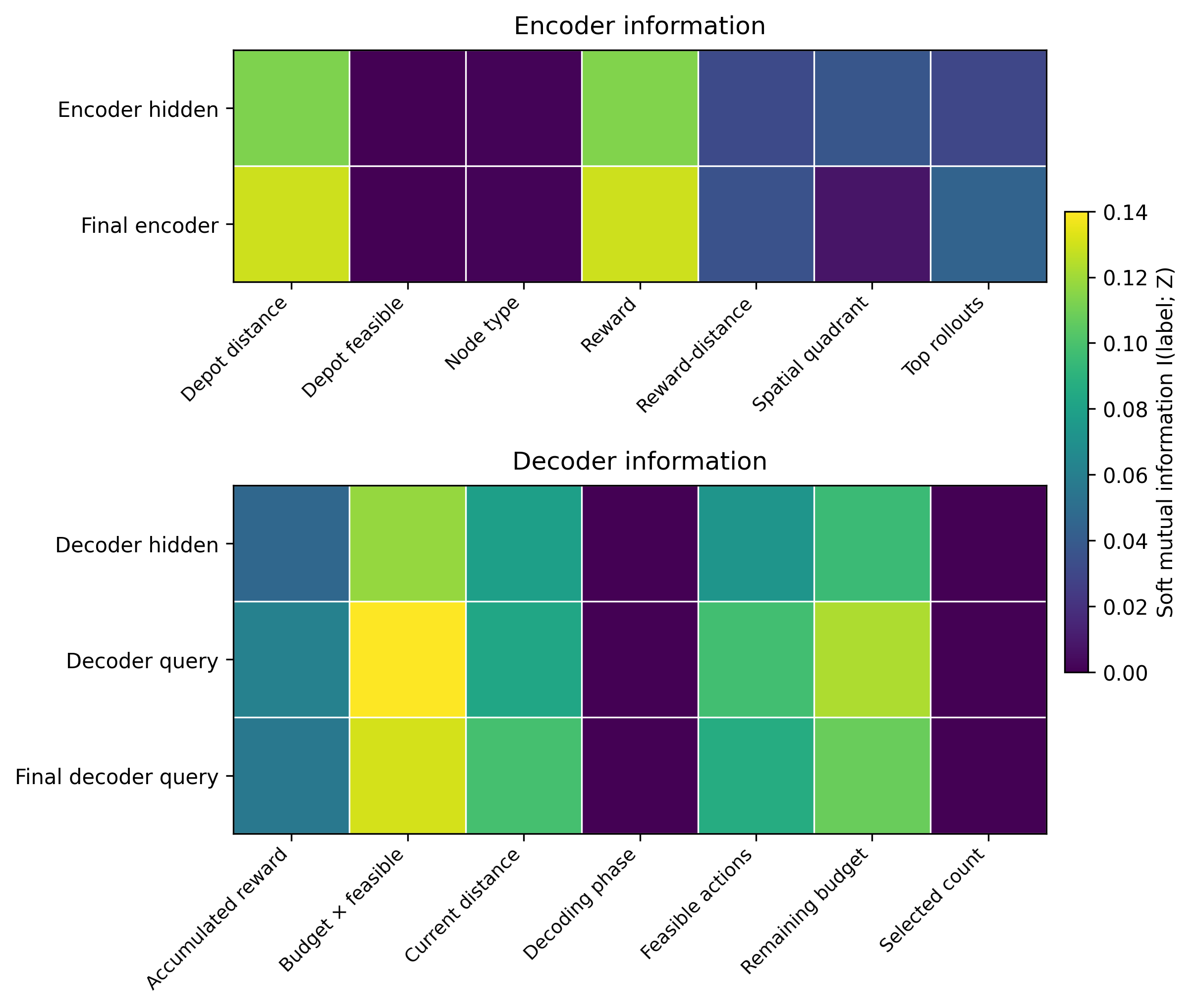}
        \caption{RWPO}
        \label{fig:info_heatmap_rwpo_n500}
    \end{subfigure}
    \hfill
    \begin{subfigure}[t]{0.49\textwidth}
        \centering
        \includegraphics[width=\linewidth]{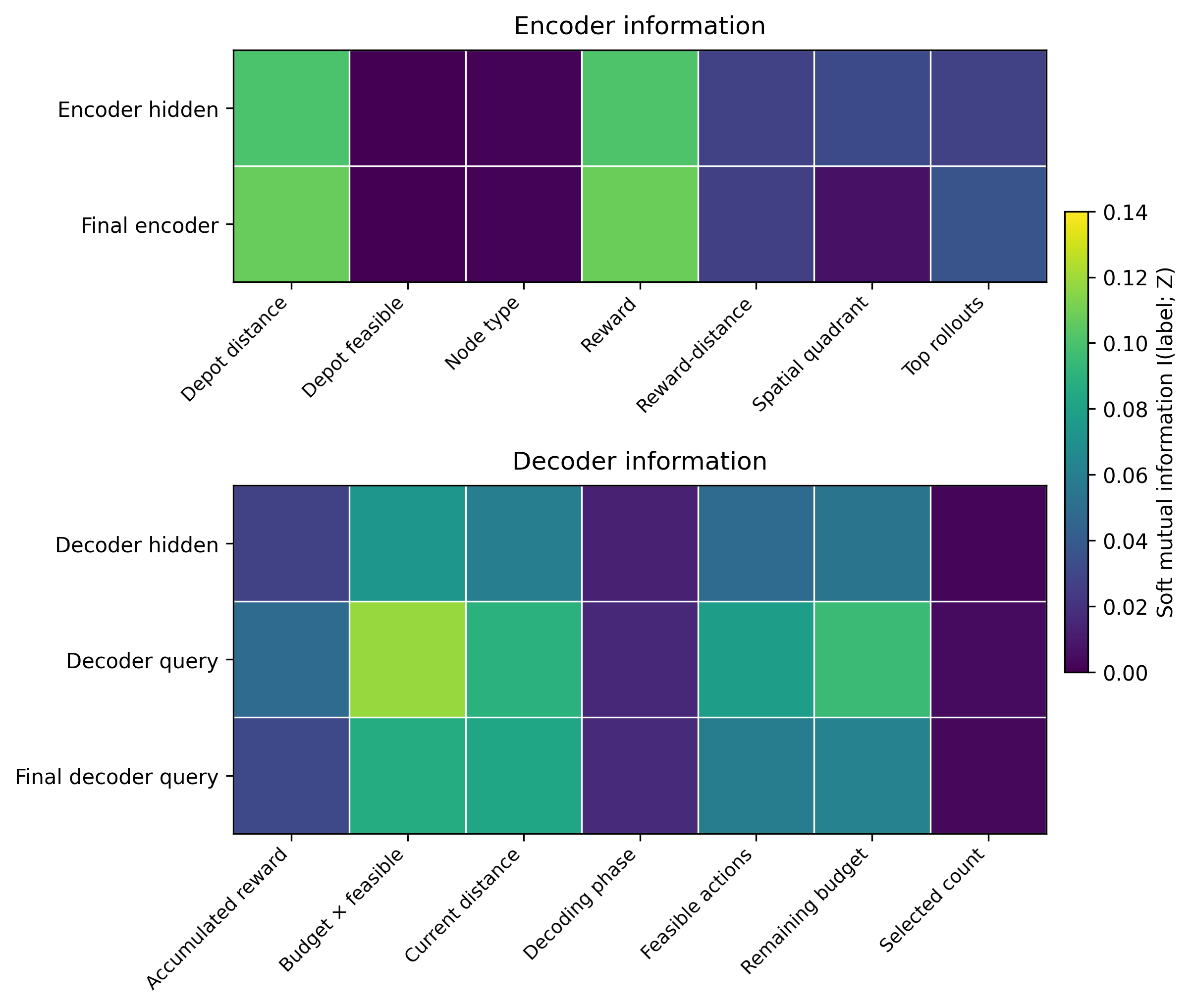}
        \caption{RWPO-FT$_{500}$}
        \label{fig:info_heatmap_rwpo_ft500_n500}
    \end{subfigure}

    \caption{
    Mutual information between encoder/decoder representations and task-relevant labels on OP instances with $N=500$.
    Encoder-side labels describe static instance-level information, including depot distance, node reward, spatial quadrant, and top-rollout membership.
    Decoder-side labels describe dynamic route-construction states, including accumulated reward, budget feasibility, current distance, decoding phase, feasible actions, remaining budget, and selected count.
    Brighter colours indicate that the corresponding representation contains more information about the label.
    }
    \label{fig:info_heatmap_n500}
\end{figure*}

Figures~\ref{fig:info_heatmap_n100} and~\ref{fig:info_heatmap_n500} reveal a relative division of labor between encoder and decoder representations across problem scales. In both settings, the encoder exhibits its strongest mutual information with static instance-level attributes, particularly depot distance and node reward, whereas the decoder-query representation contains substantially more information about dynamic route-construction variables, including budget--feasibility, feasible actions, remaining budget, and current distance. This pattern is consistent with an encoder that represents the candidate set and its reward--geometry attributes, and a decoder that tracks the evolving feasibility state of the partial route. The method comparison shows a clear scale-dependent difference: at $N=500$, PO exhibits a marked attenuation of decoder information for accumulated reward and constraint-related variables, while RWPO retains strong information about budget--feasibility, feasible actions, remaining budget, and current distance. POMO also preserves strong decoder information for several of these labels and is comparable to RWPO in some cases, so the heatmaps do not establish a uniform superiority of RWPO over POMO. The RWPO variants, including RWPO-TS, consistently maintain strong constraint-related decoder information, which is consistent with the hypothesis that reward-weighted post-training encourages more constraint-aware representations. Together with the independently observed performance degradation of PO at $N=500$, these results provide representational evidence consistent with improved large-scale generalization.

\paragraph{Scale-dependent organization of task-relevant decoder information}
To further examine whether the improved generalization of RWPO is reflected in the internal decision process, we measure how much task-relevant information is encoded in the decoder-query representation at each decoding step. Specifically, we consider (i) constraint-related signals, including \emph{feasible actions}, \emph{current distance}, \emph{remaining budget}, and \emph{budget--feasibility interaction}; (ii) objective-related signals, including \emph{accumulated reward}; and (iii) preference-related labels, including \emph{reward quantile} and \emph{top-versus-bottom trajectory membership}.
Together, these labels characterize whether the decoder representation encodes the key quantities needed to balance reward collection, feasibility preservation, and route completion.

\begin{figure*}[t]
    \centering

    \begin{subfigure}[t]{0.32\textwidth}
        \centering
        \includegraphics[width=\linewidth]{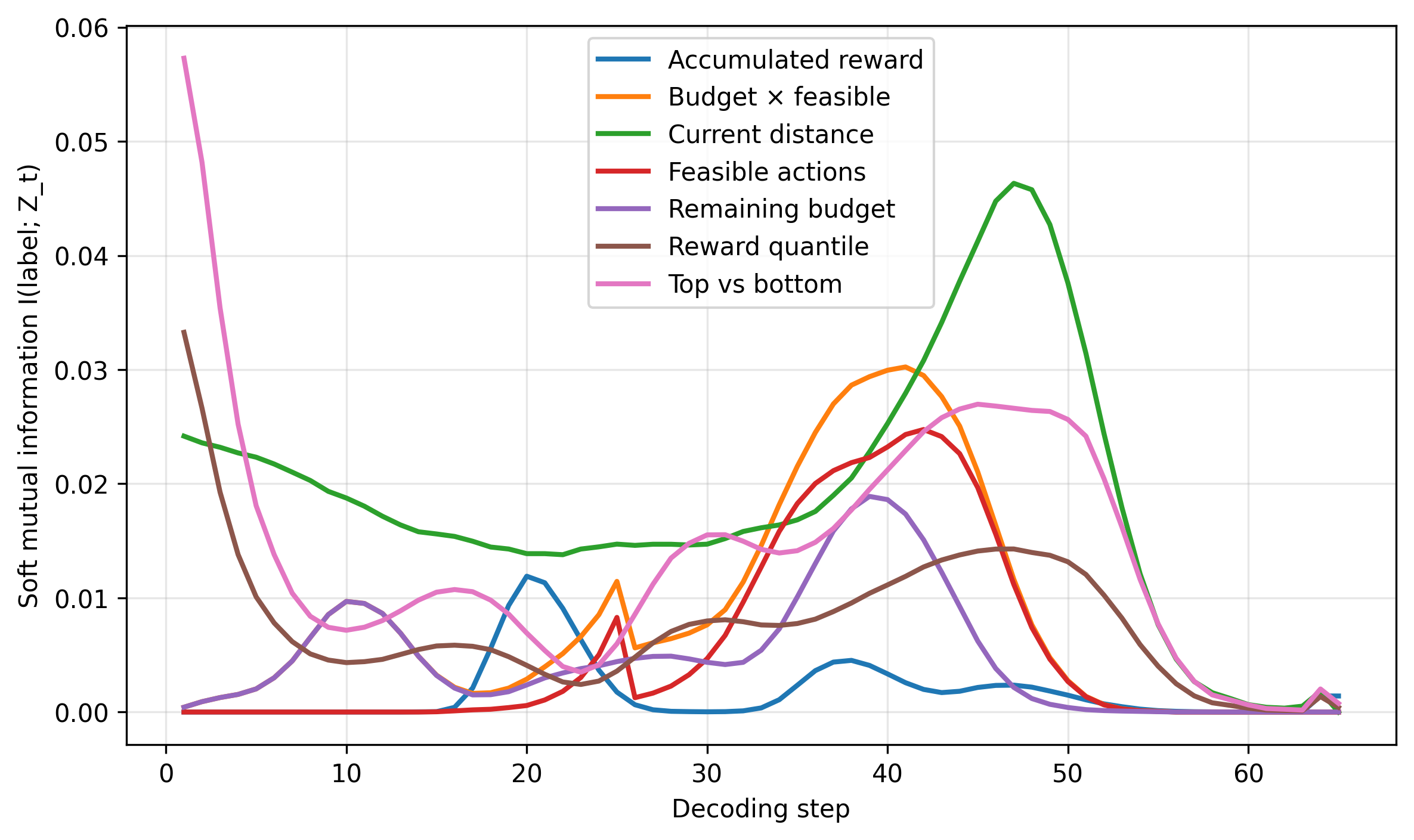}
        \caption{PO, $N=100$}
        \label{fig:info_po_n100}
    \end{subfigure}
    \hfill
    \begin{subfigure}[t]{0.32\textwidth}
        \centering
        \includegraphics[width=\linewidth]{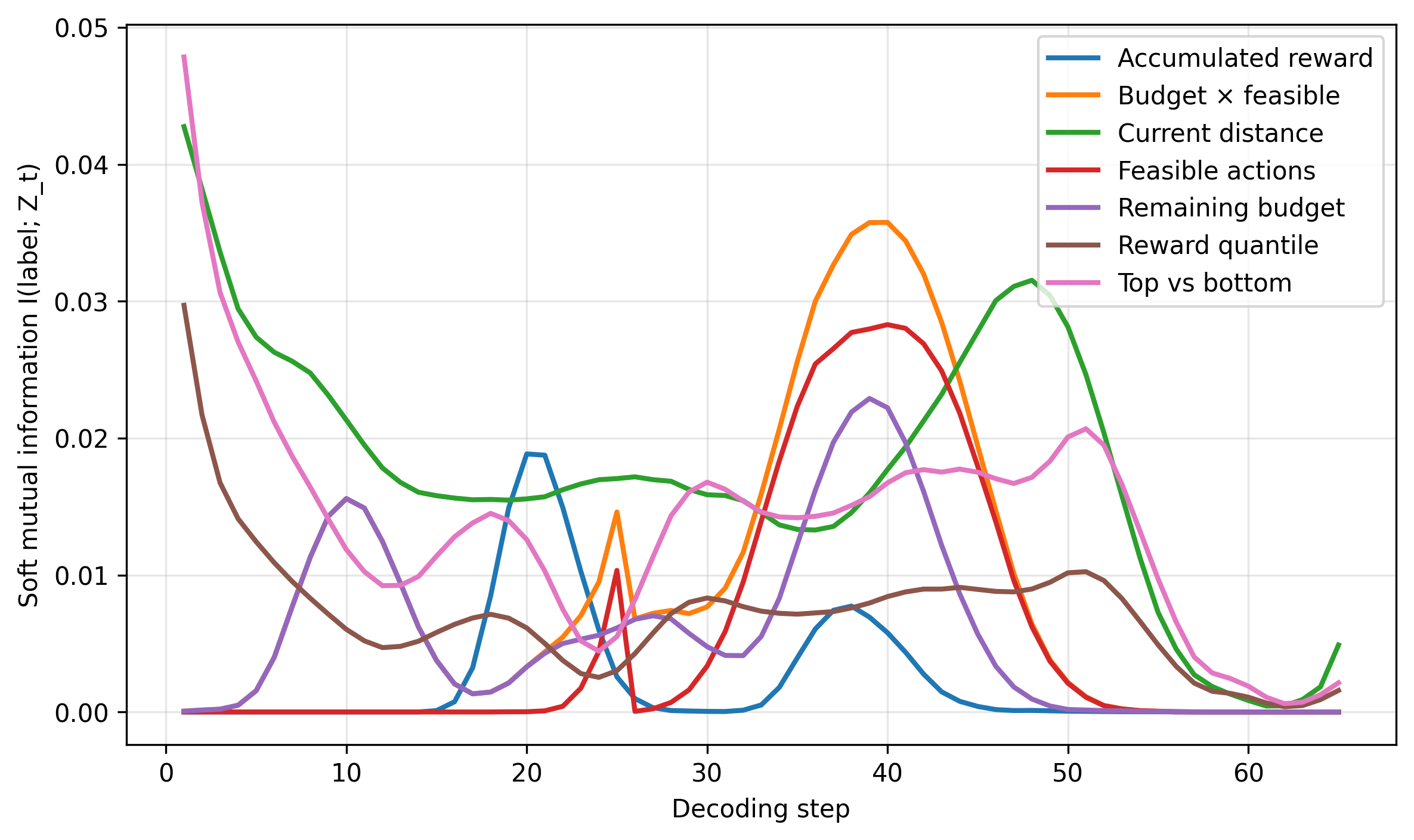}
        \caption{RWPO, $N=100$}
        \label{fig:info_rwpo_n100}
    \end{subfigure}
    \hfill
    \begin{subfigure}[t]{0.32\textwidth}
        \centering
        \includegraphics[width=\linewidth]{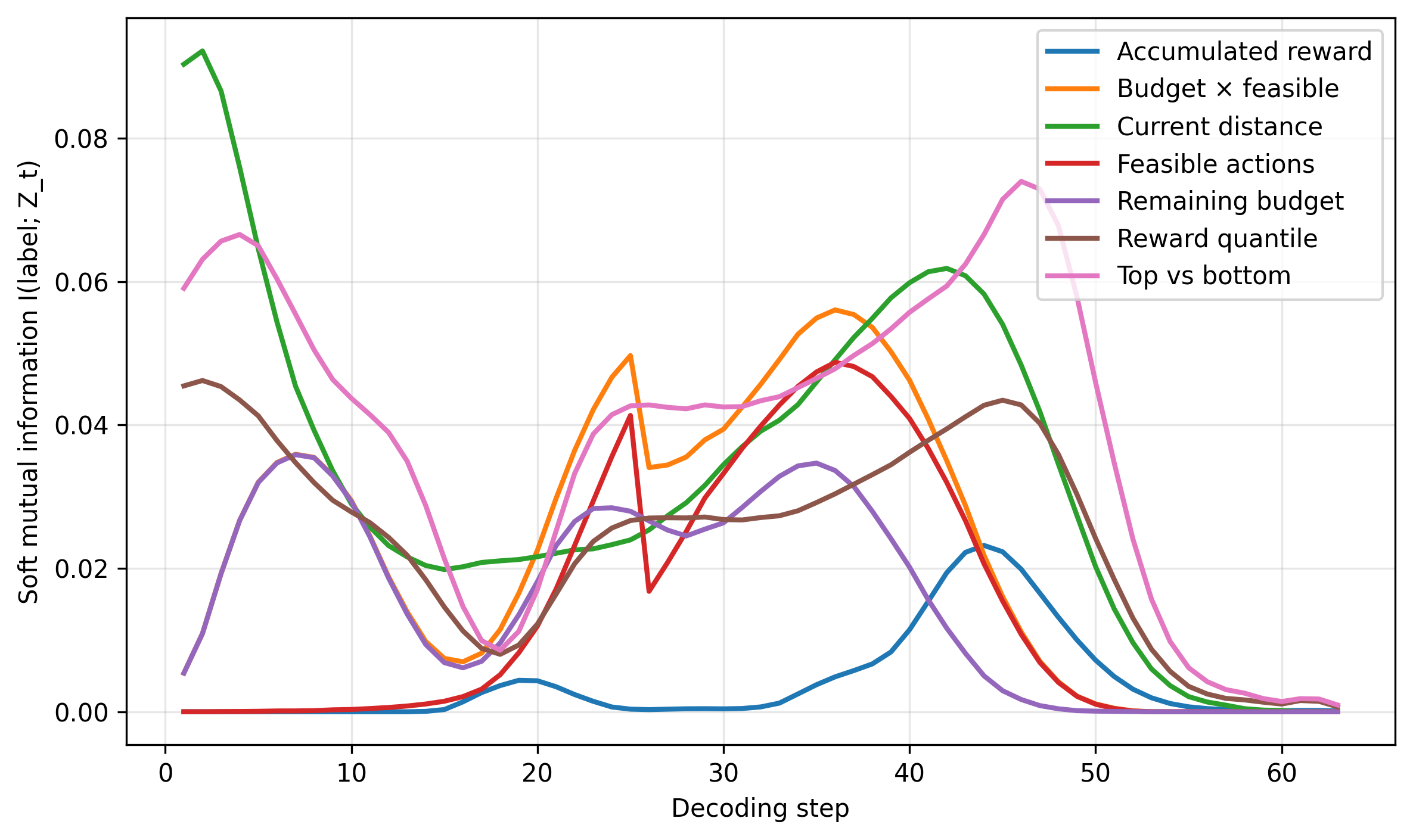}
        \caption{RWPO-TS, $N=100$}
        \label{fig:info_rwpo_ls_n100}
    \end{subfigure}

    \vspace{0.5em}
    \begin{subfigure}[t]{0.32\textwidth}
        \centering
        \includegraphics[width=\linewidth]{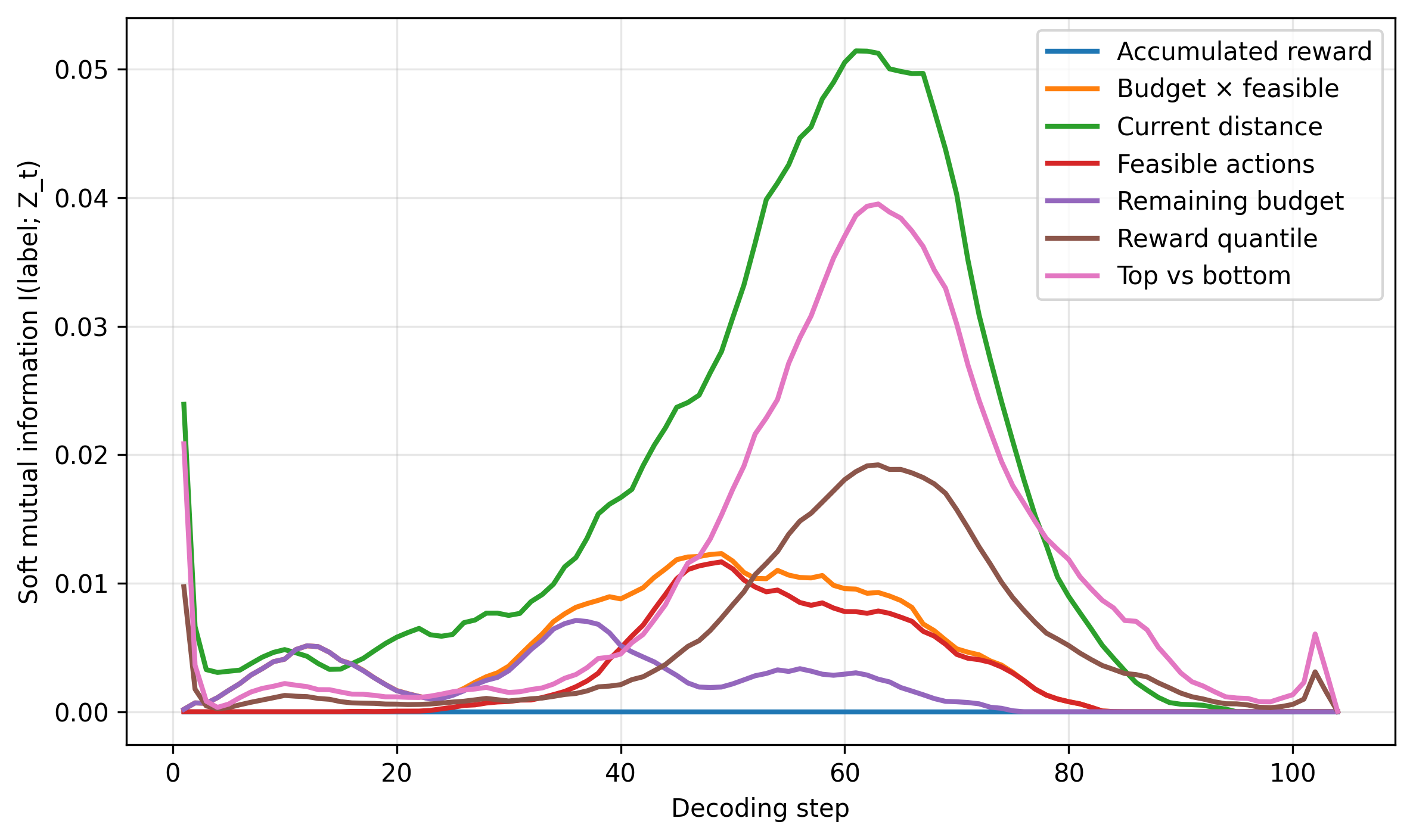}
        \caption{PO, $N=500$}
        \label{fig:info_po_n500}
    \end{subfigure}
    \hfill
    \begin{subfigure}[t]{0.32\textwidth}
        \centering
        \includegraphics[width=\linewidth]{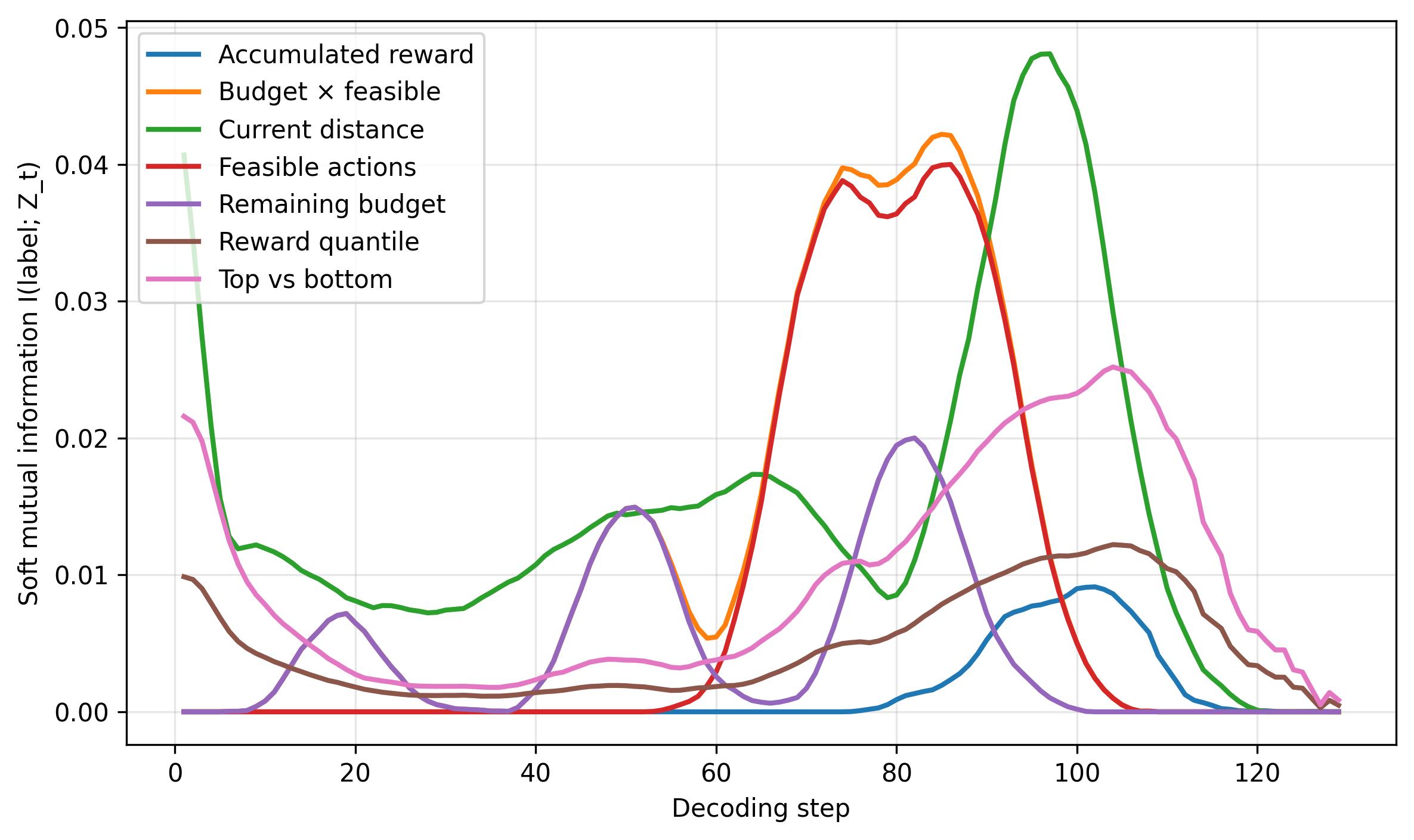}
        \caption{RWPO, $N=500$}
        \label{fig:info_rwpo_n500}
    \end{subfigure}
    \hfill
    \begin{subfigure}[t]{0.32\textwidth}
        \centering
        \includegraphics[width=\linewidth]{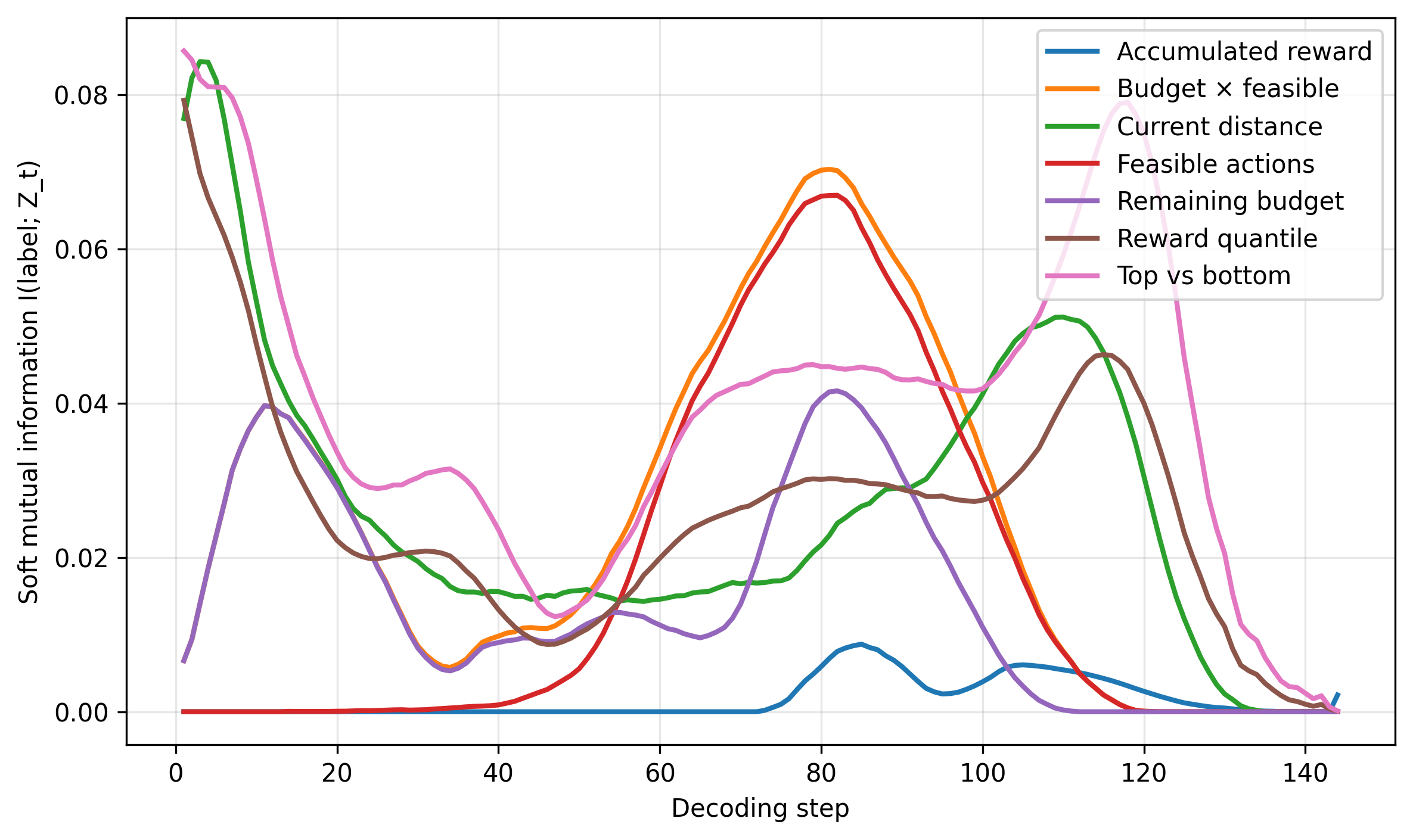}
        \caption{RWPO-FT$_{500}$-TS, $N=500$}
        \label{fig:info_rwpo_ft500_n500}
    \end{subfigure}

    \caption{
    Step-wise mutual information between decoder-query representations and trajectory-level features during OP decoding.
    }
    \label{fig:stepwise_information}
\end{figure*}

Fig.~\ref{fig:stepwise_information} shows that decoder-query representations encode different decision-relevant factors at different stages of route construction. 
The difference becomes much clearer at $N=500$. For PO, the information profile becomes concentrated primarily on current distance, top-versus-bottom trajectory membership, and reward quantile, whereas the information associated with accumulated reward, feasible actions, budget--feasibility interaction, and remaining budget is substantially weaker than at $N=100$. In contrast, RWPO retains strong encoding of feasible actions and budget-related signals during the middle stages, including a pronounced peak in the remaining-budget signal, before shifting toward current distance, top-versus-bottom trajectory membership, and reward quantile during later decoding. Thus, as the instance size grows, RWPO maintains a staged decision process that first emphasizes feasibility and budget management and subsequently emphasizes distance and trajectory quality, while PO becomes more narrowly focused on distance and outcome-related signals. The RWPO-FT$_{500}$-TS variants further amplify this qualitative pattern through larger and more sustained information peaks. Combined with the observed degradation of PO performance at $N=500$, these results are consistent with the hypothesis that \textit{RWPO improves large-scale generalization by preserving constraint- and objective-relevant information in the decoder representation.}

\subsection{Training Curves} \label{app:training_curves}

\begin{figure*}[t]
    \centering
    \begin{subfigure}[t]{0.32\textwidth}
        \centering
        \includegraphics[width=\linewidth]{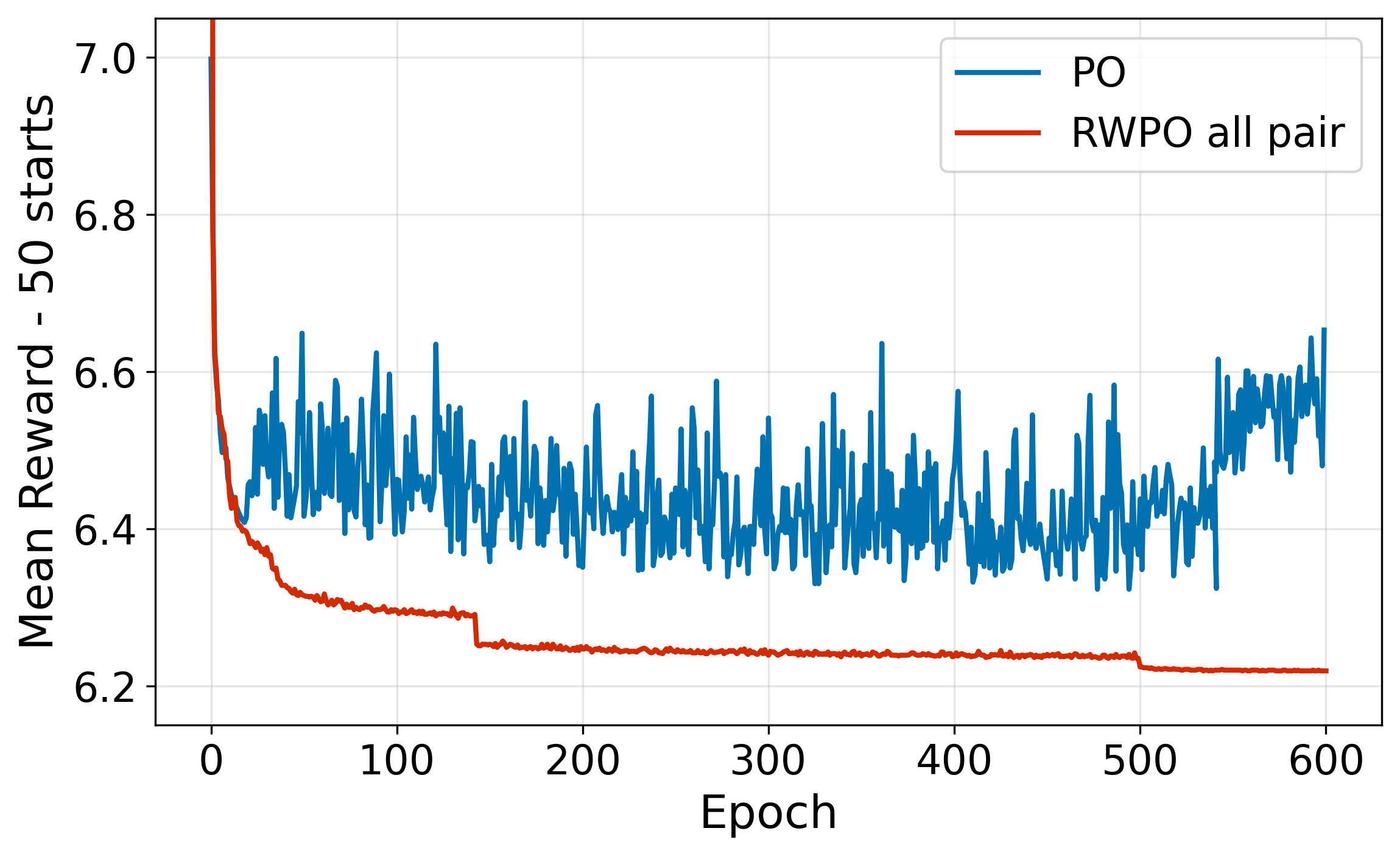}
        \caption{Mean reward on SPCTSP100}
        \label{fig:spctsp_topk_mean_reward}
    \end{subfigure}
    \hfill
    \begin{subfigure}[t]{0.32\textwidth}
        \centering
        \includegraphics[width=\linewidth]{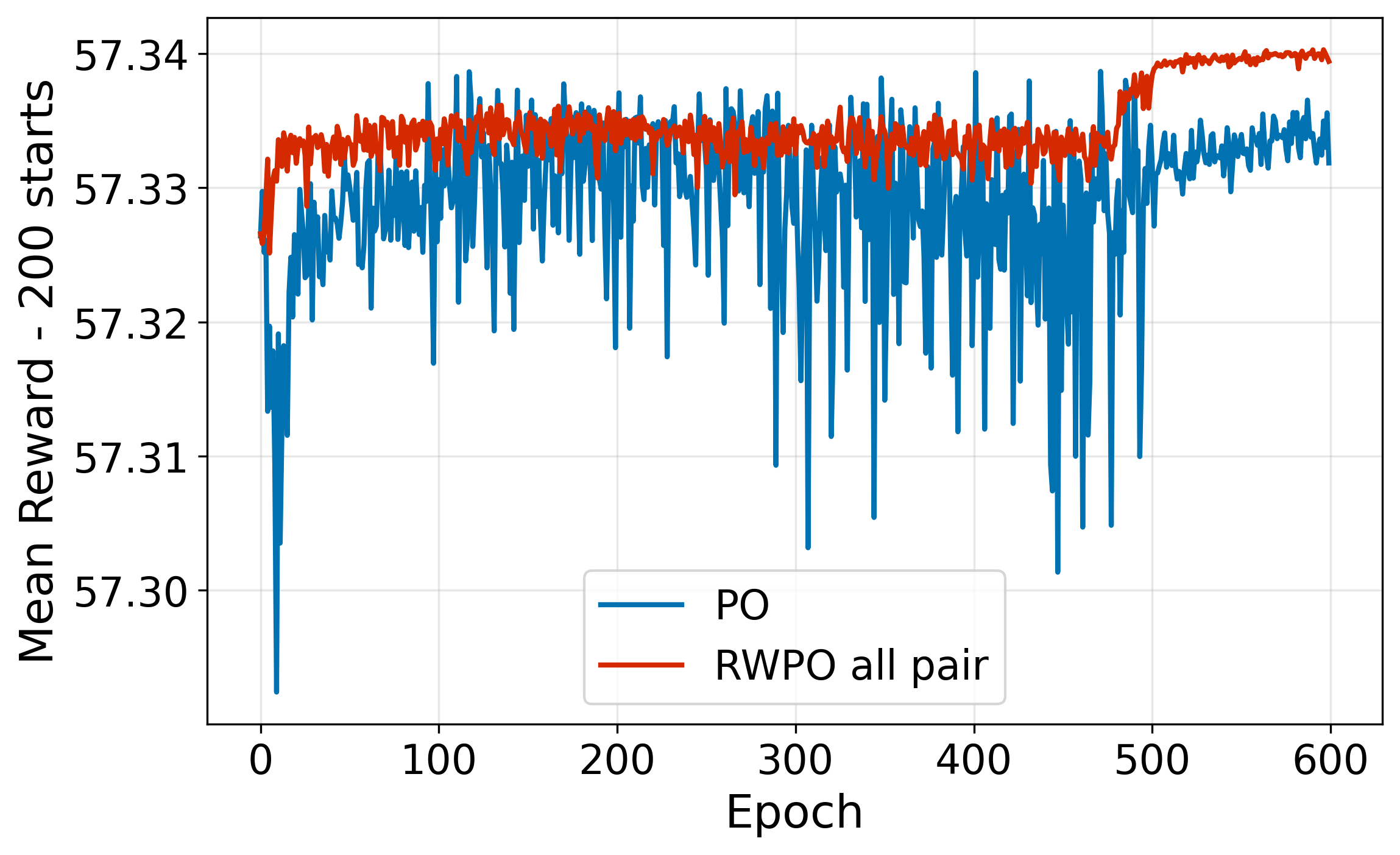}
        \caption{Mean reward on KP200}
        \label{fig:kp_topk_mean_reward}
    \end{subfigure}
    \hfill
    \begin{subfigure}[t]{0.32\textwidth}
        \centering
        \includegraphics[width=\linewidth]{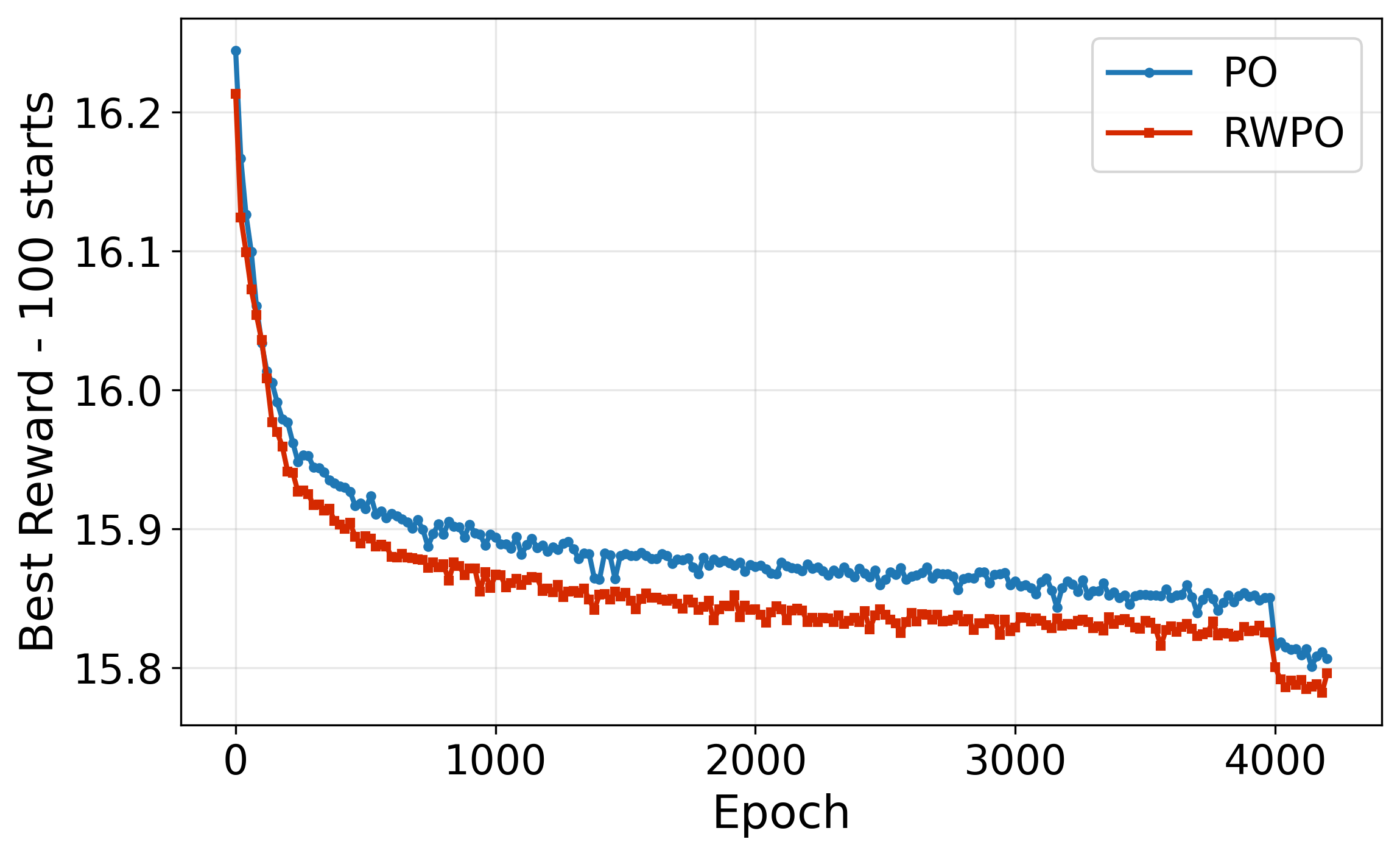}
        \caption{Best reward on CVRP100}
        \label{fig:cvrp_best_reward}
    \end{subfigure}
    \caption{Training dynamics of PO and RWPO on SPCTSP100, KP200, and CVRP100. All three panels report the mean validation objective over all top-$k$ starts. For SPCTSP and CVRP, lower objective values indicate better performance and for KP, vice versa. RWPO shows smoother and more stable optimization than PO, with the clearest final improvement on SPCTSP100 and CVRP100. On KP200, both methods reach similar final values, but RWPO exhibits substantially lower variance during training. These results complement the training curves in Fig.~\ref{fig:op_training_curves}.}
    \label{fig:training_curves_po_rwpo}
\end{figure*}

The stability advantage of RWPO extends beyond the OP. On SPCTSP100, the RWPO all-pair variant rapidly improves the mean objective and subsequently follows a smooth, low-variance trajectory, whereas PO remains highly oscillatory throughout training (Fig.\ref{fig:spctsp_topk_mean_reward}). On KP200, although PO and RWPO converge to similar final mean values, RWPO exhibits substantially lower variance, indicating more consistent optimization across the $k$ parallel starts (Fig.\ref{fig:kp_topk_mean_reward}). On CVRP100, RWPO maintains a lower objective cost than PO for nearly the entire training process, finishing with a small but consistent advantage in the best-objective metric (Fig.\ref{fig:cvrp_best_reward}). These cross-domain findings corroborate the OP and PCTSP results (Fig.\ref{fig:op_training_curves} in Section~\ref{sec:ablation_study}): while purely ordinal PO remains competitive when evaluating only the single best rollout (Fig.\ref{fig:op_best_reward50} in Section~\ref{sec:ablation_study}), its batch-wide mean performance is considerably noisier and less stable (Fig.\ref{fig:op_mean_reward50} in Section~\ref{sec:ablation_study}). In contrast, RWPO consistently improves the entire rollout distribution. This behavior directly reflects the RWPO all-pair mechanism, which extracts dense learning signals across all sampled trajectories while explicitly scaling updates by the true reward differences.

\section{Ablation Study of Hyperparameters}

\begin{table*}[t]
\centering
\caption{Ablation results for different loss functions and hyperparameter settings on OP50.
Lower (more negative) objective values indicate better performance.}
\label{tab:ablation_results_op50}
\resizebox{\textwidth}{!}{%
\begin{tabular}{l c c c c c c c c}
\toprule
Loss type & $\beta$ & $\lambda$ &
Best Reward @ 20 $\downarrow$ &
Mean Reward @ 20 $\downarrow$ &
Best Reward @ 50 $\downarrow$ &
Mean Reward @ 50 $\downarrow$ &
Best Reward $\downarrow$ (epoch) &
\\
\midrule
RWPO base & 0.0 & 0.5 &
-16.022 &
-15.155 &
-16.117 &
-15.296 &
-16.117 (50) &
\\

RWPO base & 0.2 & 0.5 &
-15.990 &
-15.071 &
-16.080 &
-15.214 &
-16.080 (50) &
\\

RWPO all pair & 0.3 & 0.0 &
-15.913 &
-14.950 &
-12.256 &
-10.205 &
-15.953 (15) &
\\

RWPO all pair & 0.0 & 0.5 &
\textbf{-16.122} &
\textbf{-15.336} &
\textbf{-16.172} &
\textbf{-15.434} &
\textbf{-16.176 (45)} &
\\

RWPO all pair & 0.2 & 0.5 &
-16.096 &
-15.292 &
-16.143 &
-15.387 &
-16.160 (45) &
\\

RWPO all pair & 0.3 & 0.5 &
-16.031 &
-15.194 &
-16.107 &
-15.332 &
-16.111 (49) &
\\

RWPO all pair & 0.3 & 1.0 &
-15.967 &
-15.037 &
-16.065 &
-15.254 &
-16.080 (50) &
\\

RWPO all pair & 0.3 & 2.0 &
-15.942 &
-14.996 &
-16.070 &
-15.223 &
-16.070 (50) &
\\
\bottomrule
\end{tabular}}
\end{table*}


\begin{table*}[t]
\centering
\caption{Ablation results for different loss functions and hyperparameter settings.
Lower (more negative) objective values indicate better performance.}
\label{tab:ablation_results_op100}
\resizebox{\textwidth}{!}{%
\begin{tabular}{l c c c c c c c c}
\toprule
Loss type & $\beta$ & $\lambda$ &
Best Reward @ 10 $\downarrow$ &
Mean Reward @ 10 $\downarrow$ &
Best Reward @ 20 $\downarrow$ &
Mean Reward @ 20 $\downarrow$ &
Best Reward $\downarrow$ (epoch) &
\\
\midrule


RWPO base & 0.0 & 0.5 &
-31.817 & -30.394 &
-32.230 & -30.872 &
-32.230 (19) \\

RWPO base & 0.3 & 0.5 &
-31.710 & -30.190 &
-32.221 & -30.827 &
-32.221 (19) \\

RWPO all pair & 0.3 & 0.0 &
-32.049 & -30.556 &
-32.294 & -30.824 &
-32.294 (19) \\

RWPO all pair & 0.0 & 0.5 &
-32.404 & -30.989 &
-32.644 & -31.300 &
-32.644 (19) \\

RWPO all pair & 0.2 & 0.5 &
\textbf{-32.678} & \textbf{-31.353} &
\textbf{-32.897} & \textbf{-31.684} &
\textbf{-32.897 (19)} \\

RWPO all pair & 0.3 & 0.5 &
-32.569 & -31.210 &
-32.815 & -31.585 &
-32.815 (19) \\

RWPO all pair & 0.3 & 1.0 &
-32.086 & -30.655 &
-32.116 & -30.754 &
-32.116 (18) \\

RWPO all pair & 0.3 & 2.0 &
-32.363 & -30.925 &
-32.538 & -31.093 &
-32.538 (19) \\
\bottomrule
\end{tabular}}
\end{table*}

We conduct a reduced-budget sensitivity analysis on OP50 and OP100 to evaluate the effects of the reward-weighting strength ($\lambda$) and the length-normalization factor ($\beta$). Because these instances are computationally lightweight, they permit efficient hyperparameter screening. Rather than exhaustively tuning hyperparameters, our objective is to identify robust configurations that transfer reliably across problem scales.

As shown in Tab.~\ref{tab:ablation_results_op50} and \ref{tab:ablation_results_op100}, introducing reward-gap weighting consistently improves the final validation objective compared to the purely ordinal baseline ($\lambda=0$). Moderate weighting strengths yield the most stable improvements across both OP50 and OP100; conversely, excessively high values provide no additional gains and risk destabilizing the preference updates. We therefore adopt $\lambda=0.5$ as a conservative default for the main OP experiments.

Furthermore, we observe that the optimal length-normalization strength scales directly with instance size. While weaker normalization suffices for OP50, OP100 strictly benefits from a stronger $\beta$ adjustment. This occurs because larger instances generate longer trajectories, inducing greater variance in cumulative log-likelihood magnitudes. To proactively handle this variance at larger scales, we fix $\beta=0.3$ across all OP experiments. We apply this same validation-based procedure to determine stable configurations for the remaining SCOP benchmarks prior to final evaluation.


\end{document}